\documentclass[final,5p,times,twocolumn]{elsarticle}

\usepackage{amssymb}
\usepackage{mathptmx}      

\DeclareMathAlphabet{\mathcal}{OMS}{cmsy}{m}{n} 
\usepackage{latexsym}
\usepackage{graphicx}
\usepackage{pbox}
\usepackage{amsfonts,amssymb}
\usepackage{amsmath}
\usepackage{changepage}
\usepackage{tikz}
\setcitestyle{sort&compress}
\usepackage{soul}
\usepackage{multirow}
\usepackage{booktabs}
\usepackage{threeparttable}
\usepackage{floatrow}
\usepackage{algorithm}
\usepackage{algcompatible}
\usepackage{tabularx}
\usepackage[titletoc,title]{appendix}
\usepackage{hyperref}
\usepackage{pbox}
\usepackage{subfig}
\hypersetup{colorlinks,linkcolor={blue},citecolor={blue},urlcolor={blue}}
\usepackage{url}
\usepackage{bm}
\usepackage[ampersand]{easylist}

\journal{arXiv}

\begin{document}

\begin{frontmatter}



\title{Multi-Label Zero-Shot Human Action Recognition via Joint Latent Ranking Embedding}


\author{Qian Wang,  Ke Chen}

\address{The University of Manchester, UK. \\qian.wang173@hotmail.com, ke.chen@manchester.ac.uk }

\begin{abstract}
Human action recognition refers to automatic recognizing human actions from a video clip, which is one of the most challenging tasks in computer vision. Due to the fact that annotating video data is laborious and time-consuming, most of the existing works in human action recognition are limited to a number of small scale benchmark datasets where there are a small number of video clips associated with only a few human actions and a video clip often contains only a single action. In reality, however, there often exist multiple human actions in a video stream. Such a video stream is often weakly-annotated with a set of relevant human action labels at a global level rather than assigning each label to a specific video episode corresponding to a single action, which leads to a multi-label learning problem. Furthermore, there are a great number of meaningful human actions in reality but it would be extremely difficult, if not impossible, to collect/annotate video clips regarding all of various human actions, which leads to a zero-shot learning scenario. To the best of our knowledge, there is no work that has addressed all the above issues together in human action recognition. In this paper, we formulate a real-world  human action recognition task as a multi-label zero-shot learning problem and propose a framework to tackle this problem in a holistic way. Our framework holistically tackles the issue of unknown temporal boundaries between different actions for multi-label learning and exploits the side information regarding the semantic relationship between different human actions for knowledge transfer. As a result, our framework leads to a joint latent ranking embedding for multi-label zero-shot human action recognition. A novel neural architecture of two component models and an alternate learning algorithm are proposed to carry out the joint latent ranking embedding learning. Thus, multi-label zero-shot recognition is done by measuring relatedness scores of action labels to a test video clip in the joint latent visual and semantic embedding spaces. We evaluate our framework with different settings, including a novel data split scheme designed especially for evaluating multi-label zero-shot learning, on two weakly annotated multi-label human action datasets: \emph{Breakfast} and \emph{Charades}. The experimental results demonstrate the effectiveness of our framework in multi-label zero-shot human action recognition.
\end{abstract}


\end{frontmatter}


\section{Introduction}
\label{sect:intro}

As one of the most challenging tasks in computer vision, human action recognition refers to automatic recognizing human actions conveyed in a video clip. In last two decades, human action recognition has been extensively studied. As there are many different human actions in reality, this task is generally formulated as a multi-class classification problem. To train a multi-class classifier for human action recognition, a great number of examples for each single action are required in the current setting. To collect such training examples, one needs to manually trim a video stream to ensure that there is only one human action appearing in a trimmed video episode. This annotation process is laborious and time-consuming and there is hence no large-scale dataset with ``fine-grained" annotation for human action recognition. In contrast to ImageNet \cite{deng2009imagenet} for object recognition, where it consists of a total of 3.2 million cleanly labelled images spreading over 5,247 categories, there are much fewer annotated video clips involving only a small number of human actions. For instance,  HMDB51 and UCF101 are among the most commonly used benchmark datasets in human action recognition, where there are 6,676 and 13,320 instances of only 51 and 101 different human actions, respectively. The limitation of human action datasets in such a scale has become an obstacle in developing a large-scale human action recognition system.

In a real scenario, a video clip often conveys multiple human actions corresponding to different concepts. Hence, a set of multiple action labels have to be used to characterize its complete semantics underlying human actions conveyed in this video clip.  For example,  video clips on YouTube are usually uploaded by users along with some descriptive terms that can be used to infer the human actions conveyed in those video clips. In this circumstance, descriptive terms may be viewed as a set of coherent labels that collectively characterize the semantics at a global level. Recently, a very large multi-label video dataset YouTube-8M \cite{abu2016youtube} has been collected by Google Research. Although the dataset is not restricted to human action video clips, it paves a new way for various video analyses including human action recognition. One of essential video analysis problems on such a data set may be formulated as multi-label learning that predicts a set of labels associated to a given instance or a set of relatedness scores corresponding to the candidate labels that could characterize this instance. In multi-label learning, a training example often consists of an input instance and a set of labels associated with this instance at a global level (no need of explicitly associating each of those labels to a relevant object within this instance).  While multi-label data are common in many domains and multi-label classification has been studied under different applications \cite{zhang2014multi-label}, e.g., semantic image tagging, text categorization and gene functionality prediction, only few works are pertinent to multi-label human action recognition in literature due to a lack of human action datasets annotated with multiple class labels. To fill in this gap, a dataset dubbed Charades \cite{sigurdsson2016hollywood} was collected especially for multi-label human action recognition and made publicly available very recently.  In addition, other datasets collected for different tasks were also considered to be used in multi-label human action recognition. Thus, such data sets provide a proper test bed for multi-label human action recognition studies.

Multi-label human action recognition often has to work on weakly labelled video data, i.e., the training data are annotated at the video level without exhaustively trimming and annotating multiple action episodes. While it is easier to collect such video clips associated with a set of labels at a global level than those with ``fine-grained" annotation,  it would be still very challenging to collect all the training examples due to the existence of many different human actions. \emph{Zero-shot learning} (ZSL) provides an alternative solution to alleviate this problem. ZSL aims to recognize the instances belonging to novel classes which are not seen during training. It has been formally shown that under certain conditions, a ZSL system trained on a dataset of finite classes could be used to predict infinite number of classes unseen during the learning \cite{zhang2016infinite}. Under the ZSL framework, we merely need to collect and annotate training examples for a moderate set of training classes and expect that a large number of novel classes can be recognized via exploiting the semantic relationship between different human actions. To this end, a ZSL algorithm needs to transfer the knowledge regarding the relations between visual features and class label semantics learned from known or training classes to unseen or test classes. The knowledge transfer is enabled by modelling the \textit{semantic representations} of different classes, which can be easily obtained from side information, e.g., descriptive texts, with a much less effort than collecting and annotating visual data. Nevertheless, most of the existing ZSL methods were proposed to tackle single-label ZSL problems but multi-label ZSL problems are much more complicated, leading to additional challenge that do not exist in single-label ZSL. Although some of single-label ZSL methods might be extensible to multi-label scenarios, their effectiveness of different ZSL algorithms have not been extensively investigated in the multi-label learning scenarios. To the best of our knowledge, there exists no work in multi-label zero-shot human action recognition.

In this paper, we address the multi-label ZSL issues in the context of human action recognition. In our problem, the training video data are weakly labelled so that the exact temporal-spatial locations of multiple human actions in a video clip remain unknown. In addition, training examples consisting of visual instances and their corresponding label sets of multiple labels are only available for those associated with training/known labels, a subset of the action label collection considered in the recognition stage. Thus, the nature of multi-label zero-shot human action recognition poses several challenges that do not exist in static data and single-label ZSL. To tackle all the challenges in a holistic way, we propose a novel \emph{joint latent ranking embedding} framework. The framework aims to learn a joint latent ranking embedding from visual and semantic domains. By using the learned joint latent ranking embedding, any visual instances and any action labels can be mapped into the joint latent visual and semantic embedding spaces where positive connections between visual instances and action labels rank ahead of negative ones.
Thus, any human actions can be recognized regardless of known or unseen actions during learning. Our framework consists of two component models: \emph{visual} and \emph{semantic} models. The visual model learns mapping a visual instance into the latent \textit{visual embedding} space, while the semantic model learns mapping action labels into the latent \textit{semantic embedding} space. The visual and the semantic models are tightly coupled to learn a proper ranking that works in the joint latent visual and semantic embedding spaces with an alternate learning algorithm on training examples annotated with only known action labels. In the test, multi-label zero-shot recognition is done by measuring relatedness scores of action labels to a test visual clip in the joint latent visual and semantic embedding spaces.

Our main contributions in this paper are summarized as follows:
\begin{itemize}
	\item By considering real scenarios, we formulate general human action recognition as a \emph{multi-label zero-shot learning} problem . To the best of our knowledge, our work presented in this paper is the first attempt in studying human action recognition from a multi-label zero-shot learning perspective, which tackles several technical challenges pertaining to this problem in a holistic way.
	\item To address the multi-label zero-shot issues arising from weakly annotated data for human action recognition, we propose a novel joint latent ranking embedding framework consisting of visual and semantic embedding models. To train two embedding models effectively, we come up with a learning algorithm that alternately optimizes the parameters in two embedding models via minimizing the proper rank loss functions.
	\item To test the performance of our proposed framework, we conduct a thorough evaluation via a comparative study on two benchmark multi-label human action datasets, \emph{Breakfast} and \emph{Charades}, with various evaluation metrics and different settings including a novel data split protocol simulating a real scenario of multi-label zero-shot human action recognition.
\end{itemize}

The rest of this paper is organized as follows. Section 2 reviews related works. Section 3 presents our framework for multi-label zero-shot human action recognition. Section 4 describes our experimental settings, and Section 5 reports the experimental results. The last section draws conclusions.

\section{Related Work}

In this section, we review the existing works relating to multi-label human action recognition, especially for those applicable to multi-label zero-shot learning scenarios, and point out the limitations of the existing works. We first overview existing multi-label classification methods and then focus on the existing works in multi-label ZSL learning despite the fact that none of such multi-label ZSL methods has been applied to human action recognition. Finally, semantic representations required by any ZSL methods are briefly reviewed.

\subsection{Multi-label Learning}

In a real scenario, semantics underlying real-world data is often complex and has to be characterized with multiple labels, e.g., web videos. In a video clip pertaining to human actions, multiple actions could happen simultaneously, e.g., sitting, eating and listening. In this scenario, no episode in such a video clip can be characterized by a single action label and a set of labels hence have to be used collectively to describe this video clip. Even though a video clip can be divided into several episodes corresponding to different human actions, the segmentation and annotation process could be difficult, tedious, laborious and time-consuming. In particular, semantic image segmentation and human action detection in video streams remain unsolved in general. As a result,
multi-label learning is often formulated as a weakly supervised learning task that predicts a set of labels associated with an instance but does not address the issue in assigning each label in the set to a specific object within this instance. To tackle a weakly supervised multi-label learning problem, two different representation methods are used to characterize input data: \emph{instance-level} and \emph{object-level} representations. An instance-level representation is a global representation of an instance, e.g., a video clip or an image, without considering objects  appearing in this instance, while an object-level representation is a local representation that describes individual objects extracted from an instance, e.g., semantically meaningful episodes/patches in the video/image. Depending on the representation of input data, multi-label learning methods can be divided into two categories.

In multi-label learning, most of the existing methods \cite{zhang2016fast,guillaumin2009tagprop,wang2016cnn,nam2014large} work on an instance-based representation, a single feature vector of an instance. Recently, Fast0Tag \cite{zhang2016fast} was proposed for multi-label image tagging by learning a mapping from visual to label space. An image containing multiple objects is represented by one aggregated visual representation. Alternatively, TagProp \cite{guillaumin2009tagprop} uses an adapted nearest neighbour model for multi-label learning in visual space where each image of multiple objects is also represented by one feature vector at the instance level. \cite{wang2016cnn} use a convolutional neural network (CNN) directly working on raw images of multiple objects to learn image-level deep visual representations for multi-label classification. \cite{nam2014large} use a deep neural network with a rank loss in learning for large-scale multi-label text classification where an input document is represented with a single feature vector. Although representing one instance at the global level is straightforward and convenient, it might neglect the intrinsic relationship between multiple objects within an instance. Thus, an instance-level representation might result in a catastrophic information loss, especially for long-term dependent and complex video data.

To overcome the weakness in neglecting the information regarding the intrinsic relationship between objects within an instance, efforts have been made to exploit such information in previous works. Despite being difficult, the segmentation of multiple objects within an instance turns out to be beneficial to multi-label learning. One framework named \textit{multi-instance multi-label learning (MIML)} \cite{zhou2007multi} demonstrates that multi-label learning can be fulfilled effectively if multiple objects within an instance have been explicitly separated or segmented even if no label is explicitly assigned to each of multiple objects within an instance during learning. In real applications, however, automatic semantic segmentation of objects in an instance is also challenging, and a manual segmentation process is laborious and time-consuming. Moreover, some recent works tend to explore object-based representations without using any explicit semantic object segmentation techniques, which seeks a synergy between the MIML and object-level representations. \cite{gu2016weakly} address this weakly supervised issue in multi-label human action detection with a two-stage solution. First, a set of potential objects or spatial-temporal volumes are generated and selected from a video instance with a set of handcrafted rules. Then the problem is transformed into a MIML problem which can be solved by those traditional multi-label learning  algorithms under the MIML framework. Similar ideas were also explored by \cite{wei2016hcp} and \cite{tang2017deep} for multi-label image classification. However, the extraction of true positive objects from the original visual instance is a very challenging yet non-trivial task, which critically determines the multi-label learning performance. To extract all the meaningful objects within an instance, a lot of candidate proposals have to be considered so that it might suffer from a high computational burden. Instead of using the MIML, \cite{cabral2015matrix} attempt to explore the information regarding multiple objects in instances via a matrix completion method. Their method works on the assumption that an instance representation may be expressed by a linear combination of hidden representations of objects appearing in this instance. Experimental results reported by \cite{cabral2015matrix} demonstrate the effectiveness of this method via an instance-level bag-of-words image representation. However, this idea does not seem applicable to other kinds of visual representations, such as those popular yet powerful deep representations.

\subsection{Multi-label Zero-shot Learning}

\emph{Zero-shot learning} (ZSL) has attracted much attention in recent years and provides a promising technique for recognizing a large number of classes without the need of the training data concerning all the classes.  Very recently, \cite{zhang2016infinite} have formally shown that it is feasible to predict a collection of infinite unseen labels with a classifier learned on training data concerning only a number of labels in this collection or a subset of this collection, where multi-label ZSL is a special case in this so-called ``infinite-label learning" paradigm. According to a ZSL taxonomy    \cite{wang2016zero}, existing ZSL approaches are divided into three categories, namely, \textit{direct mapping} \cite{lampert2014attribute, fu2010manifold, akata2015evaluation, xian2016latent, yu2018transductive}, \textit{model parameter transfer} \cite{changpinyo2016synthesized,mensink2014costa} and \textit{joint latent space learning}
\cite{frome2013devise,lei2015predicting,changpinyo2016predicting,zhang2016learning,zhang2015zero,zhang2016zero,wang2016zero, yu2018zero}.  Although most existing works focus on single-label ZSL, efforts have been made to address more complex multi-label ZSL issues \cite{mensink2014costa,zhang2016fast,ren2015multi,sandouk2016multi, nam2015predicting,fu2014transductive}.

For \emph{direct mapping}, it needs to learn a mapping directly from visual to semantic space for zero-shot recognition on the semantic space, which poses a challenge to multi-label ZSL. In single-label ZSL, a training example provides a visual-semantic representation pair used to learn a one-to-one direct mapping. In multi-label ZSL, however, one instance has to be associated with a set of multiple labels and the number of labels associated with different instances are various. As a label is represented with a semantic feature vector, e.g., a vector of attributes or a word vector, in a semantic space, it is no longer straightforward to learn a direct mapping from visual to semantic space in the context of multi-label ZSL. How to model complex semantics underlying a set of labels associated with an instance becomes a central issue in multi-label ZSL. To tackle this issue, most of existing works \cite{fu2014transductive,sandouk2016multi} make use of the composition properties of semantic representations such as word vectors by using the average of semantic representations of multiple labels to a collective semantic representation for a set of labels associated with the instance. Thus, a training example is formed with a pair of an instance-level visual representation and its corresponding collective semantic representation, which enables one to learning a direct mapping for multi-label ZSL. Apparently, such a collective representation cannot avoid information loss even though a  contextualized semantic representation \cite{sandouk2016multi} was used.
In particular, the multi-label ZSL method proposed by \cite{fu2014transductive} is subject to a fundamental limitation; their method has to take into account all the possible combinations of different unseen labels in a pre-fixed unseen label collection. Thus, the computational complexity of their algorithm grows exponentially with respect to the number of unseen labels and hence can cope with only a very small number of pre-fixed unseen labels (e.g., up to eight in their experiments).
To alleviate the information loss problem in generating a collective semantic representation, Fast0Tag \cite{zhang2016fast} introduces an alternative solution to collective semantic representations. In their method, each visual instance is mapped into a ``principal direction" in the semantic space based on an assumption that there is always such a direction for any multi-labelled instances in a semantic space, e.g., word vector space, and all the labels associated with this instance always rank ahead of irrelevant labels. In other words,  a hyperplane perpendicular to this direction can always be found to separate the relevant labels from the irrelevant ones for any multi-labelled instance.  While this assumption holds for those datasets used in their zero-shot image tagging experiments \cite{zhang2016fast}, it remains unclear for other image datasets and different domains, e.g., human action recognition.
From an alternative perspective, \cite{ren2015multi} suggest using an object-level visual presentation under the direct mapping framework for multi-label zero-shot object recognition. Before multi-label learning takes place, an image thus has to be semantically segmented into meaningful subregions and each subregion can be characterized by one label. As a result, their solution is actually a special case of the MIML \cite{zhou2007multi} but heavily relies on sophisticated semantic segmentation techniques that remain unavailable up to date. Furthermore, this method is not extensible to sequential data such as video clips.

Like the works in extending \textit{direct mapping} to multi-label ZSL, the \textit{model parameter transfer} idea is also adapted for multi-label ZSL, leading to COSTA \cite{mensink2014costa}. COSTA aims to establish a model for each unseen label via a linear weighted combination of known-label models. The known-label models are trained independently by means of a one-vs-rest binary classifier, e.g., support vector machines (SVMs). The combination coefficients are determined by the co-occurrence of multiple labels derived from either annotation of datasets in hand or external web sources. In COSTA, however, the known-label models are trained independently without considering the relationship and coherence among those labels that together describe an instance. Then, COSTA only uses label co-occurrences to model the relatedness between a pair of labels but neglects the semantics of an individual label itself. So far, this idea has been tested only on static images in the context of multi-label zero-shot object recognition.

The \textit{joint latent space learning} methodology was proposed for multimedia information retrieval and multi-label related learning and led to favorable results in real-world applications \cite{weston2010large,xu2013survey,gong2014multi,karpathy2015deep}. The core idea underlying this methodology is learning a joint latent embedding from both visual and semantic domains to narrow the semantic gap so that a task can be done effectively in the latent embedding space(s). More recently, this general idea has also been explored in single-label ZSL
\cite{frome2013devise,lei2015predicting,changpinyo2016predicting,zhang2016learning,zhang2015zero,zhang2016zero,wang2016zero,yu2018zero}.
Empirical studies suggest that those \textit{joint latent space learning} methods often outperform most of existing direct mapping and model parameter transfer methods on several benchmark datasets designed for single-label ZSL \cite{frome2013devise,lei2015predicting,changpinyo2016predicting,zhang2016learning,zhang2015zero,zhang2016zero,wang2016zero,xian2017zsl}. To the best of our knowledge, however, there exists no joint latent space learning method to tackle multi-label ZSL problems probably due to a lack of techniques in modelling complex semantics underlying a set of multiple labels describing an instance, as elucidated above for direct mapping. In this paper, we propose a novel approach to multi-label zero-shot human action recognition by exploring the joint latent space learning idea to holistically tackle those challenges described in Section \ref{sect:intro}.

\begin{figure*}[t]
	\includegraphics[width=\linewidth]{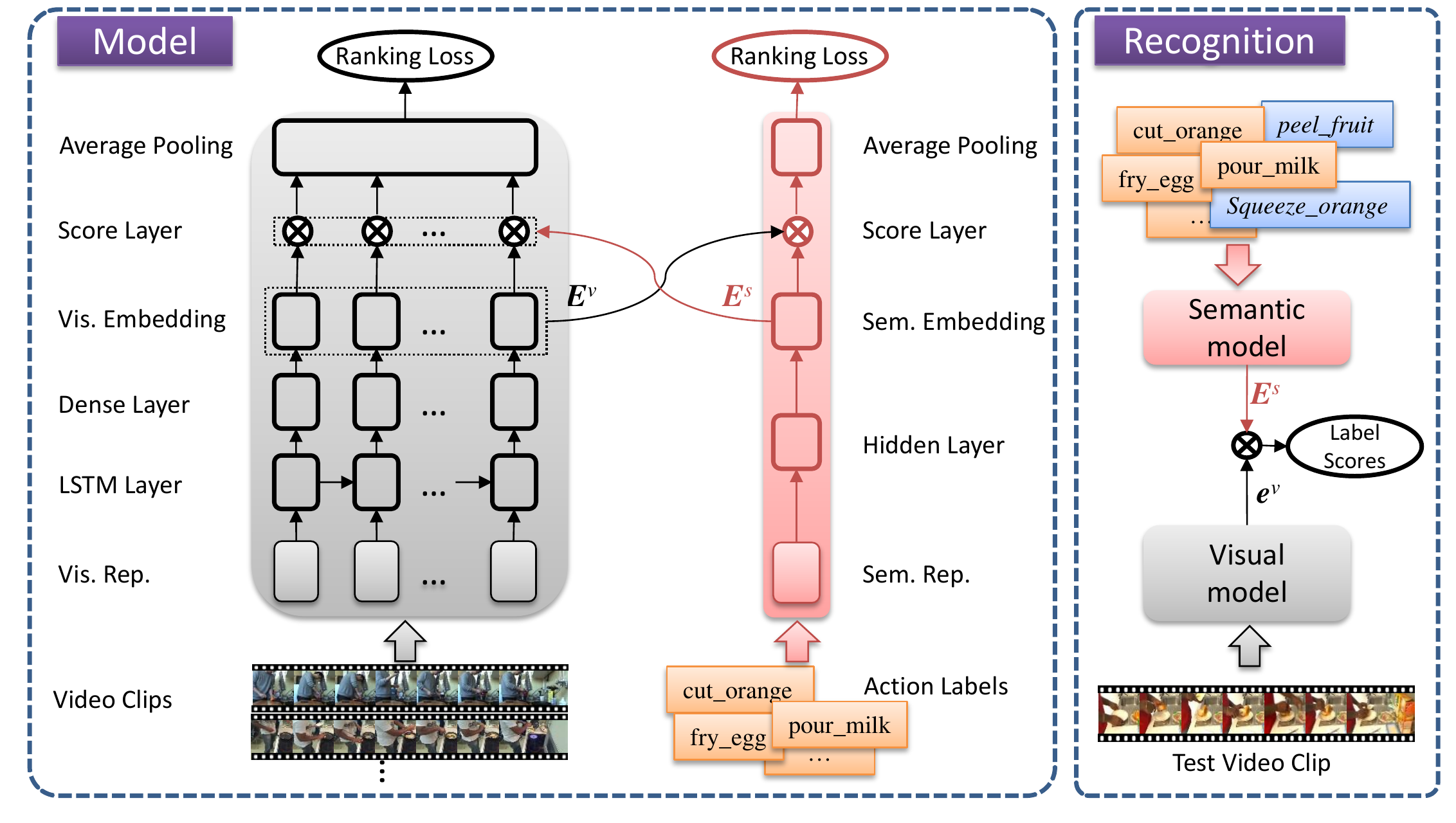}
	\caption{Our multi-label zero-shot human action recognition framework. This framework shown in the left box is composed of two component models: visual and semantic model highlighted with grey and red colors. After the joint latent ranking embedding learning, the trained visual and semantic models work together for multi-label zero-shot recognition, as shown in the right box.  Action labels marked with brown color are training classes or known labels during learning, while action labels marked with blue colour are test classes or unseen labels during learning.}
	\label{fig_framework}
\end{figure*}

\subsection{Semantic Representation}

Regardless of different ZSL scenarios, modelling semantics underlying a collection of labels and their relatedness plays a critical role in knowledge transfer required by ZSL. Miscellaneous methods in semantics modelling and representations have been developed from different perspectives including \emph{attributes of labels}, \emph{label embedding}, \emph{co-occurrence of labels} and \emph{concept embedding}.

Attributes of labels are a generic semantic representation where a label is characterized by a list of  attributes common to all the labels \cite{lampert2014attribute}.
Label embedding refers to embedding labels onto a semantic space where the semantic relatedness of labels are modeled \cite{mikolov2013distributed}. Label embedding is often carried out via learning on external textural resources. For example,  the famous \emph{Word2Vec} semantic embedding is obtained by training a skip-gram neural network on the large-scale corpora, e.g., Google News dataset \cite{mikolov2013distributed}. Such semantic representations are widely used in ZSL, e.g., \cite{zhang2016infinite,zhang2016fast,ren2015multi,fu2014transductive,wang2016zero}. Unlike label embedding obtained with external resources , co-occurrence of labels is yet another way to capture the relatedness between different labels, e.g., \cite{nam2015predicting}. Alternatively, the co-occurrence information on different class labels can also be extracted from external resources for ZSL \cite{mensink2014costa}.
In particular, co-occurrence of labels allows for capturing the relatedness between labels jointly used to describe an instance.
The label co-occurrence information may be incorporated into learning semantic embedding for a given dataset, e.g., \cite{nam2015predicting}.
In concept embedding, the semantic meaning of a label is assumed to be polysemous depending on different labels (together treated its context of this target label) jointly used to describe an instance. Hence, the semantic meaning of a label under a specific context frames a concept. As a result, concept embedding \cite{ubai2016contextualized} can be viewed as contextualized label embedding where a label may have multiple semantic representations in different contexts. The concept embedding seems specific and is only applicable to direct mapping for multi-label ZSL \cite{sandouk2016multi}. Our proposed framework for multi-label zero-shot human action action is generic so that all the semantic representations apart from the concept embedding may be used directly in our framework.

\section{Model Description}
\label{sect:model_des}

In this section, we present a novel framework for multi-label zero-shot human action recognition. First, we overview the proposed framework along with our motivation and justification. To make it self-contained, we then briefly review the LSTM unit, an important mechanism used in our framework for implicit saliency detection on video data. Next, we present the joint latent ranking embedding learning method including the rank loss functions and an alternate learning algorithm especially developed for our proposed architecture. Finally, we specify a procedure on how to apply a trained joint latent ranking embedding model to multi-label zero-shot human action recognition in test.

\subsection{Overview}
\label{sect_overview}

Our proposed framework aims at multi-label zero-shot human action recognition.
We formulate this problem as learning a mapping
$\pmb{\phi}\!\!: \pmb{x} \rightarrow \pmb{y}$, where $\pmb{x}$ is a visual input, e.g., a set of segment-level visual feature vectors extracted from a video clip, and $\pmb{y} \in \mathbb{R}^{|C|}$ is a list of label-relatedness scores for $\pmb{x}$ with respect to a action label collection, $C = \{ 1, \cdots, |C|\}$, where $C$ is further divided into two mutually exclusive label subsets, $C^{T\!r}$ and $C^U$, corresponding to known (training) and unseen actions; i.e., $ C^{T\!r} \cup C^U = C$ and $C^{T\!r} \cap C^U=\emptyset$. During learning the mapping $\pmb{\phi}$, only training examples of labels in $C^{T\!r}$ are available. However, the learned mapping $\pmb{\phi}$ is used to predict any actions appearing in a video clip no matter whether they are known actions in $C^{T\!r}$ or unseen actions in $C^U$.

To tackle the problem formulated above, we propose a joint latent ranking embedding framework. Motivated by the joint latent space learning idea used in ZSL \cite{frome2013devise,lei2015predicting,zhang2016learning,zhang2015zero,zhang2016zero,wang2016zero}, we would tackle the knowledge transfer issue in the joint latent embedding spaces where the original visual and semantic representations are mapped into. By embedding visual and semantic representations into the joint latent embedding spaces, we expect that semantic gap can be narrowed considerably and the semantic relatedness of known and unseen labels may be effectively explored and exploited in zero-shot recognition. Thus, our framework consists of two component models: \emph{visual} and \emph{semantic} models used to learn latent visual and semantic embedding, respectively. Two component models are tightly coupled to learn a joint latent ranking embedding for knowledge transfer, as illustrated in the left box of Figure \ref{fig_framework}.

For visual embedding, we encounter two major technical issues due to the nature of weakly annotated data:
a) for a visual input, it remains unknown where an episode conveying an action, and b) it remains unclear which of those action labels describing a video clip is associated with a specific video episode. Nevertheless, a video clip is an ordered sequence of frames and we could explore the temporal coherence underlying a video clip to tackle two aforementioned technical issues. Motivated by recent works in video classification and activity recognition \cite{yue2015beyond,donahue2015long,ma2017tslstm}, we employ a
\emph{long short-term memory }(LSTM) \cite{hochreiter1997long} recurrent neural network layer to capture temporal coherence underlying an action episode. Thus, the LSTM layer (c.f. Section
\ref{sect_lstm}) is first used to process a sequence of visual representations extracted from video segments. With the memorizing and forgetting mechanism of LSTM units, we expect that the LSTM layer explores the temporal structure of human actions conveyed in a video sequence; the LSTM units would memorize the input segments until parsing an episode regarding a human action is completed and then forget all the previous input segments when an episode conveying another action starts. Thus, an implicit saliency detection is carried out where no action episode boundaries are explicitly specified. For visual embedding, we further employ two fully-connected layers, dense layer of \emph{rectified linear} (ReLu) units \cite{Nair2010Relu} and visual embedding layer of linear units, to capture salient features on the temporal coherence representations yielded by the LSTM layer. While this specific visual model shown
in the left box of Figure \ref{fig_framework} is used in our experiments, its capacity can be increased by adding more hidden units and/or layers if necessary. The score and average pooling layers above the visual embedding layer are used for joint latent ranking embedding learning as presented in Section \ref{sect_jlel}. Thus, the visual model is carried out by a deep network of heterogeneous layers.

For semantic embedding, we employ a three-layer fully-connected neural network, the input layer, the hidden layer of ReLu units and the semantic embedding layer of linear units, to carry out the semantic model, as shown
in the left box of Figure \ref{fig_framework}. This learning model is capable of capturing the intricate semantic relatedness between different actions in a label collection of a moderate size, e.g. those datasets used in our experiments. If necessary, its capacity can be increased by adding more hidden units and/or layers. As a result, the neural network is fed with a specific semantic representation of action labels, e.g., word vectors and subsequently map them into the semantic embedding layer via a hidden layer. Likewise, the score and average pooling layers above the semantic embedding layer are used for joint latent ranking embedding learning. To explore the semantic relatedness between different labels in bridging the semantic gap between visual and semantic space, semantic embedding learning needs to automatically exploit the information carried in training data, e.g., frequency of label co-occurrence in a training dataset.

During the joint latent ranking embedding learning, the visual and semantic models are tightly coupled to learn a ranking criterion for the joint latent visual and semantic embedding spaces. This ranking criterion ensures that the relatedness scores of those labels associated with a visual instance are higher than those for other labels irrelevant to this instance, and the relatedness scores of those visual instances relevant to an action label are higher than those of other visual instances irrelevant to this label.
For learning, we propose an algorithm working alternately on two models for parameter estimation by promoting the correct ranking based on training examples in known classes. During training, the visual model learns the visual embedding of a video instance such that those labels relevant to this instance rank ahead of other irrelevant ones in terms of the relatedness scores estimated on the semantic latent space, $E^s$. Reciprocally, the semantic model learns the semantic embedding of action labels such that the relevant visual instances rank higher than the irrelevant ones in terms of relatedness scores calculated in the visual latent space, $E^v$. Once the learning is completed, the trained joint latent ranking embedding model can be applied to a test video clip for human action recognition. As a result, the relatedness scores corresponding to all the known and unseen action labels in a label collection are achieved by using both visual and semantic models (c.f. Section \ref{sect_multi-label ZSL}), as illustrated in the right box of Figure  \ref{fig_framework}.

\begin{figure}[t]
	\caption{LSTM unit and its information flow.}
	\label{fig_lstm}
	\centering
	\includegraphics[width=0.9\textwidth]{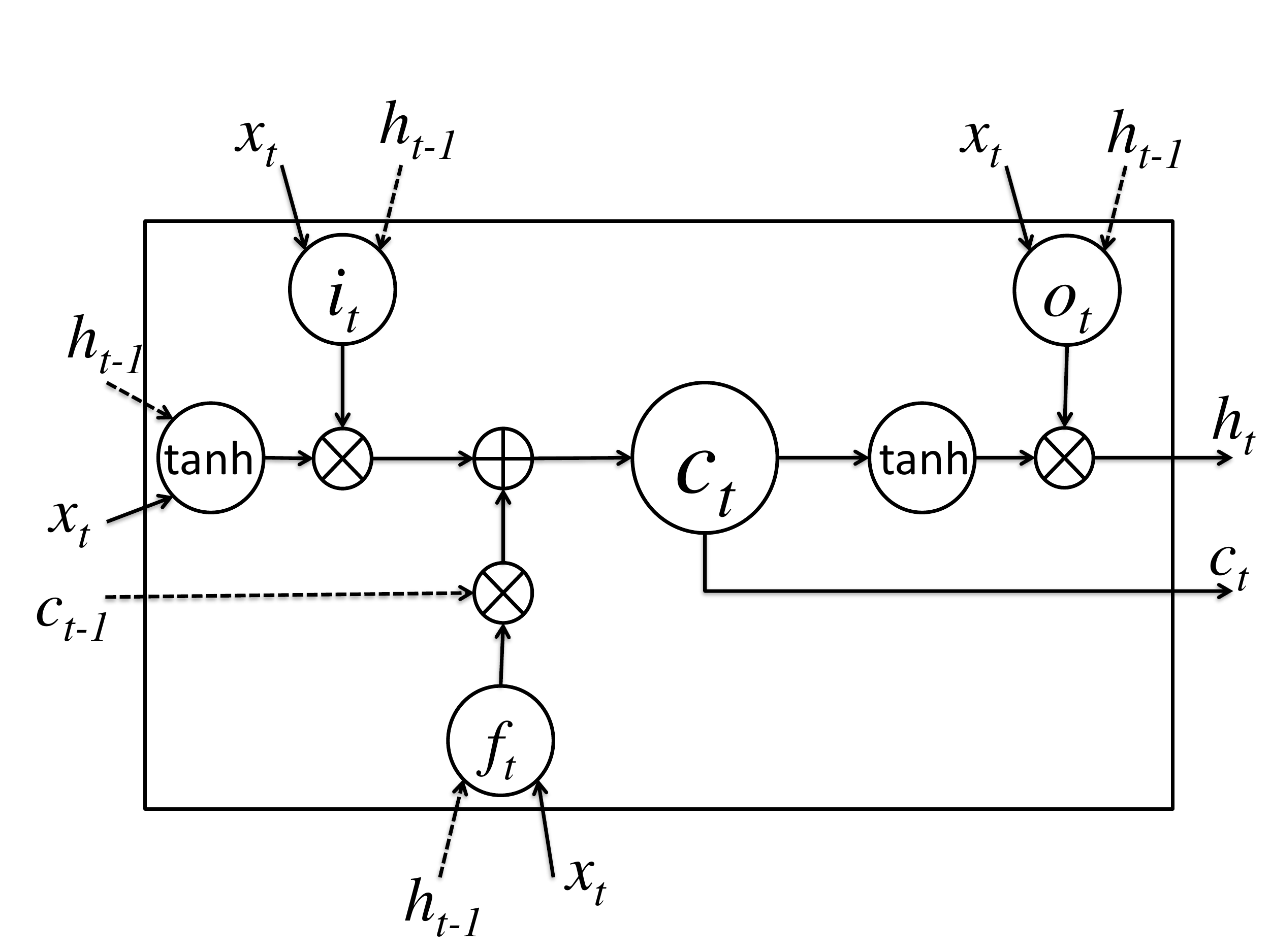}
\end{figure}

\subsection{Long Short Term Memory}
\label{sect_lstm}

As described previously, the LSTM layer plays a crucial rule in tackling two technical issues arising from weakly annotated data for visual embedding in our proposed framework. Here, we briefly review the mechanism of an LSTM unit to facilitate understanding our framework.

As shown in Figure \ref{fig_lstm}, an LSTM unit consists of a memory cell reserving the historic information at previous time steps. The output in time step $t$ is determined by the current input $x_t$ and the activation value in this memory cell. Three gates, \emph{input},\emph{output} and \emph{forgetting} gates, are used to control the information flow in the LSTM unit. The information flow in an LSTM unit \cite{hochreiter1997long, graves2013generating} is formulated as follows:

$$
i_t = \sigma(W_{xi}x_t + W_{hi}h_{t-1}+b_i),
$$
$$
f_t = \sigma(W_{xf}x_t + W_{hf}h_{t-1} + b_f),
$$
$$
c_t = f_tc_{t-1} + i_t \tanh(W_{xc}x_t + W_{hc} h_{t-1} + b_c),
$$
$$
o_t = \sigma(W_{xo}x_t + W_{ho} h_{t-1} + b_o),
$$
and
$$
h_t = o_t \tanh(c_t),
$$
where $\sigma(\cdot)$ and $\tanh(\cdot)$ are the sigmoid and the tangent hyperbolic functions, and $i$, $f$, $o$ refer to input, forget and output gates, respectively. $c$ is the cell state, representing the information stored in the cell, and $h$ is output of the LSTM unit or input receiving from the output of another LSTM unit retrospectively). $W$'s and $b$'s are weight matrices and bias vectors associated to different gates.

When LSTM units are used in a hidden layer as done in our framework, the dimension of hidden vector $h$ is determined by the number of LSTM units used in this hidden layer. Hence, it is a hyper-parameter that has to be tuned on a given dataset. The dimensions of weight matrices $W$ are thus determined by the dimensions of output vector $h$ and input $x$. The gates $i$, $f$, $o$ and the cell state $c$ are of the same dimension as that of $h$.

\begin{table}[t]
	\centering
	\caption[]{Nomenclature.}
	\label{table_notation}
	\newsavebox{\tablebox}
	\begin{lrbox}{\tablebox}
		\begin{tabularx}{\textwidth}{@{}lX}
			\toprule
			Notation & Description \\
			\toprule
			$\mathcal{D}$, $\mathcal{D}_T$ & training, test datasets \\	
			$|\cdot|$, $||\cdot||_1$ & cardinality of a set, $L_1$ norm of a vector \\
			$C$, $C^{T\!r}$, $C^U$ & label collection, training and unseen class label subsets\\	
			$E^v$, $E^s$ & visual and semantic embedding space\\				$\pmb{x}_{it}$ & visual representation for the $t$-th segment of the $i$-th example\\
			$\pmb{x}_i$ & collection of all the segment-level visual representations of the $i$-th example \\
			$\pmb{s}_c$ & semantic representation for the $c$-th label\\
			$\pmb{Y}$ & binary target label matrix of training dataset\\
			$\pmb{y}_i$ & binary target label set of the $i$-th example, i.e., the $i$-th column of matrix $\pmb{Y}$\\
			$\pmb{y}^c$ & binary indicator vector of the $c$-th label appearing examples, i.e., the $c$-th row of matrix $\pmb{Y}$\\			
			$\pmb{E}^v$, $\pmb{e}_i^v$ & visual embedding matrix and the column vector for $i$-th video clip\\
			$\pmb{E}^s$, $\pmb{e}_c^s$ & semantic embedding matrix and the column vector for the $c$-th label\\
			$d_x, d_s, d_e$ & dimensions of visual, semantic, latent embedding space\\
			$\pmb{o}_i$ & relatedness scores between the $i$-th example and all the candidate labels in visual model\\
			$\pmb{o}^c$ & relatedness scores between the $c$-th label and all the training examples in semantic model\\
			$\pmb{\phi}_v$, $\Theta_v$ & visual embedding function and parameters\\
			$\pmb{\phi}_s$, $\Theta_s$ & semantic embedding function and parameters\\
			$\pmb{\phi}$ & $\pmb{\phi}= \big \{ \pmb{\phi}_v, \pmb{\phi}_s \big \}$, mapping function for multi-label zero-shot recognition \\
			$C(\hat{\pmb{x}})$, $L(\hat{\pmb{x}})$ & the ground-truth label set of test instance  $\hat{\pmb{x}}$, the ranking list of all the labels predicted for $\hat{\pmb{x}}$ in terms of the relatedness scores  \\
			$D^c$, $L^c$ & Set of test instances of which ground-truth label sets include the $c$-th label, the ranking list of all the test instances in terms of the relatedness scores for the $c$-th label \\
			\bottomrule
		\end{tabularx}
	\end{lrbox}
	\scalebox{1}{\usebox{\tablebox}}
\end{table}

\subsection{Joint Latent Ranking Embedding Learning}
\label{sect_jlel}

Now we present the joint latent ranking embedding learning in our proposed framework.
To facilitate our presentation, we summarize the notations used in this paper in Table \ref{table_notation}.

\subsubsection{General Description}
\label{sect_jle}

Given a training set of weakly annotated video clips, $\mathcal{D}=\{\pmb{x}_i, \pmb{y}_i\}_{i=1}^{|\mathcal{D}|}$, where $\pmb{x}_i$ is the visual input  and $\pmb{y}_i \in \{+1,-1\}^{|C^{T\!r}|}$ is its binary target label vector in the $i$-th example: $+1\!/\!-1$ element indicates the presence/absence of a specific action belonging to $C^{T\!r}$ in $\pmb{x}_i$.

For a video instance $\pmb{x}_i$ in $\mathcal{D}$, we divide it evenly into $T$ segments\footnote{A segment refers to a volume of multiple consecutive frames.}, segment-level visual representations are extracted, which are collectively denoted by $\{\pmb{x}_{i1}, \pmb{x}_{i2}, \cdots, \pmb{x}_{iT}\}$. At the $t$-th time step, the segment representation $\pmb{x}_{it}$ is fed into a hidden LSTM layer and processed by this LSTM layer and two subsequent  fully-connected layers of linear activation functions (c.f. Figure \ref{fig_framework}). The latent visual embedding of the $t$-th segment,
$\pmb{e}^v_{it}$, is obtained as follows:
\begin{equation}
\label{eq_vemb}
\pmb{e}^v_{it} = \pmb{\phi}_v(\pmb{x}_{it};\Theta_v).
\end{equation}
Here, $\pmb{\phi}_v$ is the visual embedding function implemented by the parametric visual model and $\Theta_v$ is a collective notation of all the parameters in this model, including weights and biases involved in this deep network.

Likewise, as depicted in Figure \ref{fig_framework}, the $c$-th label in a label collection is first represented by a specific semantic representation, $\pmb{s}_c$, that is fed to the semantic embedding function, $\pmb{\phi}_s$, implemented by the parametric semantic model of which all the parameters are denoted by $\Theta_s$ collectively. Thus, the semantic embedding, $\pmb{e}^s_c$, of the $c$-th label is
\begin{equation}
\label{eq_semb}
\pmb{e}^s_c=\pmb{\phi}_s(\pmb{s}_c;\Theta_s).
\end{equation}

A score layer is employed in each of the visual and the semantic models. In the visual model,  the score layer takes the outputs of the visual embedding layer at all the time steps to yield the relatedness scores regarding all the labels for $\pmb{x}_i$ with a dot product between the visual embedding of each segment in  $\pmb{x}_i$ and the semantic embedding of all the labels in a label collection:
\begin{equation}
\label{eq_vscore}
\pmb{o}_{it} = <\pmb{e}_{it}^v, \pmb{E}^s>,
\end{equation}
where $\pmb{E}^s \in \mathbb{R}^{d_e\times |C^{T\!r}|}$ is a collective notation of the semantic embedding of all the labels. Here, $<\!\pmb{a},\pmb{B}\!>=\pmb{a}^T\pmb{B}$ is a vectorial notation of the dot product between a vector, $\pmb{a}$, and each column of a matrix, $\pmb{B}$. Then the relatedness scores between this video instance and different labels are achieved by averaging over the scores on all the segments of this video instance:
\begin{equation}
\label{eq_avescore}
\pmb{o}_i = \frac{1}{T} \sum_{t=1}^T \pmb{o}_{it}.
\end{equation}
Likewise, the relatedness scores between different video instances
and the $c$-th label in the label collection, $\pmb{o}^c \in \mathbb{R}^{|\mathcal{D}| \times 1}$,
is estimated in the same manner as done in the visual model based on the visual embedding of those video instances and the semantic embedding of the $c$-th label. Thus, the $i$-th element of $\pmb{o}^c$, the relatedness score regarding the $i$-th visual instance is
\begin{equation}
\label{eq_sscore}
o_i^c = \frac{1}{T} \sum_{t=1}^T <\pmb{e}_{it}^v, \pmb{e}_c^s>.
\end{equation}

For the joint latent ranking embedding learning, we need to optimize the parameters in the visual and the semantic models with training data and proper rank loss functions (the technical details are presented in Section \ref{sect_loss}). Assume that
$l_v(\cdot,\cdot)$ and $l_s(\cdot,\cdot)$ are two loss functions with respect to the visual and the semantic model, the joint latent ranking embedding learning is boiled down to simultaneously solving the following optimization problems:
\begin{equation}
\label{eq_vmodel}
\Theta_v^* = {\rm argmin}_{\Theta_v} \sum_{i=1}^{|\mathcal{D}|} l_v(\pmb{o}_i, \pmb{y}_i),
\end{equation}
\begin{equation}
\label{eq_smodel}
\Theta_s^* = {\rm argmin}_{\Theta_s} \sum_{c \in C^{T\!r}} l_s (\pmb{o}^c,\pmb{y}^c).
\end{equation}
Here, the binary indicator vector $\pmb{y}^c \in \mathbb{I}^{1 \times {|\mathcal{D}|}}$ is a row vector in the target label matrix $\pmb{Y} \in
\mathbb{I}^{|C^{T\!r}| \times {|\mathcal{D}|}}$ of a training dataset, $\mathcal{D}$, and elements of +1 in $\pmb{y}^c$ indicate that the $c$-th label appears in the target label sets of the corresponding training examples in $\mathcal{D}$. The binary indicator vector $\pmb{y}_i \in \mathbb{I}^{|C^{T\!r}| \times 1}$ is a column vector in $\pmb{Y}$, and
elements of +1 in $\pmb{y}_i$ refers to those labels in the target label set associated with the $i$-th training example in $\mathcal{D}$. The value of elements corresponding to irrelevant visual input in $\pmb{y}^c$ or labels in $\pmb{y}_i$ is always set to -1.

\subsubsection{Rank Loss Function}
\label{sect_loss}

As described in Section \ref{sect_overview}, multi-label zero-shot learning needs to establish a mapping that outputs a label-relatedness score list for a video input where the scores of the relevant labels should be ranked higher than those of irrelevant ones.
In previous studies, various rank loss functions have been developed for ranking-based learning \cite{lapin2017analysis}. To demonstrate the effectiveness of our joint latent ranking embedding framework, we adopt two simple yet typical rank loss functions, \emph{RankNet loss} \cite{burges2005learning} and the margin-based \emph{hinge rank loss} \cite{Herbrich1999mrank}, in our work although other rank loss functions \cite{lapin2017analysis} may be employed in our framework as well. RankNet loss \cite{burges2005learning} provides a generic loss function for ranking-based learning from a probabilistic perspective, while hinge rank loss \cite{Herbrich1999mrank} was originally proposed for structural SVMs and has been widely used in different tasks including single-label zero-shot learning, e.g., \cite{frome2013devise,akata2015evaluation}.

Nevertheless, we observe the following phenomenon in our experiments when using the original RankNet and hinge rank losses. By using only a ranking constraint in either of two rank losses,
all the labels are considered independently and treated equally so that the less frequently used relevant labels might be overlooked during learning. Moreover, two rank losses generally make use of pairwise constraints to explore a relationship between labels associated with an instance explicitly.
However, the relatedness scores in such rank losses are not bounded and could hence vary across different examples. Thus, some ``difficult" pairs of labels are likely to incur a larger cost that predominates the overall loss, which could make the learning biased to those pairs of labels only.
Furthermore, relatedness scores may vary in a large range for different training examples even though proper ranking relationships among them are established, which results in the poor performance. Motivated by the above observation, we introduce a regularization term to RankNet and hinge rank losses to overcome those problems.

For the target label set expressed with binary indicators, $\pmb{y}_i$, in the $i$-th training example, $(\pmb{x}_i, \pmb{y}_i)$, the elements of +1 indicate all the labels relevant to $\pmb{x}_i$ while elements of -1 express all the remaining labels irrelevant to $\pmb{x}_i$  in terms of all the known actions in $C^{T\!r}$.
Likewise, $\pmb{y}^c$, a binary indicator in $\{+1, -1\}$ regarding whether the $c$-th action appears in training examples in $\mathcal{D}$, can be handled in the same manner. Thus, the relatedness scores of $\pmb{x}_i$ to its positive and negative labels, $\pmb{o}_i$, are achieved with Eqs. (\ref{eq_semb}) and (\ref{eq_vscore}), and the relatedness scores of the $c$-th label to all the training examples, $\pmb{o}^c$, are calculated with  Eqs. (\ref{eq_vemb}) and (\ref{eq_sscore}). Based on the above quantities, we can define our regularized rank loss functions, $l_v(\pmb{o}_i, \pmb{y}_i)$ and $l_s (\pmb{o}^c,\pmb{y}^c)$.

Formally, we define the regularized RankNet loss function for visual embedding of $\pmb{x}_i$ as follows:
\begin{eqnarray}
\label{eq_vloss}
l_v(\pmb{o}_i,\pmb{y}_i)&=&\omega_i \Big(\sum_{p \in C_i^{T\!r+}} \sum_{q \in C_i^{T\!r-}} \log \big (1+\exp ( o_{iq}-o_{ip} ) \big ) + \nonumber \\
& & ~~~~~~~~\sum_{j\in C^{T\!r}} \log \big (1+\exp ( -y_{ij}o_{ij}) \big) \Big),
\end{eqnarray}
where $\omega_i = (|C_i^{T\!r+}|\cdot |C_i^{T\!r-}|+|C^{T\!r}|)^{-1}$ normalizes this per-instance regularized rank loss. Corresponding to the elements of +1 and -1 in $\pmb{y}_i$, $C_i^{T\!r+}$ and $C_i^{T\!r-}$ denote two subsets of relevant and irrelevant labels to $\pmb{x}_i$, respectively. Intuitively, minimizing the first term in Eq. (\ref{eq_vloss}) ensures that all the labels relevant to $\pmb{x}_i$ are ranked ahead of those irrelevant to $\pmb{x}_i$. The second term in Eq. (\ref{eq_vloss}) plays a regularization role; minimizing this term during learning promotes the relatedness scores by enlarging the relatedness scores to the relevant labels as well as diminishing those to the irrelevant ones simultaneously, which tackles the problems observed in our experiments.

Likewise, we define the regularized RankNet loss function for semantic embedding of label $c$ as follows:
\begin{eqnarray}
\label{eq_sloss}
l_s(\pmb{o}^c,\pmb{y}^c) &=& \omega_c \Big(\sum_{p\in \mathcal{D}^{c+}} \sum_{q \in \mathcal{D}^{c-}} \log \big ( 1+\exp ( o_q^c - o_p^c) \big ) +  \nonumber \\
& & ~~~~~~~~\sum_{j\in \mathcal{D}} \big(1+\exp(-y_j^co_j^c)\big)\Big),
\end{eqnarray}
where $\omega_c = (|\mathcal{D}^{c+}|\cdot|\mathcal{D}^{c-}|+|\mathcal{D}|)^{-1}$ normalizes the per-class regularized rank loss. $\mathcal{D}^{c+}$ and $\mathcal{D}^{c-}$ are the positive and the negative training example subsets, respectively, regarding the $c$-th label. With the same treatment as used in Eq. (\ref{eq_vloss}), minimizing Eq.(\ref{eq_sloss}) ensures that the video instances conveying the action of the $c$-th label are ranked above all those without this action. Moreover, those video instances conveying the action of the $c$-th label, indicated by $y_j^c=+1$, tend to have as high relatedness scores as possible while all other video instances without this action, indicated by $y_j^c=-1$, tend to have as low relatedness scores as possible.

Similarly, we define a regularized hinge rank loss function for visual embedding of $\pmb{x}_i$ as follows:
\begin{eqnarray}
\label{hinge_vloss}
l_v(\pmb{o}_i,\pmb{y}_i)&=& \omega_i \Big (\sum_{p \in C_i^{T\!r+}} \sum_{q \in C_i^{T\!r-}} \max \big (0, m - o_{ip} + o_{iq} \big ) +  \nonumber \\
& & ~~~~~~~~\sum_{j\in C^{T\!r}} \max \big(0, m-y_{ij}o_{ij}\big )\Big),
\end{eqnarray}
where $\omega_i = (|C_i^{T\!r+}|\cdot |C_i^{T\!r-}|+|C^{T\!r}|)^{-1}$ and $m$ is a pre-specified margin. Thus,
minimizing the first term in Eq. (\ref{hinge_vloss}) ensures that all the labels relevant to $\pmb{x}_i$ are ranked ahead of those irrelevant to $\pmb{x}_i$ with a pre-specified margin, $m$. The second term in Eq. (\ref{hinge_vloss}) plays a regularization role; minimizing this term during learning promotes the margin-based relatedness scores by enlarging the relatedness scores to the relevant labels as well as diminishing those to the irrelevant ones simultaneously.

Likewise, we define a regularized hinge rank loss function for semantic embedding of label $c$ as follows:
\begin{eqnarray}
\label{hinge_sloss}
l_s(\pmb{o}^c,\pmb{y}^c) &=&
\omega_c \Big(\sum_{p\in \mathcal{D}^{c+}} \sum_{q \in \mathcal{D}^{c-}}
\max \big (0, m - o_p^c + o_q^c \big ) +  \nonumber \\
& & ~~~~~~~~\sum_{j\in \mathcal{D}} \max \big(0, m-y_j^c o_j^c\big)\Big),
\end{eqnarray}
where $\omega_c = (|\mathcal{D}^{c+}|\cdot|\mathcal{D}^{c-}|+|\mathcal{D}|)^{-1}$ and $m$ is a pre-specified margin.
With the same treatment as used in Eq. (\ref{hinge_vloss}), minimizing Eq.(\ref{hinge_sloss}) ensures that the video instances conveying the action of the $c$-th label are ranked above all those without this action with a margin, $m$. Moreover, those video instances conveying the action of the $c$-th label, indicated by $y_j^c=+1$, tend to have as high relatedness scores as possible while all other video instances without this action, indicated by $y_j^c=-1$, tend to have as low relatedness scores as possible.

As a result, we can employ either our regularized RankNet loss functions in Eqs. (\ref{eq_vloss}) and (\ref{eq_sloss}) or the regularized hinge rank loss functions in Eqs. (\ref{hinge_vloss}) and (\ref{hinge_sloss}) to train visual and semantic embedding models in our framework.

\subsubsection{Learning Algorithm}
\label{sect_opt}

As formulated in Eqs. (\ref{eq_vmodel}) and (\ref{eq_smodel}), learning is going to find the optimal parameters, ${\Theta}_v^*$ and $\Theta_s^*$, by minimizing two loss functions, $l_v(\pmb{o}_i, \pmb{y}_i)$ and $l_s (\pmb{o}^c,\pmb{y}^c)$, defined in Section \ref{sect_loss}.
However, the relatedness scores required in $l_v(\pmb{o}_i, \pmb{y}_i)$ regarding the visual model involve the output of the semantic model, $E^s$, and vice versa (c.f. Figure \ref{fig_framework}).
Moreover, $l_v(\pmb{o}_i, \pmb{y}_i)$ requires the relatedness scores between all the candidate labels and each of training examples, while
$l_s (\pmb{o}^c,\pmb{y}^c)$ needs the relatedness scores between all the training examples and each of all the action labels in $C^{T\!r}$. Thus,
our optimization problems are very complex and unsolvable simultaneously with commonly used local search methods, e.g., gradient-descent based methods.

Motivated by the works dealing with similar optimization problems, e.g., \cite{kavukcuoglu2010fast,jiang2017exploiting}, we come up with a learning algorithm to train the visual and the semantic models alternately during learning. In our alternate learning strategy, our learning algorithm begins with randomly initializing the parameters in the semantic model and then use the initialized parameter to generate the initial semantic embedding.  By using the initial semantic embedding in $l_v(\pmb{o}_i, \pmb{y}_i)$, the visual model can be trained with a local search method such as the mini-batch stochastic gradient decent method. After one epoch, the current parameters in the visual model are frozen and used to generate the visual embedding for all the examples.  By using the current visual embedding in $l_s (\pmb{o}^c,\pmb{y}^c)$, the semantic model is trained in the same manner. This alternate learning process carries on until a stopping condition is satisfied. The details of this \emph{alternate} learning algorithm is described in Algorithm \ref{alg_alternating}.

\begin{algorithm}[t]
	\caption{Joint Latent Ranking Embedding Learning}
	\label{alg_alternating}
	\renewcommand{\algorithmicrequire}{\textbf{Input:}}
	\renewcommand{\algorithmicensure}{\textbf{Output:}}
	\begin{algorithmic}[1]
		\REQUIRE Randomly initialize parameters, $\Theta_v^0$ and $\Theta_s^0$, in the visual and the semantic models, respectively;  extract the visual representations of training example, $\pmb{x}_i,~ i=1,\cdots,N$, and the semantic representations of all the training labels, $\pmb{s}_c, ~\forall c \in C^{T\!r}$; input the target label matrix of the training set, $Y$; pre-set the dimensionality of joint latent ranking embedding space, $d_e$.
		\ENSURE Optimal model parameters: $\Theta_v^*$ and $\Theta_s^*$.
		\STATE Generate the initial semantic embedding $\pmb{\phi}_s(\pmb{s}_c;\Theta_s^0), ~\forall c \in C^{T\!r}; ~~t \leftarrow 0$.
		\REPEAT
		\STATE $t  \leftarrow t + 1$;
		\STATE $\Theta_v^t = {\rm argmin}_{\Theta_v} \sum_{i=1}^N l_v(\pmb{o}_i, \pmb{y}_i)$ with the current semantic embedding for one epoch;
		\STATE Generate the visual embedding with the current visual model, $\pmb{\phi}_v(\pmb{x}_i;\Theta_v^t),~i=1,\cdots,N$;
		\STATE $\Theta_s^t = {\rm argmin}_{\Theta_s} \sum_{c \in C^{T\!r}} l_s (\pmb{o}^c,\pmb{y}^c)$  with the current visual embedding for one epoch;
		\STATE Generate the semantic embedding with the current semantic model $\pmb{\phi}_s(\pmb{s}_c;\Theta_s^t), ~\forall c \in C^{T\!r}$;
		\UNTIL Stopping condition is met.
		\STATE  $\Theta_v^* \leftarrow \Theta_v^t$ and $\Theta_s^* \leftarrow \Theta_s^t$.
	\end{algorithmic}
\end{algorithm}

It is worth stating that two rank loss functions defined for visual and semantic model are related and the optimisation of one model would naturally promote the other towards its optimal solution. Hence, our alternate learning algorithm may converge after running finite epochs with the same properties held for similar methods \cite{kavukcuoglu2010fast,jiang2017exploiting}.

\subsection{Multi-Label Zero-Shot Recognition}
\label{sect_multi-label ZSL}

Once the joint latent ranking embedding learning is completed, we obtain a mapping function:
$\pmb{\phi}(\pmb{x}, \pmb{c})=\big \{ \pmb{\phi}_v(\pmb{x}|\Theta_v^*), \pmb{\phi}_s(\pmb{c}|\Theta_s^*) \big \}$ where $\pmb{\phi}_v(\pmb{x}|\Theta_v^*)$ and $\pmb{\phi}_s(\pmb{c}|\Theta_s^*)$ are the visual and the semantic embedding functions implemented by the trained visual and semantic models, respectively. Then, we can use this mapping function for multi-label zero-shot human action recognition.

For recognition, we first extract the semantic representations of all the labels, including both \emph{known} and \emph{unseen} labels, in a considered label collection: $\pmb{s}_c, ~\forall c \in C$; $C=C^{T\!r} \cup C^U$ and $C^{T\!r} \cap C^U =\emptyset$. By using the semantic embedding function, we achieve the sematic embedding of all the labels:
$\hat{\pmb{e}}_c^{s}=\pmb{\phi}_s(\pmb{s}_c|\Theta_s^*),~\forall c \in C$.
For a test video clip, we divide it into $T$ segments and extract its segment-level representations collectively denoted by $\hat{\pmb{x}} = \{\hat{\pmb{x}}_1,\hat{\pmb{x}}_2, ..., \hat{\pmb{x}}_T\}$. Technical details for extracting semantic and visual representations can be found in Section \ref{sect_representations}.
By feeding $\hat{\pmb{x}}$ to the visual embedding function, we achieve its visual embedding:
$\hat{\pmb{e}}^{v}=\pmb{\phi}_v(\hat{\pmb{x}}|\Theta_v^*)$. Thus, the relatedness scores between this test video clip, $\hat{\pmb{x}}$, and all the actions in the considered label collection $C$, including known and unseen labels during learning, is achieved by
\begin{equation}
\label{eq:recognition}
{\rm S} (\hat{\pmb{x}}, c) = <\hat{\pmb{e}}^{v}, \hat{\pmb{e}}_c^{s}>, ~~~\forall c \in C.
\end{equation}
Finally, we achieve a ranking action label list, $L(\hat{\pmb{x}})$, for this test video clip by sorting its relatedness scores measured against all the labels in $C$ with Eq. (\ref{eq:recognition}):
\begin{equation}
\label{eq:rank_x}
L(\hat{\pmb{x}})=\big \{  c_i \big \}_{i=1}^{|C|},
\end{equation}
where $\forall c_i, c_j \in C$, ${\rm Score} (\hat{\pmb{x}}, c_i) \geq {\rm Score} (\hat{\pmb{x}}, c_j)$ if $i < j$.

In our experiments regarding the use of two different rank losses in our framework, we observe that the regularized RankNet and hinge rank losses often behave differently in several evaluation scenarios (c.f. Section \ref{sect:ex_setting}). Although two rank losses generally yield the comparable performance overall, a closer look suggests that those correctly recognized video instances are quite different when two different rank losses are used in our framework, respectively. Hence, we would employ a simple fusion method to exploit the complementary aspect resulted from the use of two different rank losses.
In order to fuse the results yielded by the models trained with two different rank losses, we first normalize the relatedness scores of $\hat{\pmb{x}}$ to the considered label collection, $C$, generated by each of two models as follows:
\begin{equation}
\label{score_norm}
\tilde{{\rm S}} (\hat{\pmb{x}}, c) = \frac {{\rm S}(\hat{\pmb{x}}, c) - {\rm S}_{min}} {{\rm S}_{max} - {\rm S}_{min}},
\end{equation}
where ${\rm S}_{max}$ and ${\rm S}_{min}$ are the highest and lowest related scores of video clips, respectively, measured on a test set. Let $\tilde{{\rm S}}^{(Reg)} (\hat{\pmb{x}}, c)$ and $\tilde{{\rm S}}^{(Hinge)} (\hat{\pmb{x}}, c)$ denote the normalized relatedness scores yielded by two models trained with our regularized rank and the hinge rank losses, respectively. Then, the fused relatedness scores, $\tilde{{\rm S}}^{(Fusion)} (\hat{\pmb{x}}, c)$ is simply an average between $\tilde{{\rm S}}^{(RankNet)} (\hat{\pmb{x}}, c)$ and $\tilde{{\rm S}}^{(Hinge)} (\hat{\pmb{x}}, c)$; i.e.,
\begin{equation}
\label{score_fusion}
\tilde{{\rm S}}^{(Fusion)} (\hat{\pmb{x}}, c) =
\frac { \tilde{{\rm S}}^{RankNet} (\hat{\pmb{x}}, c)~ +~ \tilde{{\rm S}}^{Hinge} (\hat{\pmb{x}}, c) } 2.
\end{equation}
Based on the fused relatedness scores, a ranking action label list, $L^{(Fusion)}(\hat{\pmb{x}})$, is achieved in the same manner as specified in Eq.(\ref{eq:rank_x}) for any test video clip, $\hat{\pmb{x}}$.

\section{Experimental Setting}
\label{sect:ex_setting}

In this section, we describe our experimental design and settings, including datasets, visual and semantic representations, model learning, evaluation scenarios and criteria used in our experiments. Moreover, we design a number of comparative experiments to exhibit the gain resulting from different components in our framework and to demonstrate the effectiveness of our framework by a comparison to several state-of-the-art multi-label ZSL methods that could be applied to general human action recognition.

\subsection{Datasets and Splits}
\label{sub:data_split}

We first describe datasets and their split settings used in our experiments for simulation of multi-label ZSL scenarios.

\subsubsection{Datasets}
To evaluate our framework, we employ two publicly available video datasets: \textit{Breakfast} \cite{kuehne2014language} and \textit{Charades} \cite{sigurdsson2016hollywood}, in our experiments. In both datasets, at least two actions are involved in each video clip and the duration of each video clip is relatively long, which implies the temporal coherence information may be explored and exploited in human action recognition. Hence, both datasets are suitable to evaluate weakly annotated multi-label human action recognition. Below, we summarize the main aspects of two video datasets.

\noindent
\textbf{Breakfast}:
In this dataset \cite{kuehne2014language}, there are 1,989 video clips totally, where a video clip conveys several cooking actions. Totally, there are 49 cooking actions (excluding the ``silence" label), such as `stirring", ``pouring\_milk" and ``opening\_the\_fridge". Those actions are performed by 52 people in different kitchens. Although this dataset is not collected especially for multi-label human action recognition, we would use it as a proof-of-concept test bed.

\noindent
\textbf{Charades}:
This dataset \cite{sigurdsson2016hollywood} is collected from hundreds of people recording videos in their own home especially for video-based human activity analysis in daily lives. Hence, it is very challenging for multi-label human action recognition. In this dataset,
there are 9,848 video clips involving 157 different human actions totally, acting out casual everyday activities. An average duration of video clips is around 30 seconds and an average number of actions involved in a video clip is 6.8. Those actions are performed by  267  people  from  three  continents,  and more  than  one  person appear in over  15\% of  all the video clips.  The raw video data (scaled to 480p) are used  in our experiments, which are  available from the Charades project page\footnote{\url{http://allenai.org/plato/charades/}}.

\subsubsection{Dataset Splits}
\label{sect_datasplit}

\begin{figure}
	\includegraphics[width = 0.4\paperwidth]{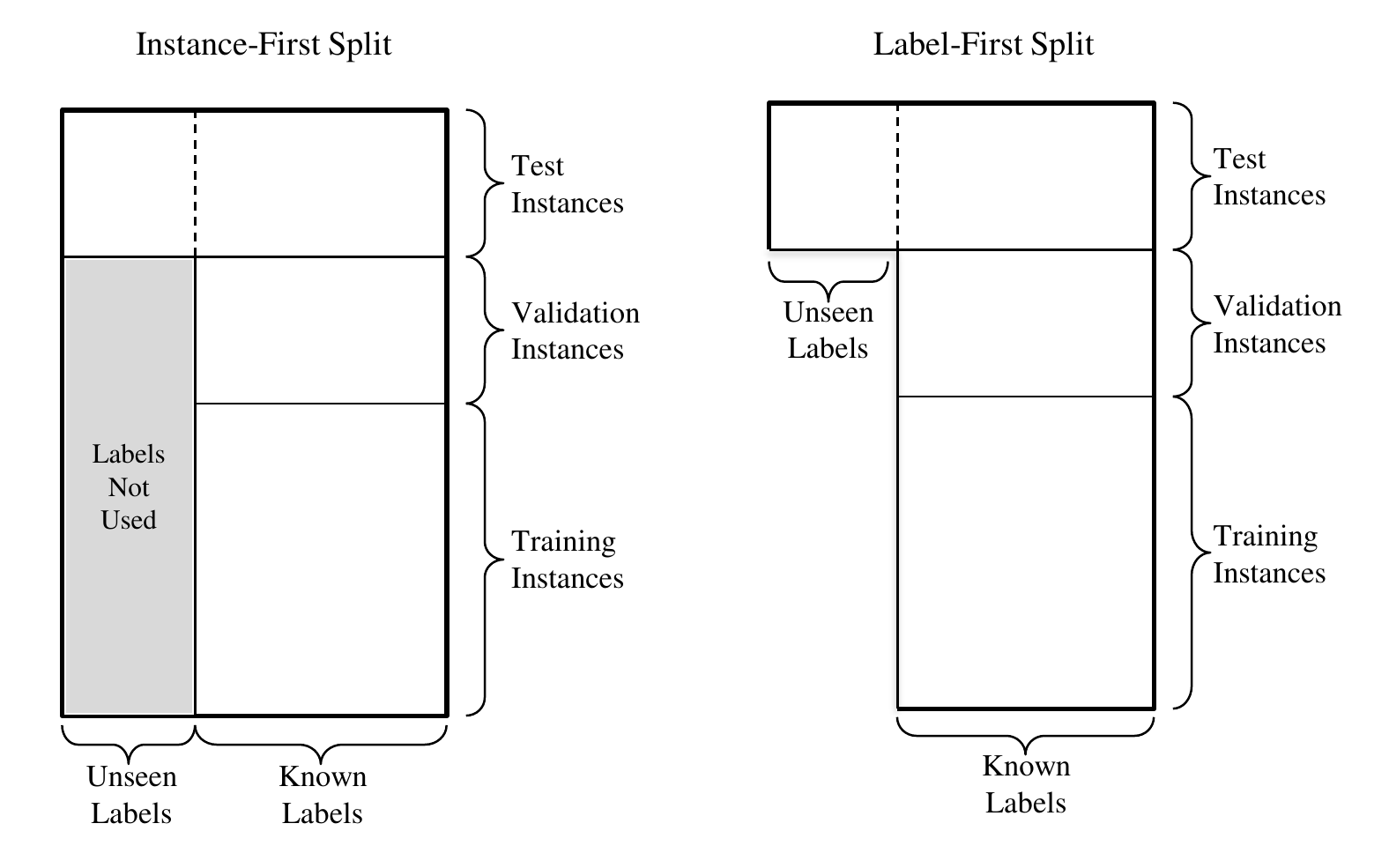}
	\caption{Two different data split settings used in our experiments. In the \emph{instance-first} split (left plot), a data set is simply split into three mutually exclusive subsets: training, validation and test subsets. Any unseen labels associated with instances  in training and validation subsets (shaded area) are removed from their target label sets before being used in learning. In the \emph{label-first} split (right plot), a number of labels are first specified as unseen labels. Instances associated with any of unseen labels form a test subset. The remaining data of known labels are further divided into training and validation subsets to be used in learning.}
	\label{fig_datasplit}
\end{figure}

To simulate a zero-shot scenario, we need to split a dataset into training and test sets where a training set contains examples associated with only known classes while a test set has test instances involving at least one unseen class. Unlike single-label ZSL where a dataset is automatically split into training and test sets once unseen classes are specified, the dataset split issue in multi-label ZSL becomes much more complicated.
In our experiments, we make two different split settings,  \emph{instance-first split} (IFS) and \emph{label-first split} (LFS), as illustrated in Figure \ref{fig_datasplit}.

\begin{table*}[ht]
	
	{\normalsize
		
		\centering
		
		\caption[]{Information on two different data split settings.}
		
		\label{table_datasplit}
		
		
		\begin{lrbox}{\tablebox}
			
			\begin{tabular}{c|c||c|c|c||c|c}\toprule \hline
				
				Dataset & Split Method & \# Training Inst. & \# Validation Inst. & \# Test Inst. & \# Known Labels & \# Unseen Labels\\
				
				\hline
				
				\multirow{2}{*}{Breakfast} &  Instance-first & 1196 & 126 & 667 & 39 & 10 \\
				
				\cline{2-7}
				
				& Label-first & 1019/917/823 & 200  & 770/872/966 & 39  & 10\\
				
				\hline
				
				\multirow{2}{*}{Charades} &  Instance-first & 6385 & 1600 & 1863 & 117 & 40 \\
				
				\cline{2-7}
				
				& Label-first & 4580/4176/3987 & 1000  & 4268/4672/4861 & 137  & 20\\
				
				\hline
				
				\bottomrule
				
			\end{tabular}
			
		\end{lrbox}
		
		\scalebox{0.77}{\usebox{\tablebox}}
		
	}
	
\end{table*}

\noindent
\textbf{Instance-First Split}\\
This is a commonly used data split setting in all the existing multi-label ZSL works, e.g., \cite{mensink2014costa,nam2015predicting,zhang2016fast}.
In this setting, instances in a dataset is first split into training, validation and test subsets. The training and the validation subsets are used for parameter estimation and hyper-parameter tuning, the dimension of latent embedding space $d_e$ and the number of iterations of Algorithm \ref{alg_alternating} for our model. The test set that may or may not involve unseen labels is reserved for performance evaluation. Then, we divide the action label collection into mutually exclusive known and unseen label sets. Before learning, any unseen labels in the target label set of an instance in the training and the validation subsets are removed as shown in the left plot of Figure \ref{fig_datasplit}. In other words, only known labels in the target label sets of an instance in those two datasets are used in learning. It is worth clarifying that unlike single-label ZSL, it is often infeasible to simulate a ZSL scenario by manipulating the validation set due to insufficient data in two datasets used in our experiments. Hence, the validation for hyper-parameter tuning in our experiments has to follow the typical protocol used in multi-label learning \cite{zhang2014multi-label}.

To split the Breakfast dataset with this setting, we adopt the pre-split by data collectors \cite{kuehne2014language}, where the video clips of 13 people are reserved for test.  We further divide the rest video clips for training and validation:  video clips of 32 people for training and the remaining video clips of seven people for validation. As a result, the numbers of video clips for training, validation and test are 1,196, 126 and 667, respectively.
Then we randomly split the 49 labels into known and unseen labels:  10 labels reserved as unseen labels and the rest 39 as known labels.

For the Charades dataset, we also adopt its pre-split provided by data collectors  \cite{sigurdsson2016hollywood}, where 7,985  and 1,863 video clips are used for training and test, respectively.  We further divide training data into two subsets: 6,385 for training and 1,600 for validation in our experiments. Then we randomly choose 40 out of 157 human actions as unseen classes and the rest 117 human actions are hence known actions. \\

\noindent
\textbf{Label-First Split}\\
Although the instance-first data split setting is widely used in multi-label ZSL, it suffers from a fundamental limitation. It is well known that multiple labels together could frame a specific concept and removing any label from this label cohort may lead to a less accurate semantic meaning and biases in learning. Furthermore, the instance-first split allows for accessing to visual features of instances involved in unseen actions. To overcome this limitation, we propose a novel data split setting for multi-label ZSL named \emph{label-first} split. In this new setting, all the labels in a label collection used in a dataset is first divided into two mutually exclusive subsets: known and unseen labels. Then, all the instances having any unseen
labels are reserved for test and the rest instances of known labels only are further divided into two subsets for training and validation, as shown in the right plot of Figure \ref{fig_datasplit}. Due to sparsity of training data, the validation in the label-first split also adopts the protocol used in multi-label learning \cite{zhang2014multi-label}.

To split the Breakfast dataset with this setting, we randomly choose 10 labels for unseen labels and the rest 39 labels are designated as  known labels accordingly. Hence, this dataset is naturally split into two sets for training and test. The training data are further divided for training and validation. For validation, we randomly choose 200 instances from the training data. Likewise, the Charades dataset is split by using 20 randomly chosen label for unseen labels. Thus, the remaining 137 labels become known labels. From the instances of known labels, we randomly choose 1,000 instance used for validation. It is worth stating that the current datasets do not allows for reserving a large number of classes as unseen classes in either the IFS or the LFS setting. In the IFS, the more labels reserved as unseen labels, the less accurate mapping learned from visual to semantic domains due to the existence of visual features of unseen actions and a lack of the corresponding action labels in such training examples. In the LFS, the more labels reserved as unseen labels, the fewer training examples available. Hence, the training examples do not convey the essential information required in learning.

For reliability, we repeat our experiments on each dataset under each split setting for three trials. During a trial,  training data given in the pre-split of each dataset is randomly divided into training and validation subsets with the instance-first split setting, and a known/unseen label split on each dataset is chosen randomly with the label-first split setting. For clarity, we summarize the data split information\footnote{All the data splits and source code used in our experiments are available on our project website: \url{http://staff.cs.manchester.ac.uk/~kechen/MLZSHAR}.} on two datasets in Table \ref{table_datasplit}.

\subsection{Visual and Semantic Representations}
\label{sect_representations}

In our experiments, we use visual representations extracted with the existing C3D deep network \cite{tran2015learning} and word vectors as semantic representations \cite{mikolov2013distributed}.

As suggested by \cite{tran2015learning}, the C3D features are extracted for a segment of 16 frames with eight frames overlapping between two adjacent segments. Thus, a training/test video clip is always divided into $T$ segments with the treatment as follows.
To ensure that each video clip can be divided into $T$ segments, any video clip must have $8*(T+1)$ frames. To this end, we simply down-sample those video clips of more frames with a proper sampling rate so that $T$ C3D feature vectors can be extracted and collectively form a \emph{segment-based} visual representation for this video clip. When a video clip has fewer frames, we first extract C3D feature vectors from those frames in this video clip and then pad all-zero vectors to the visual representation until there are $T$ feature vectors. Also,
we can convert $T$ feature vectors into a holistic \emph{instance-level} visual representation via averaging those $T$ feature vectors. By using such an instance-level visual representation in our comparative study, we would demonstrate a performance gain benefiting from exploring/exploiting temporal coherence information underlying segments in a video clip.
In our experiments, the segment-based visual representation is always used in our model while the instance-level visual representation is used in the baseline and the state-or-the-art models (c.f. Section \ref{comparative}) unless a different setting is specified.
Based on a cross-validation experiment, we choose $T=300$ for Breakfast and $T=20$ for Charades.
Although only C3D features are used in our experiments, it is worth mentioning that other kinds of visual representations, e.g., IDT features \cite{wang2013action} and deep image features extracted on a frame basis, can also be used straightforwardly in our framework.

In our experiments, we adopt Word2Vec as our semantic representation. Word2Vec was trained with a skip-gram neural network on the Google News dataset of 100 billion words \cite{mikolov2013distributed}.
As a result, one action label is represented by a 300-dimensional word vector. Although only 300-dimensional word vectors are used in our experiments, word vectors of different dimensionality may be used, and moreover, other semantic representations, e.g., attributes, can be used in our framework without any difficulty if available.

It is worth emphasizing that the same treatment described above is applied in both the learning and the recognition phases to extract visual and semantic representations.

\subsection{Model Learning}
\label{sub:model_learn}

In our experiments, model learning is implemented on Keras \cite{chollet2015}, a high-level neural networks library, running on top of either TensorFlow or Theano. As we use two neural networks to carry out the visual and the semantic models (c.f. the left box in Figure \ref{fig_framework}), we need to decide the specific network architectures and relevant hyper-parameters on two datasets during the model learning. The optimal hyper-parameters  are found by a grid-based search via a cross-validation procedure. The Adam \cite{kingma2014adam}, a stochastic optimisation method, is used for training our model with its default configuration.

The visual model takes a sequence of segment-level C3D representations of $d_x = 4,096$ features as input to the LSTM layer where there are $N_1$ LSTM units. To improve the generalization, we also apply the \emph{dropout} procedure \cite{srivastava2014dropout} to the LSTM layer where a dropout rate needs specifying. The output of LSTM units are fed to a fully connected dense layer of $N_2$ neurons, and the output of this dense layer are further fed to the visual embedding layer of $d_e$ neurons. During learning, there are no hyper-parameters involved in the score and the average pooling layer in the visual model.

As described in Section \ref{sect_overview}, the semantic model is carried out by a fully-connected three-layer feed-forward neural network. The word vectors of $d_s=300$ dimensions are first input to a hidden layer of $N_1$ \textit{ReLu} units. Subsequently, the output of this hidden layer are fed to the semantic embedding layer of $d_e$ linear units.
Note that for joint latent ranking embedding learning, the dimension of the semantic embedding space is set to the same of the visual embedding space in our experiments. Likewise, there are no hyper-parameters involved in the score and the average pooling layer in the semantic model during learning.

\subsection{Evaluation Scenarios}
\label{sect_evaluation_scenarios}

Multi-label zero-shot recognition is complex given the fact that a test instance may be associated with a label set including both known and unseen class labels. Thus, there are different evaluation scenarios in previous works \cite{sandouk2016multi, zhang2016fast}; each focuses on a specific aspect. Following their settings  \cite{sandouk2016multi, zhang2016fast}, we evaluate our framework along with other learning models used in our comparative study described later on in this section in three different scenarios: \\
\noindent
\textbf{Known-action only}: In this setting, the performance is evaluated regarding only known (training) actions. This scenario boils down to the conventional supervised multi-label learning. In this circumstance, we no longer take
any unseen action labels into account during test; for a test instance, its relatedness score ranking list contains only those regarding known action labels in $C^{T\!r}$ and any unseen action label in $C^U$ in its ground-truth label set, if there is, will be removed such that the modified ground-truth set includes only known action labels in $C^{T\!r}$.\\
\noindent
\textbf{Unseen-action only}: In this setting, the performance is evaluated regarding only unseen (test) actions. This scenario boils down to a standard ZSL setting. In this situation, we no longer consider any known action labels; for a test instance, its relatedness score ranking list contains only those regarding unseen action labels in $C^U$ and any known action label in $C^{T\!r}$  in its ground-truth label set, if there is, will be removed such that the modified ground-truth set includes only unseen  action labels in $C^U$.\\
\noindent
\textbf{Generalized ZSL}: In this setting, the performance is evaluated regarding all the actions of which labels appearing in a label collection $C$ without considering if an action label is known or unseen during learning. This scenario has been named \emph{generalized ZSL} in the machine learning community. In this situation, both known and unseen action labels are treated equally;  for a test instance, its relatedness score ranking list contains those regarding all the action labels in $C$ and the evaluation is made against its ground-truth label set that could be a mixture of known and unseen labels. It is worth highlighting that the generalized ZSL setting is required by multi-label zero-shot human action recognition in a real application.

\subsection{Evaluation Metrics}
\label{sect_evaluation}

There are a variety of evaluation metrics for multi-label learning. Depending on the output of a multi-label learning system, the evaluation metrics are generally divided into two types: \emph{ranking-based} and \emph{bipartition-based} metrics \cite{nam2015predicting, sorower2010literature}. Ranking-based metrics work for the situation that a learning system yields a ranking list of continuous-valued relatedness scores on all the candidate labels. In contrast,  bipartition-based metrics are used when a learning system produces only a binary indicator vector for all the candidate labels, where 1/0 element expresses the presence/absence. Since our model yields a ranking list of continuous-valued relatedness scores, we employ two commonly used ranking-based metrics for performance evaluation
\cite{nam2015predicting,sorower2010literature,mensink2014costa,zhang2016fast,li2016socializing}, \emph{I}nstance-centric \emph{M}ean \emph{A}verage \emph{P}recision (I-MAP) and   \emph{L}abel-centric \emph{M}ean \emph{A}verage \emph{P}recision (L-MAP). In addition, we employ other metrics, \emph{precision}, \emph{recall} and $F_1$ score, which have also been used in the performance evaluation of multi-label learning \cite{zhang2014multi-label,gong2013deep}.

To facilitate our presentation, we first define the \emph{precision-at-k} \cite{manning2009IR} in a generic form:
\begin{equation}
\label{Eq:P@k}
P @ k(A, B) =
\frac 1 k \big | A \cap B[1,\cdots,k] \big |,
\end{equation}
where $A$ is a ground-truth set, $B$ is a set of all the  retrieved entities ranked in terms of relevance, and $B[1,\cdots,k]$ indicates top $k$ entities in $B$.
Given a test dataset, $\mathbb{D}_T=\big \{ \hat{\pmb{x}}_i \big \}_{i=1}^{|\mathbb{D}_T|}$, a learning model yields a \emph{label-based} ranking list for a test instance, $\hat{\pmb{x}}_i \in \mathbb{D}_T$,  in terms of its relatedness scores to all the labels in $C$ (c.f. Eqs. (\ref{eq:recognition}) and (\ref{eq:rank_x})):
$L(\hat{\pmb{x}}_i) = \big \{ c_j \big \}_{j=1}^{|C|}$, where $\forall c_{p}, c_{q} \in C$, ${\rm Score} (\hat{\pmb{x}}_i,c_{p}) \geq {\rm Score} (\hat{\pmb{x}}_i, c_{q})$ if $p < q$. Let $C(\hat{\pmb{x}}_i)$ denote the ground-truth label set of $\hat{\pmb{x}}_i$. I-MAP over a test dataset $\mathbb{D}_T$ is defined by
\begin{equation}
\label{eq:I-MAP}
{\rm I\!\!-\!\!MAP} = \frac 1 {|\mathbb{D}_T|} \sum_{i=1}^ {|\mathbb{D}_T|} \frac {\sum_{c=1}^{|C|}
	P @ c \Big ( C(\hat{\pmb{x}}_i),L(\hat{\pmb{x}}_i) \Big )\delta \big (c, C(\hat{\pmb{x}}_i) \big )} { |C(\hat{\pmb{x}}_i)|},
\end{equation}
where $\delta \big (c, C(\hat{\pmb{x}}_i) \big )=1$ if $c \in C(\hat{\pmb{x}}_i)$ and
$\delta \big (c, C(\hat{\pmb{x}}_i) \big )=0$ otherwise.

While I-MAP measures the accuracy in terms of test instances, L-MAP is used to evaluate the performance from a different perspective in light of candidate labels. Given a specific label $c \in C$, a model predicts the relatedness scores against the action specified by the $c$-th label for all the test instances in $\mathbb{D}_T$.
Hence, we can achieve an \emph{instance-based} ranking list for the $c$-th label, $L^c= \big \{ \hat{\pmb{x}}_{i_j} \big \}_{j=1}^{|\mathbb{D}_T|}$, in terms of their relatedness scores against the $c$-th label where $\forall \hat{\pmb{x}}_{i_p}, \hat{\pmb{x}}_{i_q} \in \mathbb{D}_T$, ${\rm Score}(\hat{\pmb{x}}_{i_p},c) \geq {\rm Score}(\hat{\pmb{x}}_{i_q},c)$ if $p<q$.
Let $D^c$ denote the collection of those test instances of which their ground-truth label sets indeed include the $c$-th label. Thus, the L-MAP over a test dataset, $\mathbb{D}_T$, is defined by
\begin{equation}
\label{eq:L-MAP}
{\rm L\!\!-\!\!MAP} = \frac 1 {|C|} \sum_{c=1}^{|C|} \frac {\sum_{i=1}^{|\mathbb{D}_T|}
	P @ i\Big ( D^c,L^c \Big )\delta \big (\hat{\pmb{x}}_i, D^c) \big )} { |D^c|},
\end{equation}
where $\delta(\hat{\pmb{x}}_i, D^c)=1$ if $\hat{\pmb{x}}_i \in D^c$ and
$\delta(\hat{\pmb{x}}_i, D^c)=0$ otherwise.

Those widely used evaluation metrics in information retrieval have also been used in evaluating multi-label learning systems, e.g., \cite{zhang2014multi-label,gong2013deep}. In our experiments, we adopt overall top-$k$ precision, recall and ${\rm F_1}$ score measured over a test dataset, $\mathbb{D}_T$, which are defined as follows:

\begin{equation}
\label{eq_precision}
{\rm precision}(k) = \frac {\sum_{i=1}^{|\mathbb{D}_T|}
	P @ k \big ( C(\hat{\pmb{x}}_i), L(\hat{\pmb{x}}_i) \big )}
{k*|\mathbb{D}_T|},
\end{equation}

\begin{equation}
\label{eq_recall}
{\rm recall}(k)= \frac {\sum_{i=1}^{|\mathbb{D}_T|}
	P @ k \big ( C(\hat{\pmb{x}}_i), L(\hat{\pmb{x}}_i) \big )}
{\sum_{i=1}^{|\mathbb{D}_T|} |C(\hat{\pmb{x}}_i)|},
\end{equation}

\begin{equation}
\label{eq_f1}
{\rm F_1}(k)=\frac{2*{\rm precision}(k)*{\rm recall}(k)}{{\rm precision}(k)+{\rm recall}(k)}.
\end{equation}

\subsection{Comparative Study}
\label{comparative}

In our experiments, we systematically conduct a  comparative study from two different perspectives: ablation study and state-of-the-art models. As a result, a number of baseline systems are designed to demonstrate roles played by the main components in our framework while several state-of-the-art multi-label ZSL algorithms are adapted for human action recognition. For the comparative study, we evaluate each of different models on three evaluation scenarios with evaluation metrics described in Sections \ref{sect_evaluation_scenarios} and \ref{sect_evaluation} under the exactly same conditions, including visual and semantic representations. As there are alternative pooling strategies that could be used to implement our framework, we further investigate those pooling strategies by comparing them to the average pooling used in our framework.

\subsubsection{Baseline Models}
\label{sect_ablation}
To investigate the roles played by different component mechanisms employed in our framework, we design four
baseline models, \emph{random guess of scores}, \emph{non-recurrent connection, without semantic embedding} and \emph{randomized label representation}, by manipulating our framework with different purposes described as follows: \\
\noindent
\textbf{Random guess of scores (RGS)}:
This is a general baseline that provides a lowest performance bound used for a reference to improvement made by a learning model. In our work, we randomly generate relatedness scores of all the candidate labels for a test instance. Then the performance of this baseline model is evaluated based on the random guess of scores. For reliability, we repeat the RGS process 100 times in our experiments and the statistics of the RGS performance including mean and standard error of mean (SEM) are reported.\\
\noindent
\textbf{Non-recurrent connection (NRC)}:
In our framework, a LSTM layer of recurrent connections is employed to capture temporal coherence underlying sequential video data in the visual embedding learning. To examine the role played by the LSTM layer, we replace the recurrent connected layer with a fully connected layer without recurrent connections and keep all other components in our framework unchanged. By this setting, our model is converted into a baseline model named \emph{non-recurrent connection}. During learning, obviously, this baseline model no longer explicitly makes use of the temporal dependency information underlying sequential segments in a video clip. Algorithm \ref{alg_alternating} is used directly for parameter estimation.\\
\noindent
\textbf{Without semantic embedding (WSE)}:
In our framework, there is a semantic model for semantic embedding with the motivation that the use of a joint latent ranking embedding space narrows the semantic gap between visual and semantic domains and the zero-shot recognition should be done in the joint latent ranking embedding space. However, some existing works, e.g., Fast0Tag
\cite{zhang2016fast}, do not learn a semantic embedding and the zero-shot recognition takes place directly in the semantic space. To examine the effectiveness of our semantic embedding, we come up with a baseline model named \emph{without semantic embedding} by removing the semantic model from our framework. Thus, the original semantic representations are used to replace the semantic embedding representations, $\pmb{E}^s$, required by the score layer in the visual model, which is amount to mapping the visual space directly onto the original semantic space. As this baseline model has only the visual model, the learning becomes simpler; i.e.,  solving the optimization problem formulated in Eq. (\ref{eq_vmodel}) based on the original semantic representation with the Adam \cite{kingma2014adam}.
It is worth clarifying that this baseline model is similar to Fast0Tag \cite{zhang2016fast} apart from an LSTM-based visual embedding model and the segment-level visual representation used in this baseline model while a feed-forward neural network and instance-level visual representation are employed by Fast0Tag for visual embedding.\\
\noindent
\textbf{Randomized label representation (RLR)}:
One of the most important issues in ZSL is exploring/exploiting the side information conveyed in the semantic domain. As our framework works for multi-label zero-shot recognition, we would investigate whether the semantic relatedness information encoded in the semantic embedding, inherited from the original semantic representations, is effectively used in knowledge transfer. To this end, we design another baseline model named \emph{randomized label representation} by replacing the word vector of a label with a vector of the same dimensionality that is generated randomly and normalized with the $l_2$ norm to ensure that it has the same range as that of the word vector. Apparently, the semantic relatedness information no longer exists in such randomized label representations. For parameter estimation,  Algorithm \ref{alg_alternating} is used directly via replacing the semantic representations of labels with the randomized label representations in training data.\\

\subsubsection{State-of-the-Art Methods}
\label{sect_comparative}

Although, to the best of our knowledge, there exists no work in multi-label zero-shot human action recognition, we notice that there are a few multi-label ZSL algorithms. In our comparative study, we adopt and extend those multi-label ZSL algorithms for human action recognition for a thorough evaluation of our proposed framework. Below, we briefly describe those multi-label ZSL algorithms used in our experiments. \\
\noindent
\textbf{Direct Semantic Prediction (DSP)}: DSP is a well-known  baseline model used in previous works for multi-label ZSL, e.g., \cite{sandouk2016multi}. DSP is derived from  \emph{direct attribute prediction} originally proposed for single-label ZSL \cite{lampert2014attribute}. The idea behind DSP is learning a mapping function $\phi: X \rightarrow \mathcal{S}$  from visual to semantic space directly for ZSL.  By using the composition property of word vectors, given a multi-labelled video clip, we use the mean word vector achieved by averaging those word vectors of the labels associated with this video clip to be its semantic representation. Thus, the multi-label ZSL problem is boiled down to single-label ZSL.
In our experiments, we employ support vector regressor models to learn the mapping function $\phi(\cdot)$.  Given a test instance $\pmb{\hat{x}}$, the learned $\phi(\cdot)$ is used to predict its compositional semantic representation $\pmb{\hat{s}} = \phi(\pmb{\hat{x}})$, then the prediction scores regarding different labels are estimated by measuring distances between the predictions and the word vectors of all the labels in a label collection $C$:
${\rm Score}(\pmb{\hat{x}}, c) = <\!\pmb{\hat{s}},\pmb{s}_c\!>,~\forall c \in C$,
which leads to a label-based ranking list, $L(\pmb{\hat{x}})$, in terms of semantic relatedness.\\
\noindent
\textbf{Convex combination of Semantic Embedding (ConSE)}:
ConSE is a ZSL algorithm  proposed by \cite{norouzi2013zero}, which can be naturally applied to multi-label ZSL.
As same as formulated in DSP, ConSE also learns a mapping to predict a compositional semantic representation from the visual representation of a given video clip. Instead of learning a direct mapping function as DSP does, however, ConSE  learns  the conditional probabilities,  $P(c|\pmb{x})$ for $\forall c \in C^{T\!r}$, regarding all the known actions with training data.
For recognition, the compositional or collective semantic representation of a test instance $\pmb{\hat{x}}$ is estimated by a convex combination
of the semantic representations of top-5 known actions of the highest conditional probabilities. The combination weights are the $l_2$ normalized conditional probabilities of those top-5 known actions:
$\pmb{\hat{s}} = \sum_{c \in C^{T\!r}} P(c|\pmb{\hat{x}})\pmb{s}_c$.
Thus, the prediction scores regarding all the labels in an action label collection $C$ are
${\rm Score}(\pmb{\hat{x}}, c)=<\!\pmb{\hat{s}},\pmb{s}_c\!>, ~\forall c \in C$, which leads to
a label-based ranking list, $L(\pmb{\hat{x}})$.\\
\noindent
\textbf{COSTA}: COSTA is a method proposed by \cite{mensink2014costa} especially for multi-label zero-shot classification. In this method, multi-label classification is converted into a number of binary classification problems via a one-vs-rest setting. $|C^{T\!r}|$ linear binary SVMs are trained based on the examples regarding $|C^{T\!r}|$ known actions.
Then the parameters of the SVM for an unseen label $c \in C^U$ is estimated by a weighted combination of the parameters of $|C^{T\!r}|$ trained SVMs corresponding to known actions:
$
\pmb{w}_c = \sum_{k=1}^{|C^{T\!r}|} \alpha_k  \beta_{ck} \pmb{w}_k.$
Here, $\pmb{w}_k$ is the parameters of the SVM regarding the
$k$-th label in $C^{T\!r}$ and $\alpha_k$ is a combination coefficient regarding the importance of this SVM achieved via learning.
$\beta_{ck} = {\exp(-d_{ck})} / {\sum_{j=1}^{|C^{T\!r}|} \exp(-d_{cj})}$, which is a factor indicating the relatedness between a unseen label, $c \in C^U$, and a known label, $k \in C^{T\!r}$, and  $d_{ck}$ is the semantic distance between labels $c$ and $k$ measured via their word vectors.
Thus, the SVM with the parameters $\pmb{w}_c$ can be used to predict the unseen label $c$ for a given test instance. Note that our experimental results not reported in this paper due to the limited space suggest that learning $\alpha_k$ is not only time consuming but also yields the poorer performance than that where all $|C^{T\!r}|$ SVMs are treated equally; i.e., $\alpha_k=1$ for $k=1, \cdots, |C^{T\!r}|$. Later on, we only report the best performance under this setting.\\
\noindent
\textbf{Fast0Tag}: Fast0Tag is one of the latest state-of-the-art methods proposed for multi-label image tagging and multi-label ZSL \cite{zhang2016fast}. The main idea behind Fast0Tag is learning a mapping function $\phi: X \rightarrow S$ from visual to semantic space for multi-label zero-shot tagging and recognition. Unlike DSP, a ranking-based loss function, \textit{RankNet}, is used to train a deep network to carry out $\phi(\cdot)$ so that for an video clip, its relevant labels should be ranked ahead of those irrelevant ones. For recognition, the mapping function yields the predicted semantic representation, $\phi(\hat{{\pmb x}})$, for a test instance, $\hat{{\pmb x}}$. Then, we can achieve the relatedness scores to all the labels in an action label collection and the label-based ranking list as same as done in DSP and ConSE. In our experiments, we strictly follow the same settings suggested by \cite{zhang2016fast}. \\
\noindent
\textbf{Fast0Tag+}: Our work presented in this paper suggests that the use of learned semantic embedding leads to better performance than the use of the original semantic representations directly. To further investigate this idea, we make an extension of Fast0Tag by incorporating our semantic model into the Fast0Tag model and name our extension \textit{Fast0Tag+}. As a result, Fast0Tag+ has an architecture resembling ours (c.f. the left box of Fig.\ref{fig_framework}), where the visual model is carried out by the original Fast0Tag architecture while the semantic model is the same as ours presented in Section \ref{sect:model_des}. The original rank loss functions in Fast0Tag are used and our alternate learning algorithm described in Algorithm \ref{alg_alternating} is used for parameter estimation. For recognition, the same procedure presented in Section
\ref{sect_multi-label ZSL} is used for a given test instance.
Here, we argue that this extension would provide further evidence in examining the effectiveness of semantic embedding learning.

\subsubsection{Pooling Strategy}
To investigate the effect of different pooling strategies over temporal relatedness scores, we conduct a comparative experiment by replacing the average pooling with either the maximum pooling or the local average global maximum pooling in our framework. For the maximum pooling,  Eq.(\ref{eq_avescore}) for the average pooling is thus altered to
\begin{equation}
\label{eq_maxscore}
\pmb{o}_i = \max_{t=1}^T \pmb{o}_{it}.
\end{equation}
For the local average global maximum pooling, we firstly divide the $T$ segments into $T_s$ groups with  a $50\%$ overlap between two consecutive groups. As a result, there are $N_g=2*T/T_s$ consecutive segments in each group. We calculate the average score in each group and find the maximum as follows:
\begin{equation}
\label{eq_lagmscore}
\pmb{o}_i = \max_{t_s=1}^{T_s} \frac{1}{N_g} \sum_{t=(t_s-1)N_g/2+1}^{(t_s+1)N_g/2} \pmb{o}_{it}.
\end{equation}
Note that the local average global maximum pooling strategy is generic, and the average pooling and maximum pooling can be viewed as its special cases without between-group overlapping: the average pooling when $T_s=1, N_g=T$ and the maximum pooling when $T_s=T, N_g=1$, respectively. To make a thorough investigation, we set $T_s=10, N_g=60$ and $T_s=20, N_g=30$ as two experimental settings for Breakfast dataset. For Charades dataset, we set $T_s=5, N_g=8$ and $T_s=10, N_g=4$. All other experimental settings are kept the same for a fair comparison.

In our comparative study, the optimal hyper-parameters involved in baseline and state-of-the-art learning models are sought during their learning with the same cross-validation procedure as described in Section \ref{sub:model_learn}. Moreover, five state-of-the-art methods described above and ours are extensible to multi-label recognition straightforwardly; i.e.,
all the actions are known in advance and their training examples are available during learning. Thus, we also report the multi-label recognition performance, which not only extends our comparative study in a wider scope but also provides a benchmark to see how much the performance of each method is degraded in a zero-shot circumstance. For experiments in comparison of different pooling strategies, all the components and setting are kept unchanged except the pooling operations.

\section{Experimental Results}
\label{sect:results}

In this section, we report the detailed experimental results in different settings and exemplify some typical test instances via visual inspection.

\subsection{Results on Learning}\label{sub:resLearning}
We first report the experimental results regarding learning including optimal hyper-parameters for all the models used in our experiments and the evolution of the learning process for our model trained with our proposed alternate learning algorithm (Algorithm \ref{alg_alternating}) under different data split settings.

As described in Section \ref{sub:model_learn}, we employ a grid-based search procedure via cross-validation to find out the optimal hyper-parameters in terms of both the loss used to train a model and the I-MAP performance (c.f. Section \ref{sect_evaluation}) as this metric directly evaluate the relatedness of a video instance to all the labels in a considered action label collection. We seek an optimal value from a set of candidate hyper-parameters involved in all different learning models used in our experiments with the exactly same procedure as follows:\\

\noindent
\textbf{Network architecture}:
The optimal architecture of neural networks in a model used in our experiments is investigated by tuning different number of neurons in each hidden layer. In our proposed model, there are totally four structural hyper-parameters.  The number of hidden units in the LSTM layer is selected from the candidate set, $N_1=256,512,1024$. In the visual model, the  number of neurons in the hidden layer above the LSTM layer is investigated with $N_2=1024,2048$. In the semantic model,  the number of neurons in the first hidden layer is selected from $N_1=300,500,700$. As a critical hyper-parameter in our algorithm, the dimension of latent embedding space $d_e$, the number of neurons in the visual/semantic embedding layers, is investigated by setting the candidate values, $d_e =200,500,800$. For the \textit{non-recurrent} baseline model,  the number of neurons in the first hidden layer replacing the LSTM layer is chosen from  $N_1=1024,2048,4096$. For Fast0Tag and Fast0Tag+, the number of neurons in the first and second hidden layers are selected from $N_1= 2048, 4096, 8092$ and $N_2 = 1024, 2048$, respectively.\\
\textbf{Learning rate}:
For all the neural networks in the proposed model, the baseline and the state-of-the-art models, candidate learning rates are $\{\texttt{1e\!-\!2}, \texttt{1e\!-\!4}\}$ and $\{\texttt{1e\!-\!4}, \texttt{1e\!-\!6}\}$ for the visual and  the semantic models, respectively.\\
\noindent
\textbf{Number of epochs}:
Learning is stopped when the I-MAP performance on a validation set is no longer improved within the last 10 epochs and the loss reaches a low level on both training and validation sets. Then, the optimal model chosen is the one that yields the highest value of I-MAP on the validation set.\\
\noindent
\textbf{Dropout rate}: The dropout rate used in the first layer of a neural model during learning is selected from $\{0, 0.5\}$.\\
\noindent
\textbf{Margin}: The margin used in the hinge rank loss is selected for $m =0.1, 1, 10$.\\
\noindent
\textbf{SVM hyper-parameters:} In our comparative study, ConSE \cite{norouzi2013zero} and COSTA  \cite{mensink2014costa} employ a linear SVM for classification and DSP \cite{lampert2014attribute} uses a linear SVR for regression. In our experiments, an optimal soft-margin value is sought from $C=0.01, 1, 100$. For SVR, the percentage of support vectors is always set to $\epsilon=0.1$ as suggested in literature.

As a result, the resultant optimal hyper-parameter values in different experimental settings are summarized in Table \ref{table_hyper}.

\begin{figure*}[htbp]
	\caption{The evolution of the regularized RankNet losses, $L_v$ and $L_s$, on training data and the I-MAP values on validation data during the joint visual and semantic embedding learning with our alternate learning algorithm described in Algorithm \ref{alg_alternating}. All the results are achieved based on  Split 1 on two datasets under the IFS and the LFS settings (c.f. Table \ref{table_datasplit}).}
	\centering
	\subfloat{{
			\includegraphics[width=.4\textwidth]{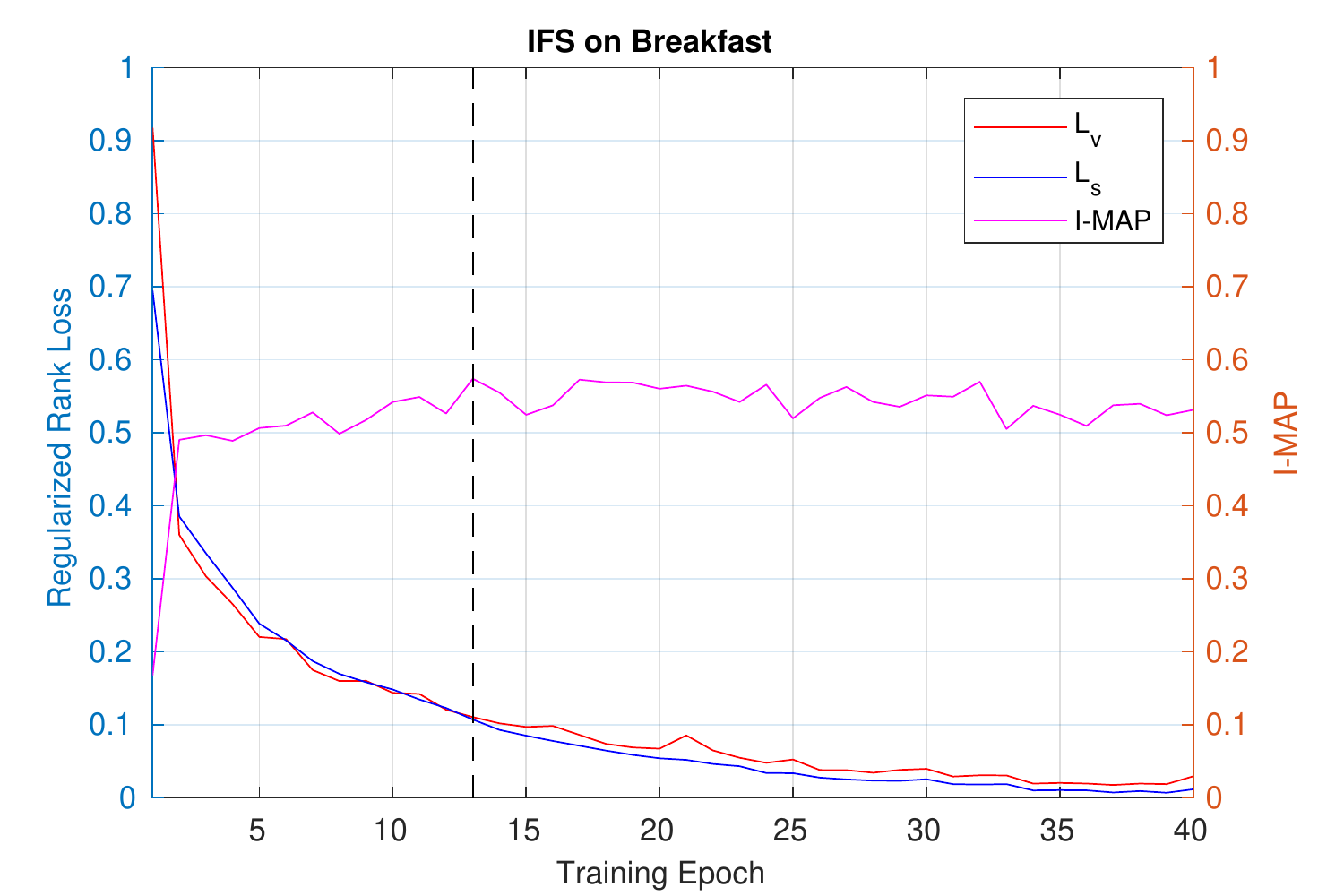}
	}}
	\qquad
	\subfloat{	\includegraphics[width=.4\textwidth]{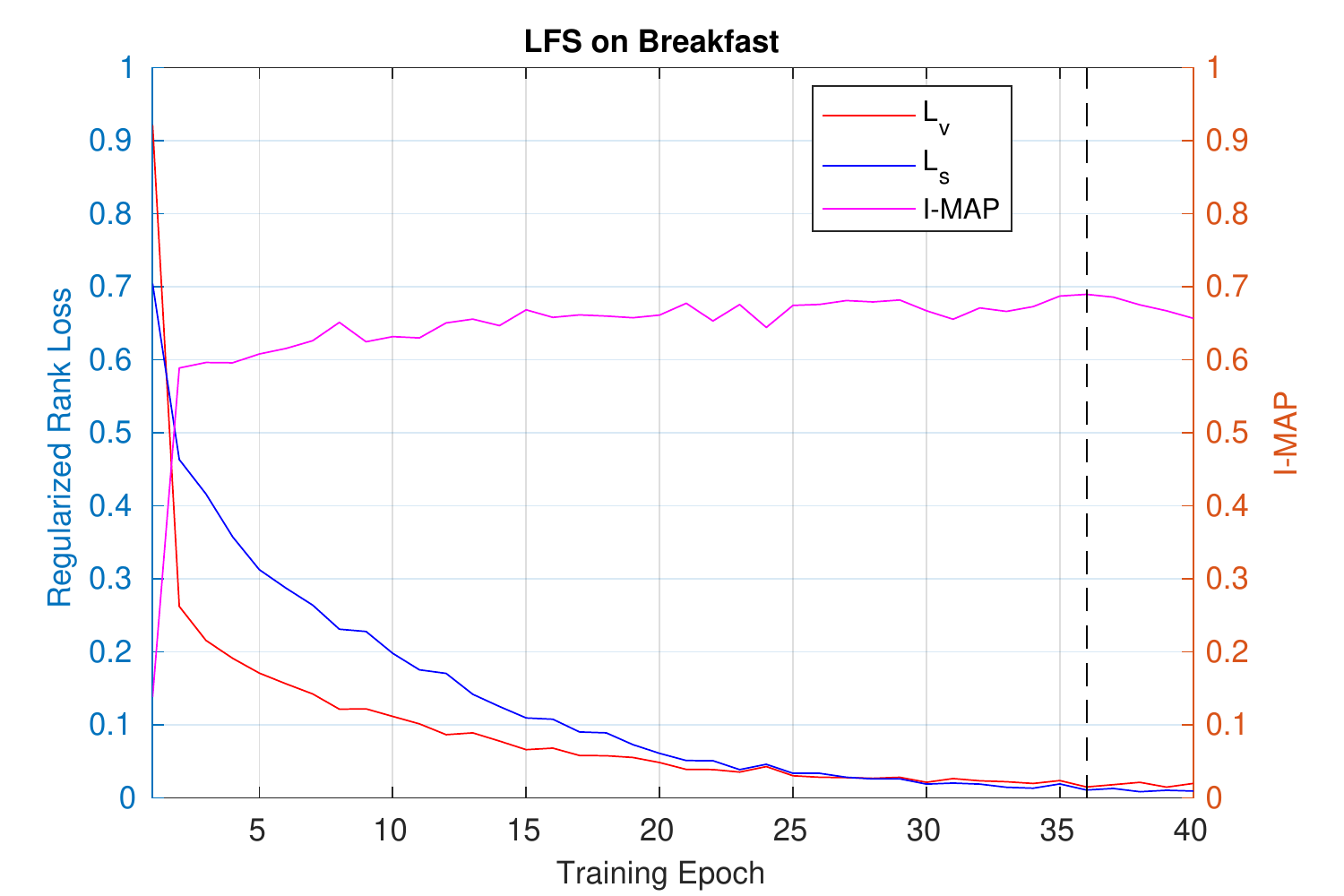}
		
	}
	\qquad
	\subfloat{
		\includegraphics[width=.4\textwidth]{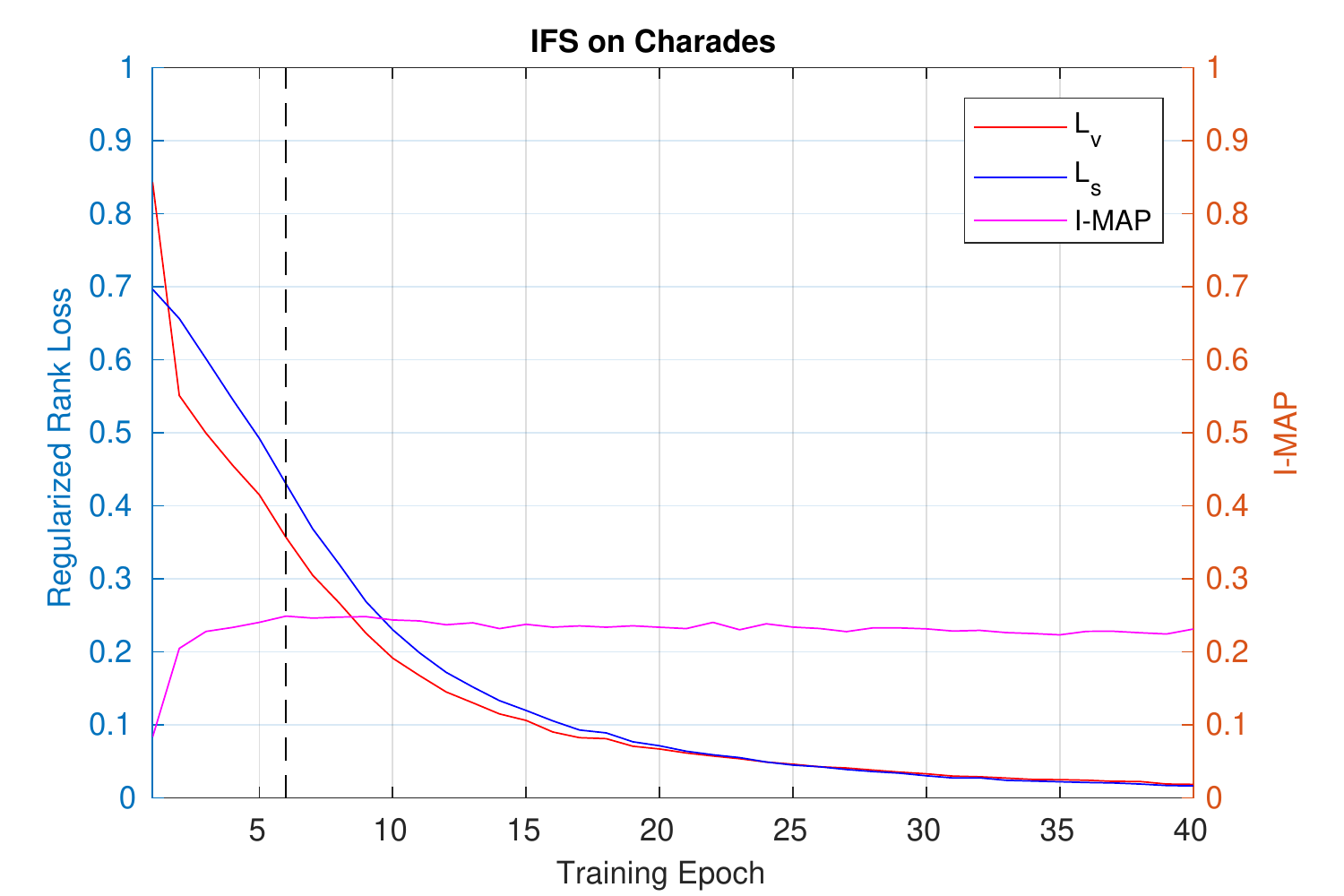}
	}
	\qquad
	\subfloat{
		\includegraphics[width=.4\textwidth]{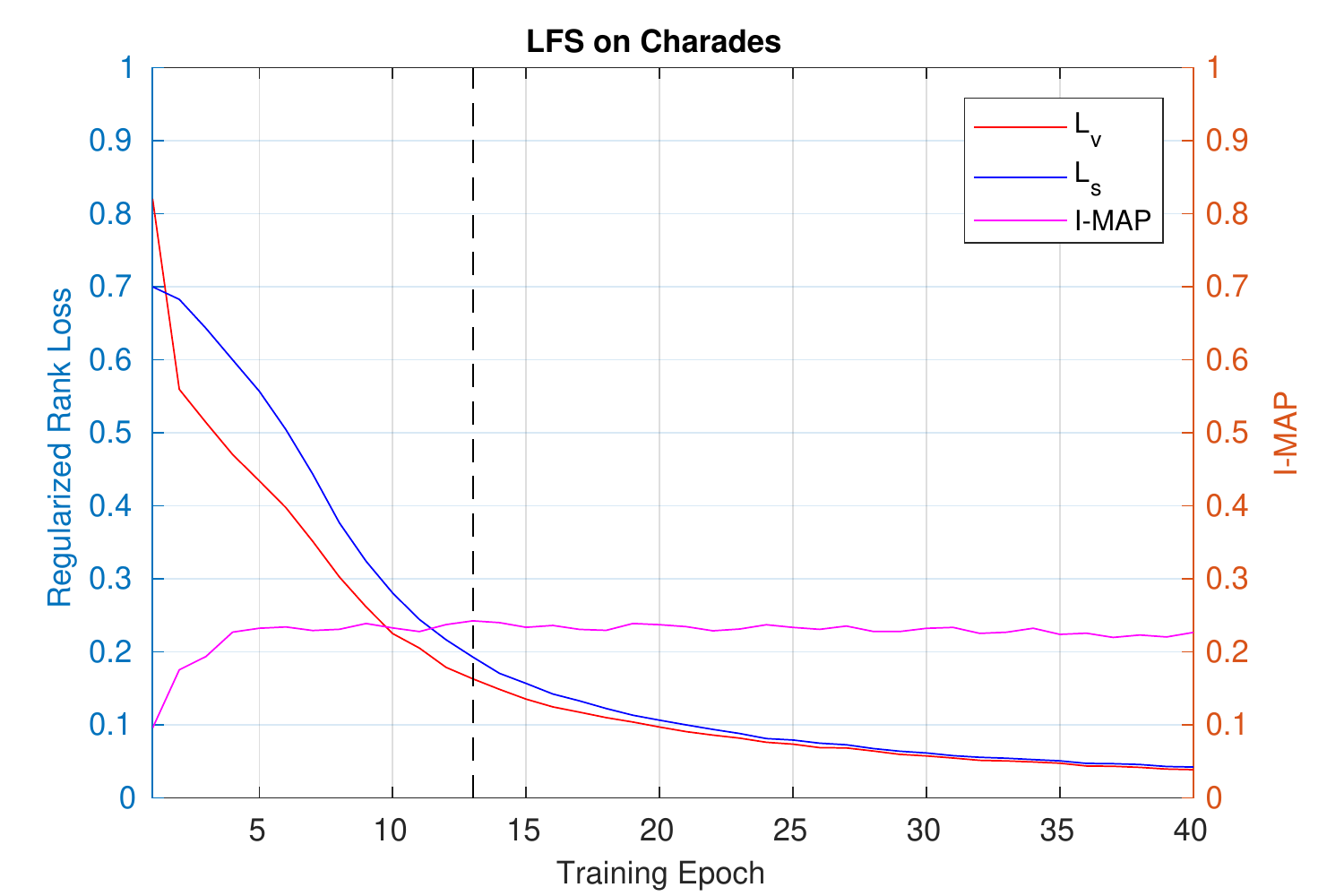}
	}
	\label{fig_lossCurves}
\end{figure*}

To train our model described in Section \ref{sect:model_des}, our proposed learning algorithm optimizes two rank loss functions alternately for joint visual and semantic embedding learning. With the regularized RankNet loss functions, we would exhibit the learning behavior during the training. As illustrated in Figure \ref{fig_lossCurves}, the regularized rank losses, $L_v$ and $L_s$, with respect to the visual and the semantic models keep decreasing steadily on training data as the training epochs increase regardless of the data split settings and datasets. Nevertheless, we adopt the early-stop strategy to avoid overfitting. However, we observe that the change of two ranking losses on validation data fluctuates wildly in learning so that we cannot decide a proper early-stop point easily. Instead we use the I-MAP measured on validation data to decide the proper early-stop points, as shown in Figure \ref{fig_lossCurves} where the bars of dash line indicate the actual point that the learning is stopped for different training datasets. In general, all our experiments in learning (including not shown in Figure \ref{fig_lossCurves}) suggest that our alternate learning algorithm always converges regardless of different rank losses and datasets under different data split settings.

\begin{table*}[!htbp]
	{\large
		\centering
		\caption[]{Optimal hyper-parameter values of different learning models found by grid search. \textbf{Notation}:
			IFS -- Instance-First Split; LFS -- Label-First Split;
			V -- Visual model; S -- Semantic model; lr -- learning rate; C, $\epsilon$ -- soft-margin  and percentage of support vectors in SVM/SVR; $m$ -- margin in the hinge ranking loss.  $N_1 \rightarrow N_2 \rightarrow d_e$ indicates a neural network architecture where $N_1$(dropout rate) is the number of neurons in the first hidden layer and dropout rate used in learning;  $N_2$ is the number of hidden neurons in the second hidden layer; and $d_e$ is the number of neurons in the latent embedding layer.}
		\label{table_hyper}
\begin{lrbox}{\tablebox}
	\begin{tabular}{c|c|c|c|c|c}
		\hline\hline
		\multirow{2}{*}{\textbf{Dataset}} & \multirow{2}{*}{\textbf{Data Split}} & \multirow{2}{*}{\textbf{Model}} &  \multicolumn{3}{c}{\textbf{Split}}\\ \cline{4-6}
		& & &  \textbf{1} &  \textbf{2} &  \textbf{3} \\
		
		\hline
		\multirow{40}{*}{Breakfast} & \multirow{20}{*}{IFS}
		& \multirow{2}{*}{NRC(RankNet)} & ${\rm V\!:}~{\rm lr}=1e\!\!-\!\!4;1024 (0.5) \rightarrow 2048 \rightarrow 500$ &${\rm V\!:}~{\rm lr}=1e\!\!-\!\!4;1024 (0.5) \rightarrow 2048 \rightarrow 500$  &${\rm V\!:}~{\rm lr}=1e\!\!-\!\!4;2048 (0.5) \rightarrow 1024 \rightarrow 500$  \\
		&&& ${\rm S\!:}~{\rm lr}=1e\!\!-\!\!6;500 \rightarrow 500$ &${\rm S\!:}~{\rm lr}=1e\!\!-\!\!6;500 \rightarrow 500$ & ${\rm S\!:}~{\rm lr}=1e\!\!-\!\!6;500 \rightarrow 500$  \\
		&& \multirow{2}{*}{NRC(Hinge)} & ${\rm V\!:}~{\rm lr}=1e\!\!-\!\!4;1024 (0.5) \rightarrow 2048 \rightarrow 500$ &${\rm V\!:}~{\rm lr}=1e\!\!-\!\!4;1024 (0.5) \rightarrow 2048 \rightarrow 500$  &${\rm V\!:}~{\rm lr}=1e\!\!-\!\!4;1024 (0.5) \rightarrow 2048 \rightarrow 500$  \\
		&&& ${\rm S\!:}~{\rm lr}=1e\!\!-\!\!6;500 \rightarrow 500;m=1$ &${\rm S\!:}~{\rm lr}=1e\!\!-\!\!6;500 \rightarrow 500;m=1$ & ${\rm S\!:}~{\rm lr}=1e\!\!-\!\!6;700 \rightarrow 500;m=1$  \\
		&& WSE(RankNet) & ${\rm V\!:}~{\rm lr}=1e\!\!-\!\!4;512(0)\rightarrow 2048$ & ${\rm V\!:}~{\rm lr}=1e\!\!-\!\!4;512(0)\rightarrow 1024$ & ${\rm V\!:}~{\rm lr}=1e\!\!-\!\!4;1024(0)\rightarrow 2048$ \\
		&& WSE(Hinge) & ${\rm V\!:}~{\rm lr}=1e\!\!-\!\!4;1024(0.5)\rightarrow 2048;m=1$ & ${\rm V\!:}~{\rm lr}=1e\!\!-\!\!4;512(0)\rightarrow 1024;m=1$ & ${\rm V\!:}~{\rm lr}=1e\!\!-\!\!4;1024(0)\rightarrow 1024;m=1$ \\
		&& \multirow{2}{*}{RLR(RankNet)} & ${\rm V\!:}~{\rm lr}=1e\!\!-\!\!4;512 (0) \rightarrow 2048 \rightarrow 800$  &${\rm V\!:}~{\rm lr}=1e\!\!-\!\!4;256 (0.5) \rightarrow 1024 \rightarrow 500$  &${\rm V\!:}~{\rm lr}=1e\!\!-\!\!4;256 (0.5) \rightarrow 2048 \rightarrow 200$  \\
		&&& ${\rm S\!:}~{\rm lr}=1e\!\!-\!\!6;700 \rightarrow 800$ &${\rm S\!:}~{\rm lr}=1e\!\!-\!\!6;500 \rightarrow 500$ &${\rm S\!:}~{\rm lr}=1e\!\!-\!\!6;300 \rightarrow 200$  \\
		&& \multirow{2}{*}{RLR(Hinge)} & ${\rm V\!:}~{\rm lr}=1e\!\!-\!\!4;1024 (0) \rightarrow 2048 \rightarrow 500$  &${\rm V\!:}~{\rm lr}=1e\!\!-\!\!4;512 (0.5) \rightarrow 1024 \rightarrow 500$  &${\rm V\!:}~{\rm lr}=1e\!\!-\!\!4;512 (0.5) \rightarrow 2048 \rightarrow 200$  \\
		&&& ${\rm S\!:}~{\rm lr}=1e\!\!-\!\!6;700 \rightarrow 500;m=10$ &${\rm S\!:}~{\rm lr}=1e\!\!-\!\!6;500 \rightarrow 500;m=10$ &${\rm S\!:}~{\rm lr}=1e\!\!-\!\!6;500 \rightarrow 200;m=10$  \\
		\cline{3-6}
		&& DSP & $C=1,\epsilon=0.1$  & $C=1,\epsilon=0.1$ & $C=1,\epsilon=0.1$\\
		&& ConSE & $C=1$ & $C=1$& $C=1$\\
		&& COSTA & $C=1$ & $C=1$& $C=1$\\
		&& Fast0Tag & ${\rm V\!:}~{\rm lr}=1e\!\!-\!\!4;8192 (0)\rightarrow 1024$  & ${\rm V\!:}~{\rm lr}=1e\!\!-\!\!4; 8192 (0.5) \rightarrow 2048$ & ${\rm V\!:}~{\rm lr}=1e\!\!-\!\!2;8192 (0.5) \rightarrow 2048$\\
		&& \multirow{2}{*}{Fast0Tag+} & ${\rm V\!:}~{\rm lr}=1e\!\!-\!\!4;4096 (0) \rightarrow 1024 \rightarrow 800$  &${\rm V\!:}~{\rm lr}=1e\!\!-\!\!4;8192 (0.5) \rightarrow 2048 \rightarrow 800$  &${\rm V\!:}~{\rm lr}=1e\!\!-\!\!4;8192 (0) \rightarrow 1024 \rightarrow 200$  \\
		&&& ${\rm S\!:}~{\rm lr}=1e\!\!-\!\!6;500 \rightarrow 800$ &${\rm S\!:}~{\rm lr}=1e\!\!-\!\!6;700 \rightarrow 800$ &${\rm S\!:}~{\rm lr}=1e\!\!-\!\!6;300 \rightarrow 200$  \\
		&& \multirow{2}{*}{Ours(RankNet)}& ${\rm V\!:}~{\rm lr}=1e\!\!-\!\!4;512 (0) \rightarrow 2048 \rightarrow 800$  &${\rm V\!:}~{\rm lr}=1e\!\!-\!\!4;1024 (0) \rightarrow 2048 \rightarrow 500$  &${\rm V\!:}~{\rm lr}=1e\!\!-\!\!4;1024 (0) \rightarrow 2048 \rightarrow 500$  \\
		&&& ${\rm S\!:}~{\rm lr}=1e\!\!-\!\!6;700 \rightarrow 800$ &${\rm S\!:}~{\rm lr}=1e\!\!-\!\!6;700 \rightarrow 500$ &${\rm S\!:}~{\rm lr}=1e\!\!-\!\!6;700 \rightarrow 500$  \\
		&& \multirow{2}{*}{Ours(Hinge)} & ${\rm V\!:}~{\rm lr}=1e\!\!-\!\!4;256 (0.5) \rightarrow 2048 \rightarrow 500$  &${\rm V\!:}~{\rm lr}=1e\!\!-\!\!4;256 (0.5) \rightarrow 1024 \rightarrow 500$  &${\rm V\!:}~{\rm lr}=1e\!\!-\!\!4;1024 (0) \rightarrow 1024 \rightarrow 500$  \\
		&&& ${\rm S\!:}~{\rm lr}=1e\!\!-\!\!6;700 \rightarrow 500$; $m=1$ &${\rm S\!:}~{\rm lr}=1e\!\!-\!\!6;700 \rightarrow 500$; $m=1$ &${\rm S\!:}~{\rm lr}=1e\!\!-\!\!6;500 \rightarrow 500$; $m=1$  \\
		\cline{2-6}
		& \multirow{20}{*}{LFS}
		& \multirow{2}{*}{NRC(RankNet)} & ${\rm V\!:}~{\rm lr}=1e\!\!-\!\!4;2048 (0.5) \rightarrow 2048 \rightarrow 200$    &${\rm V\!:}~{\rm lr}=1e\!\!-\!\!4;2048 (0) \rightarrow 2048 \rightarrow 800$  &${\rm V\!:}~{\rm lr}=1e\!\!-\!\!4;2048 (0) \rightarrow 2048 \rightarrow 500$  \\
		&&& ${\rm S\!:}~{\rm lr}=1e\!\!-\!\!6;500 \rightarrow 200$ &${\rm S\!:}~{\rm lr}=1e\!\!-\!\!6;500 \rightarrow 800$ &${\rm S\!:}~{\rm lr}=1e\!\!-\!\!6;700 \rightarrow 500$  \\
		&& \multirow{2}{*}{NRC(Hinge)} & ${\rm V\!:}~{\rm lr}=1e\!\!-\!\!4;4096 (0) \rightarrow 2048 \rightarrow 500$    &${\rm V\!:}~{\rm lr}=1e\!\!-\!\!4;4096 (0) \rightarrow 2048 \rightarrow 500$  &${\rm V\!:}~{\rm lr}=1e\!\!-\!\!4;2048 (0.5) \rightarrow 2048 \rightarrow 200$  \\
		&&& ${\rm S\!:}~{\rm lr}=1e\!\!-\!\!6;300 \rightarrow 500;m=10$ &${\rm S\!:}~{\rm lr}=1e\!\!-\!\!6;300 \rightarrow 500;m=10$ &${\rm S\!:}~{\rm lr}=1e\!\!-\!\!6;500 \rightarrow 200;m=1$  \\
		&& WSE(RankNet) & ${\rm V\!:}~{\rm lr}=1e\!\!-\!\!4;1024(0)\rightarrow 1024$ & ${\rm V\!:}~{\rm lr}=1e\!\!-\!\!4;1024(0)\rightarrow 1024$ & ${\rm V\!:}~{\rm lr}=1e\!\!-\!\!4;1024(0.5)\rightarrow 1024$ \\
		&& WSE(Hinge) & ${\rm V\!:}~{\rm lr}=1e\!\!-\!\!4;1024(0.5)\rightarrow 2048;m=10$ & ${\rm V\!:}~{\rm lr}=1e\!\!-\!\!4;1024(0.5)\rightarrow 2048;m=10$ & ${\rm V\!:}~{\rm lr}=1e\!\!-\!\!4;512(0)\rightarrow 2048;m=1$ \\
		&& \multirow{2}{*}{RLR(RankNet)} & ${\rm V\!:}~{\rm lr}=1e\!\!-\!\!4;256 (0.5) \rightarrow 1024 \rightarrow 500$  &${\rm V\!:}~{\rm lr}=1e\!\!-\!\!4;512 (0.5) \rightarrow 1024 \rightarrow 200$  &${\rm V\!:}~{\rm lr}=1e\!\!-\!\!4;512 (0.5) \rightarrow 2048 \rightarrow 200$  \\
		&&& ${\rm S\!:}~{\rm lr}=1e\!\!-\!\!6;500 \rightarrow 500$ &${\rm S\!:}~{\rm lr}=1e\!\!-\!\!6;700 \rightarrow 200$ &${\rm S\!:}~{\rm lr}=1e\!\!-\!\!6;500 \rightarrow 200$  \\
		&& \multirow{2}{*}{RLR(Hinge)} & ${\rm V\!:}~{\rm lr}=1e\!\!-\!\!4;512 (0) \rightarrow 2048 \rightarrow 800$  &${\rm V\!:}~{\rm lr}=1e\!\!-\!\!4;256 (0) \rightarrow 1024 \rightarrow 800$  &${\rm V\!:}~{\rm lr}=1e\!\!-\!\!4;512 (0) \rightarrow 2048 \rightarrow 800$  \\
		&&& ${\rm S\!:}~{\rm lr}=1e\!\!-\!\!6;500 \rightarrow 800;m=10$ &${\rm S\!:}~{\rm lr}=1e\!\!-\!\!6;700 \rightarrow 800;m=10$ &${\rm S\!:}~{\rm lr}=1e\!\!-\!\!6;700 \rightarrow 800;m=10$  \\
		\cline{3-6}
		&& DSP & $C=100,\epsilon=0.1$  & $C=100,\epsilon=0.1$& $C=100,\epsilon=0.1$\\
		&& ConSE & $C=100$ & $C=100$& $C=100$\\
		&& COSTA & $C=100$ & $C=100$& $C=100$\\
		&& Fast0Tag & ${\rm V\!:}~{\rm lr}=1e\!\!-\!\!4;4096 (0.5)\rightarrow 2048$  & ${\rm V\!:}~{\rm lr}=1e\!\!-\!\!4; 4096 (0) \rightarrow 1024$ & ${\rm V\!:}~{\rm lr}=1e\!\!-\!\!2;8192 (0.5) \rightarrow 2048$\\
		&& \multirow{2}{*}{Fast0Tag+} & ${\rm V\!:}~{\rm lr}=1e\!\!-\!\!4;8192 (0.5) \rightarrow 1024 \rightarrow 800$  &${\rm V\!:}~{\rm lr}=1e\!\!-\!\!4;8192 (0.5) \rightarrow 1024 \rightarrow 800$  &${\rm V\!:}~{\rm lr}=1e\!\!-\!\!4;4096 (0) \rightarrow 1024 \rightarrow 200$  \\
		&&& ${\rm S\!:}~{\rm lr}=1e\!\!-\!\!6;500 \rightarrow 800$ &${\rm S\!:}~{\rm lr}=1e\!\!-\!\!6;500 \rightarrow 800$ &${\rm S\!:}~{\rm lr}=1e\!\!-\!\!6;300 \rightarrow 200$  \\
		&& \multirow{2}{*}{Ours(RankNet)} & ${\rm V\!:}~{\rm lr}=1e\!\!-\!\!4;512 (0) \rightarrow 1024 \rightarrow 800$  &${\rm V\!:}~{\rm lr}=1e\!\!-\!\!4;256 (0) \rightarrow 1024 \rightarrow 800$  &${\rm V\!:}~{\rm lr}=1e\!\!-\!\!4;512 (0.5) \rightarrow 1024 \rightarrow 200$  \\
		&&& ${\rm S\!:}~{\rm lr}=1e\!\!-\!\!6;500 \rightarrow 800$ &${\rm S\!:}~{\rm lr}=1e\!\!-\!\!6;700 \rightarrow 800$ &${\rm S\!:}~{\rm lr}=1e\!\!-\!\!6;700 \rightarrow 200$  \\
		&& \multirow{2}{*}{Ours(Hinge)} & ${\rm V\!:}~{\rm lr}=1e\!\!-\!\!4;512 (0.5) \rightarrow 1024 \rightarrow 200$  &${\rm V\!:}~{\rm lr}=1e\!\!-\!\!4;512 (0.5) \rightarrow 1024 \rightarrow 800$  &${\rm V\!:}~{\rm lr}=1e\!\!-\!\!4;256 (0) \rightarrow 2048 \rightarrow 500$  \\
		&&& ${\rm S\!:}~{\rm lr}=1e\!\!-\!\!6;500 \rightarrow 200$; $m=10$ &${\rm S\!:}~{\rm lr}=1e\!\!-\!\!6;500 \rightarrow 800$; $m=10$ &${\rm S\!:}~{\rm lr}=1e\!\!-\!\!6;500 \rightarrow 500$; $m=1$  \\
		\hline \hline
		
		\multirow{40}{*}{Charades} & \multirow{20}{*}{IFS}
		& \multirow{2}{*}{NRC(RankNet)} & ${\rm V\!:}~{\rm lr}=1e\!\!-\!\!4;1024 (0) \rightarrow 1024 \rightarrow 500$  &${\rm V\!:}~{\rm lr}=1e\!\!-\!\!4;1024 (0) \rightarrow 1024 \rightarrow 500$  &${\rm V\!:}~{\rm lr}=1e\!\!-\!\!4;2048 (0) \rightarrow 1024 \rightarrow 200$  \\
		&&& ${\rm S\!:}~{\rm lr}=1e\!\!-\!\!6;500 \rightarrow 500$ &${\rm S\!:}~{\rm lr}=1e\!\!-\!\!6;500 \rightarrow 500$ &${\rm S\!:}~{\rm lr}=1e\!\!-\!\!6;700 \rightarrow 200$  \\
		&& \multirow{2}{*}{NRC(Hinge)} & ${\rm V\!:}~{\rm lr}=1e\!\!-\!\!4;2048 (0) \rightarrow 2048 \rightarrow 800$  &${\rm V\!:}~{\rm lr}=1e\!\!-\!\!4;2048 (0) \rightarrow 2048 \rightarrow 800$  &${\rm V\!:}~{\rm lr}=1e\!\!-\!\!4;4096 (0.5) \rightarrow 2048 \rightarrow 500$  \\
		&&& ${\rm S\!:}~{\rm lr}=1e\!\!-\!\!6;700 \rightarrow 800;m=10$ &${\rm S\!:}~{\rm lr}=1e\!\!-\!\!6;500 \rightarrow 800;m=10$ &${\rm S\!:}~{\rm lr}=1e\!\!-\!\!6;300 \rightarrow 500;m=10$  \\
		&& WSE(RankNet) & ${\rm V\!:}~{\rm lr}=1e\!\!-\!\!4;1024(0.5)\rightarrow 2048$ & ${\rm V\!:}~{\rm lr}=1e\!\!-\!\!4;512(0.5)\rightarrow 2048$ & ${\rm V\!:}~{\rm lr}=1e\!\!-\!\!4;256(0.5)\rightarrow 2048$ \\
		&& WSE(Hinge) & ${\rm V\!:}~{\rm lr}=1e\!\!-\!\!4;1024(0.0)\rightarrow 2048$ & ${\rm V\!:}~{\rm lr}=1e\!\!-\!\!4;1024(0.0)\rightarrow 1024$ & ${\rm V\!:}~{\rm lr}=1e\!\!-\!\!4;1024(0.0)\rightarrow 2048$ \\
		&& \multirow{2}{*}{RLR(RankNet)} & ${\rm V\!:}~{\rm lr}=1e\!\!-\!\!4;256 (0.5) \rightarrow 1024 \rightarrow 500$  &${\rm V\!:}~{\rm lr}=1e\!\!-\!\!4;256 (0.5) \rightarrow 1024 \rightarrow 500$  &${\rm V\!:}~{\rm lr}=1e\!\!-\!\!4;256 (0.5) \rightarrow 1024 \rightarrow 500$  \\
		&&& ${\rm S\!:}~{\rm lr}=1e\!\!-\!\!6;500 \rightarrow 500$ &${\rm S\!:}~{\rm lr}=1e\!\!-\!\!6;500 \rightarrow 500$ &${\rm S\!:}~{\rm lr}=1e\!\!-\!\!6;500 \rightarrow 500$  \\
		&& \multirow{2}{*}{RLR(Hinge)} & ${\rm V\!:}~{\rm lr}=1e\!\!-\!\!4;1024 (0.5) \rightarrow 2048 \rightarrow 500$  &${\rm V\!:}~{\rm lr}=1e\!\!-\!\!4;1024 (0.5) \rightarrow 2048 \rightarrow 500$  &${\rm V\!:}~{\rm lr}=1e\!\!-\!\!4;1024 (0.5) \rightarrow 2048 \rightarrow 500$  \\
		&&& ${\rm S\!:}~{\rm lr}=1e\!\!-\!\!6;700 \rightarrow 500;m=10$ &${\rm S\!:}~{\rm lr}=1e\!\!-\!\!6;700 \rightarrow 500;m=10$ &${\rm S\!:}~{\rm lr}=1e\!\!-\!\!6;700 \rightarrow 500;m=10$  \\
		\cline{3-6}
		&& DSP & $C=1,\epsilon=0.1$  & $C=1,\epsilon=0.1$& $C=1,\epsilon=0.1$\\
		&& ConSE & $C=1$ & $C=1$& $C=1$\\
		&& COSTA & $C=1$ & $C=1$& $C=1$\\
		&& Fast0Tag & ${\rm V\!:}~{\rm lr}=1e\!\!-\!\!4;4096 (0.5)\rightarrow 2048$  & ${\rm V\!:}~{\rm lr}=1e\!\!-\!\!4; 8192 (0) \rightarrow 2048$ & ${\rm V\!:}~{\rm lr}=1e\!\!-\!\!4;4096 (0.5) \rightarrow 2048$\\
		&& \multirow{2}{*}{Fast0Tag+} & ${\rm V\!:}~{\rm lr}=1e\!\!-\!\!4;8192 (0.5) \rightarrow 2048 \rightarrow 800$  &${\rm V\!:}~{\rm lr}=1e\!\!-\!\!4;8192 (0.5) \rightarrow 2048 \rightarrow 800$  &${\rm V\!:}~{\rm lr}=1e\!\!-\!\!4;8192 (0.5) \rightarrow 1024 \rightarrow 800$  \\
		&&& ${\rm S\!:}~{\rm lr}=1e\!\!-\!\!6;700 \rightarrow 800$ &${\rm S\!:}~{\rm lr}=1e\!\!-\!\!6;700 \rightarrow 800$ &${\rm S\!:}~{\rm lr}=1e\!\!-\!\!6;500 \rightarrow 800$  \\
		&& \multirow{2}{*}{Ours(RankNet)}& ${\rm V\!:}~{\rm lr}=1e\!\!-\!\!4;256 (0.5) \rightarrow 2048 \rightarrow 800$  &${\rm V\!:}~{\rm lr}=1e\!\!-\!\!4;256 (0.5) \rightarrow 2048 \rightarrow 800$  &${\rm V\!:}~{\rm lr}=1e\!\!-\!\!4;1024 (0.5) \rightarrow 1024 \rightarrow 800$  \\
		&&& ${\rm S\!:}~{\rm lr}=1e\!\!-\!\!6;700 \rightarrow 800$ &${\rm S\!:}~{\rm lr}=1e\!\!-\!\!6;700 \rightarrow 800$ &${\rm S\!:}~{\rm lr}=1e\!\!-\!\!6;500 \rightarrow 800$  \\
		&& \multirow{2}{*}{Ours(Hinge)} & ${\rm V\!:}~{\rm lr}=1e\!\!-\!\!4;256 (0.5) \rightarrow 2048 \rightarrow 500$  &${\rm V\!:}~{\rm lr}=1e\!\!-\!\!4;256 (0.5) \rightarrow 2048 \rightarrow 800$  &${\rm V\!:}~{\rm lr}=1e\!\!-\!\!4;1024 (0.5) \rightarrow 1024 \rightarrow 800$  \\
		&&& ${\rm S\!:}~{\rm lr}=1e\!\!-\!\!6;700 \rightarrow 500$; $m=1$ &${\rm S\!:}~{\rm lr}=1e\!\!-\!\!6;700 \rightarrow 800$; $m=1$ &${\rm S\!:}~{\rm lr}=1e\!\!-\!\!6;500 \rightarrow 800$; $m=1$  \\
		\cline{2-6}
		& \multirow{20}{*}{LFS}
		
		& \multirow{2}{*}{NRC(RankNet)} & ${\rm V\!:}~{\rm lr}=1e\!\!-\!\!4;1024 (0) \rightarrow 2048 \rightarrow 800$  &${\rm V\!:}~{\rm lr}=1e\!\!-\!\!4;2048 (0) \rightarrow 1024 \rightarrow 200$  &${\rm V\!:}~{\rm lr}=1e\!\!-\!\!4;4096 (0) \rightarrow 2048 \rightarrow 500$  \\
		&&& ${\rm S\!:}~{\rm lr}=1e\!\!-\!\!6;700 \rightarrow 800$ &${\rm S\!:}~{\rm lr}=1e\!\!-\!\!6;700 \rightarrow 200$ &${\rm S\!:}~{\rm lr}=1e\!\!-\!\!6;300 \rightarrow 500$  \\
		&& \multirow{2}{*}{NRC(Hinge)} & ${\rm V\!:}~{\rm lr}=1e\!\!-\!\!4;2048 (0) \rightarrow 2048 \rightarrow 500$  &${\rm V\!:}~{\rm lr}=1e\!\!-\!\!4;2048 (0) \rightarrow 1024 \rightarrow 800$  &${\rm V\!:}~{\rm lr}=1e\!\!-\!\!4;2048 (0) \rightarrow 2048 \rightarrow 500$  \\
		&&& ${\rm S\!:}~{\rm lr}=1e\!\!-\!\!6;500 \rightarrow 800;m=10$ &${\rm S\!:}~{\rm lr}=1e\!\!-\!\!6;700 \rightarrow 200;m=1$ &${\rm S\!:}~{\rm lr}=1e\!\!-\!\!6;500 \rightarrow 800;m=10$  \\
		&& WSE(RankNet) & ${\rm V\!:}~{\rm lr}=1e\!\!-\!\!4;1024(0.5)\rightarrow 2048$ & ${\rm V\!:}~{\rm lr}=1e\!\!-\!\!4;512(0.5)\rightarrow 2048$ & ${\rm V\!:}~{\rm lr}=1e\!\!-\!\!4;1024(0.5)\rightarrow 2048$ \\
		&& WSE(Hinge) & ${\rm V\!:}~{\rm lr}=1e\!\!-\!\!4;1024(0.0)\rightarrow 2048;m=10$ & ${\rm V\!:}~{\rm lr}=1e\!\!-\!\!4;256(0.0)\rightarrow 2048;m=1$ & ${\rm V\!:}~{\rm lr}=1e\!\!-\!\!4;1024(0.5)\rightarrow 2048;m=10$ \\
		&& \multirow{2}{*}{RLR(RankNet)} & ${\rm V\!:}~{\rm lr}=1e\!\!-\!\!4;256 (0.5) \rightarrow 1024 \rightarrow 500$  &${\rm V\!:}~{\rm lr}=1e\!\!-\!\!4;256 (0.5) \rightarrow 2048 \rightarrow 800$  &${\rm V\!:}~{\rm lr}=1e\!\!-\!\!4;256 (0.5) \rightarrow 2048 \rightarrow 800$  \\
		&&& ${\rm S\!:}~{\rm lr}=1e\!\!-\!\!6;500 \rightarrow 500$ &${\rm S\!:}~{\rm lr}=1e\!\!-\!\!6;700 \rightarrow 800$ &${\rm S\!:}~{\rm lr}=1e\!\!-\!\!6;700 \rightarrow 800$  \\
		&& \multirow{2}{*}{RLR(Hinge)} & ${\rm V\!:}~{\rm lr}=1e\!\!-\!\!4;512 (0.5) \rightarrow 2048 \rightarrow 800$  &${\rm V\!:}~{\rm lr}=1e\!\!-\!\!4;512 (0.5) \rightarrow 2048 \rightarrow 800$  &${\rm V\!:}~{\rm lr}=1e\!\!-\!\!4;512 (0.5) \rightarrow 2048 \rightarrow 800$  \\
		&&& ${\rm S\!:}~{\rm lr}=1e\!\!-\!\!6;700 \rightarrow 800;m=10$ &${\rm S\!:}~{\rm lr}=1e\!\!-\!\!6;700 \rightarrow 800;m=10$ &${\rm S\!:}~{\rm lr}=1e\!\!-\!\!6;500 \rightarrow 800;m=10$  \\
		\cline{3-6}
		&& DSP & $C=1,\epsilon=0.1$  & $C=1,\epsilon=0.1$& $C=1,\epsilon=0.1$\\
		&& ConSE & $C=1$ & $C=1$& $C=1$\\
		&& COSTA & $C=1$ & $C=1$& $C=1$\\
		&& Fast0Tag & ${\rm V\!:}~{\rm lr}=1e\!\!-\!\!4;4096 (0.5)\rightarrow 1024$  & ${\rm V\!:}~{\rm lr}=1e\!\!-\!\!4; 8192 (0.5) \rightarrow 2048$ & ${\rm V\!:}~{\rm lr}=1e\!\!-\!\!4;4096 (0.5) \rightarrow 1024$\\
		&& \multirow{2}{*}{Fast0Tag+} & ${\rm V\!:}~{\rm lr}=1e\!\!-\!\!4;8192 (0.5) \rightarrow 2048 \rightarrow 800$  &${\rm V\!:}~{\rm lr}=1e\!\!-\!\!4;8192 (0) \rightarrow 1024 \rightarrow 500$  &${\rm V\!:}~{\rm lr}=1e\!\!-\!\!4;4096 (0.5) \rightarrow 1024 \rightarrow 500$  \\
		&&& ${\rm S\!:}~{\rm lr}=1e\!\!-\!\!6;700 \rightarrow 800$ &${\rm S\!:}~{\rm lr}=1e\!\!-\!\!6;500 \rightarrow 500$ &${\rm S\!:}~{\rm lr}=1e\!\!-\!\!6;500 \rightarrow 500$  \\
		&& \multirow{2}{*}{Ours(RankNet)} & ${\rm V\!:}~{\rm lr}=1e\!\!-\!\!4;512 (0.5) \rightarrow 2048 \rightarrow 800$  &${\rm V\!:}~{\rm lr}=1e\!\!-\!\!4;512 (0.5) \rightarrow 2048 \rightarrow 800$  &${\rm V\!:}~{\rm lr}=1e\!\!-\!\!4;1024 (0.5) \rightarrow 2048 \rightarrow 500$  \\
		&&& ${\rm S\!:}~{\rm lr}=1e\!\!-\!\!6;700 \rightarrow 800$ &${\rm S\!:}~{\rm lr}=1e\!\!-\!\!6;500 \rightarrow 800$ &${\rm S\!:}~{\rm lr}=1e\!\!-\!\!6;700 \rightarrow 500$  \\
		&& \multirow{2}{*}{Ours(Hinge)} & ${\rm V\!:}~{\rm lr}=1e\!\!-\!\!4;512 (0.5) \rightarrow 1024 \rightarrow 500$  &${\rm V\!:}~{\rm lr}=1e\!\!-\!\!4;256 (0.5) \rightarrow 2048 \rightarrow 500$  &${\rm V\!:}~{\rm lr}=1e\!\!-\!\!4;1024 (0.5) \rightarrow 1024 \rightarrow 800$  \\
		&&& ${\rm S\!:}~{\rm lr}=1e\!\!-\!\!6;700 \rightarrow 500$; $m=1$ &${\rm S\!:}~{\rm lr}=1e\!\!-\!\!6;700 \rightarrow 500$; $m=1$ &${\rm S\!:}~{\rm lr}=1e\!\!-\!\!6;500 \rightarrow 800$; $m=1$  \\
		\hline\hline
		
	\end{tabular}
\end{lrbox}
\scalebox{0.5}{\usebox{\tablebox}}
}
\end{table*}

\begin{table*}[htbp]
	{\large
		\centering
		\caption[]{Multi-label zero-shot recognition performance (mean$\pm$SEM\%) of the baseline models and the full model using \textbf{RankNet loss} in three evaluation scenarios under different data split settings. \textbf{Notation}:
			GZSL -- generalized ZSL scenario; KnownA -- Known-action only scenario; UnseenA -- Unseen-action only scenario. Other notations are the same as described in Table \ref{table_hyper}.
		}
		\label{table_baseline_log}
\begin{lrbox}{\tablebox}
	\begin{tabular}{c|c|c|ccccc|ccccc}
		\hline\hline
		\multirow{2}{*}{\textbf{Data Split}} & \multirow{2}{*}{\textbf{Evaluation Scenario}}  & \multirow{2}{*}{\textbf{Model}} &  \multicolumn{5}{c|}{\textbf{Breakfast}} & \multicolumn{5}{c}{\textbf{Charades}}\\ \cline{4-13}
		& & &\textbf{L-MAP} & \textbf{I-MAP}   & \textbf{P}  & \textbf{R}  & \textbf{${\rm F_1}$} & \textbf{L-MAP} & \textbf{I-MAP}   & \textbf{P}  & \textbf{R}  & \textbf{${\rm F_1}$}\\
		
		\hline
		\multirow{15}{*}{IFS}& \multirow{5}{*}{GZSL}
		& RGS&$10.9\pm 0.0$&$15.4\pm 0.0$& $8.8\pm 0.0$& $10.2\pm 0.0$& $9.4\pm 0.0$& $5.9\pm 0.0$ & $8.5\pm 0.0$ &$5.6\pm 0.0$ &$3.2\pm 0.0$ &$4.1\pm 0.0$\\
		&& NRC& $27.9\pm 0.7$ & $49.1\pm 1.1$& $35.1\pm 0.9$& $40.5\pm 1.1$& $37.6\pm 1.0$& $9.2\pm 0.1$ &$20.9\pm 0.7$ &$20.7\pm 0.9$ &$11.9\pm 0.5$ &$15.1\pm 0.7$\\
		&& WSE& $30.2\pm 0.4$ &$50.1\pm 0.4$ &$36.7\pm 0.6$ &$42.4\pm 0.7$ &$39.3\pm 0.6$ &$9.3\pm 0.1$ &$20.6\pm 0.4$ &$20.5\pm 0.5$ &$11.7\pm 0.3$ &$14.9\pm 0.3$\\
		&& RLR& $29.6\pm 0.3$ & $50.4\pm 0.5$& $36.7\pm 0.7$& $42.3\pm 0.8$& $39.3\pm 0.8$& $9.4\pm 0.1$ &$20.7\pm 1.0$ &$21.1\pm 1.4$ &$12.0\pm 0.8$ &$15.3\pm 1.0$\\
		&& Ours& $\mathbf{32.8\pm 0.7}$& $\mathbf{53.5\pm 1.2}$ & $\mathbf{38.6\pm 1.6}$& $\mathbf{44.5\pm 1.9}$& $\mathbf{41.4\pm 1.8}$& $\mathbf{9.7\pm 0.1}$ &$\mathbf{22.4\pm 0.4}$ &$\mathbf{22.7\pm 0.4}$ &$\mathbf{13.0\pm 0.2}$ &$\mathbf{16.5\pm 0.3}$\\
		\cline{2-13}
		&\multirow{5}{*}{KnownA}
		& RGS&$11.4\pm 0.2$&$17.4\pm 0.1$& $9.6\pm 0.2$& $12.9\pm 0.0$& $11.0\pm 0.1$& $6.1\pm 0.1$ & $9.4\pm 0.1$ &$5.8\pm 0.1$ &$4.3\pm 0.0$ &$4.9\pm 0.0$\\
		&& NRC& $30.3\pm 1.0$ & $53.6\pm 0.8$& $35.4\pm 0.7$& $47.5\pm 0.6$& $40.6\pm 0.6$& $10.0\pm 0.2$&$25.4\pm 0.5$ &$23.0\pm 0.6$ &$17.1\pm 0.2$ &$19.6\pm 0.4$\\
		&& WSE& ${32.1\pm 0.9}$ &$55.4\pm 1.0$ &${37.6\pm 0.3}$ &${50.6\pm 0.5}$ &${43.2\pm 0.2}$ &$10.1\pm 0.3$ &$24.6\pm 0.3$ &$22.3\pm 0.3$ &$16.6\pm 0.0$ &$19.0\pm 0.1$\\
		&& RLR& $33.5\pm 1.1$ & $56.3\pm 0.8$& $37.4\pm 1.0$& $50.2\pm 0.8$& $42.8\pm 0.9$& $\mathbf{10.7\pm 0.1}$&$\mathbf{26.7\pm 0.7}$ &$\mathbf{23.9\pm 0.9}$ &$\mathbf{17.8\pm 0.4}$ &$\mathbf{20.4\pm 0.6}$\\
		&&  Ours&$\mathbf{35.3\pm 1.3}$ & $\mathbf{58.0\pm 0.4}$& $\mathbf{38.2\pm 1.6}$& $\mathbf{51.3\pm 1.4}$& $\mathbf{43.8\pm 1.5}$& ${10.5\pm 0.2}$&${26.1\pm 0.3}$ &${23.5\pm 0.5}$ &${17.5\pm 0.1}$ &${20.0\pm 0.2}$\\
		\cline{2-13}
		&\multirow{5}{*}{UnseenA}
		& RGS&$8.5\pm 0.6$& $30.7\pm 0.6$& $6.1\pm 0.6$& $50.0\pm 0.0$& $10.9\pm 0.9$& $5.4\pm 0.0$ & $13.9\pm 0.0$ &$5.1\pm 0.0$ &$12.5\pm 0.0$ &$7.2\pm 0.0$\\
		&& NRC& $17.1\pm 0.7$  & $47.5\pm 2.4$& $8.7\pm 0.8$& $70.8\pm 2.5$& $15.4\pm 1.2$& $6.8\pm 0.4$&$20.4\pm 1.1$ &$8.4\pm 0.7$ &$20.8\pm 1.9$ &$11.9\pm 1.0$\\
		&& WSE&${21.7\pm 1.7}$ &${47.7\pm 3.0}$ &$9.1\pm 1.9$ &$72.4\pm 8.8$ &$16.1\pm 3.2$ &${6.9\pm 0.4}$ &$20.5\pm 1.6$ &$8.5\pm 0.7$ &$21.2\pm 2.1$ &$12.1\pm 1.0$\\
		&& RLR& $13.2\pm 2.5$ & $34.3\pm 7.1$& $7.0\pm 1.6$& $56.6\pm 9.7$& $12.4\pm 2.7$& $5.5\pm 0.4$&$13.4\pm 0.6$ &$4.7\pm 0.4$ &$11.6\pm 0.6$ &$6.7\pm 0.5$\\
		&&  Ours&$\mathbf{22.3\pm 1.1}$ & $\mathbf{53.1\pm 5.8}$& $\mathbf{9.5\pm 1.5}$& $\mathbf{77.2\pm 8.4}$& $\mathbf{16.9\pm 2.5}$& $\mathbf{7.1\pm 0.4}$&$\mathbf{22.4\pm 2.1}$ &$\mathbf{9.5\pm 1.0}$ &$\mathbf{23.5\pm 2.7}$ &$\mathbf{13.5\pm 1.5}$\\
		\hline\hline
		\multirow{15}{*}{LFS}& \multirow{5}{*}{GZSL}
		& RGS&$15.2\pm 1.5$&$17.8\pm 0.4$& $11.3\pm 0.4$& $10.2\pm 0.0$& $10.7\pm 0.2$& $5.2\pm 0.1$ & $7.9\pm 0.1$& $5.1\pm 0.1$ &$3.2\pm 0.0$ &$3.9\pm 0.0$\\
		&& NRC& $21.9\pm 2.2$ & $27.6\pm 1.2$& $19.2\pm 1.0$& $17.3\pm 1.1$& $18.2\pm 1.0$& $8.9\pm 0.2$&$20.5\pm 0.5$ &$21.2\pm 0.8$ &$13.3\pm 0.5$ &$16.4\pm 0.6$\\
		&& WSE& $\mathbf{25.2\pm 1.4}$ & ${31.0\pm 2.7}$ &${23.0\pm 2.0}$ &${20.7\pm 1.4}$ &${21.7\pm 1.6}$ &$9.1\pm 0.0$ &$20.3\pm 0.0$ &$21.0\pm 0.2$ &$13.2\pm 0.2$ &$16.2\pm 0.2$\\
		&& RLR& $22.6\pm 2.1$ & $29.6\pm 1.9$& $22.0\pm 0.7$& $19.8\pm 0.2$& $20.9\pm 0.3$& $\mathbf{9.3\pm 0.2}$&$19.6\pm 0.4$ &$19.8\pm 0.6$ &$12.4\pm 0.4$ &$15.2\pm 0.5$\\
		&&  Ours&${25.0\pm 1.4}$ & $\mathbf{32.6\pm 2.9}$& $\mathbf{23.6\pm 2.1}$& $\mathbf{21.2\pm 1.5}$& $\mathbf{22.3\pm 1.7}$& ${9.2\pm 0.1}$&$\mathbf{20.8\pm 0.3}$ &$\mathbf{21.6\pm 0.7}$ &$\mathbf{13.5\pm 0.4}$ &$\mathbf{16.6\pm 0.5}$\\
		\cline{2-13}
		&\multirow{5}{*}{KnownA}
		& RGS&$14.7\pm 1.9$& $17.8\pm 0.2$& $10.1\pm 0.2$& $12.8\pm 0.0$& $11.3\pm 0.1$& $5.0\pm 0.1$ & $8.0\pm 0.1$ &$4.8\pm 0.1$ &$3.7\pm 0.0$ &$4.2\pm 0.0$\\
		&& NRC& $22.2\pm 2.5$ & $30.2\pm 1.5$& $18.9\pm 1.1$& $24.2\pm 1.8$& $21.2\pm 1.4$& $8.9\pm 0.3$&$22.3\pm 0.6$ &$21.1\pm 0.7$ &$15.9\pm 0.4$ &$18.1\pm 0.5$\\
		&& WSE& $\mathbf{25.3\pm 2.0}$ &${34.2\pm 3.3}$ &${23.1\pm 1.9}$ &${29.5\pm 2.7}$ &${25.9\pm 2.3}$ &$9.0\pm 0.0$ &$22.2\pm 0.3$ &$21.1\pm 0.4$ &$15.9\pm 0.0$ &$18.1\pm 0.1$\\
		&& RLR& $24.8\pm 2.6$ & $34.3\pm 3.1$& $21.7\pm 0.7$& $27.6\pm 0.4$& $24.3\pm 0.6$& $\mathbf{9.5\pm 0.1}$&$\mathbf{23.0\pm 0.4}$ &${21.2\pm 0.5}$ &${16.0\pm 0.4}$ &${18.2\pm 0.4}$\\
		&&  Ours& ${25.2\pm 1.6}$ & $\mathbf{35.7\pm 3.6}$& $\mathbf{23.3\pm 2.0}$& $\mathbf{29.6\pm 2.8}$& $\mathbf{26.0\pm 2.3}$& ${9.1\pm 0.1}$&$\mathbf{23.0\pm 0.6}$ &$\mathbf{21.6\pm 0.7}$ &$\mathbf{16.3\pm 0.3}$ &$\mathbf{18.6\pm 0.5}$\\
		\cline{2-13}
		&\multirow{5}{*}{UnseenA}
		& RGS&$16.8\pm 1.6$& $34.2\pm 1.3$& $16.2\pm 1.6$& $50.0\pm 0.0$& $24.4\pm 1.8$& $6.9\pm 0.1$ & $19.4\pm 0.1$ &$6.7\pm 0.1$ &$25.0\pm 0.0$ &$10.5\pm 0.1$\\
		&& NRC& $21.2\pm 1.4$ & ${42.3\pm 5.4}$& ${18.7\pm 3.2}$& $57.1\pm 7.0$& ${28.0\pm 4.4}$& $9.4\pm 0.1$&$\mathbf{30.7\pm 2.2}$ &$\mathbf{11.6\pm 0.7}$ &$\mathbf{43.2\pm 2.8}$ &$\mathbf{18.2\pm 1.1}$\\
		&& WSE& $\mathbf{25.2\pm 0.4}$ &$42.5\pm 4.6$ &${20.0\pm 3.1}$ &$61.5\pm 6.8$ &${30.1\pm 4.3}$ &$\mathbf{9.7\pm 0.1}$ &$30.1\pm 1.2$ &$11.5\pm 0.4$ &$43.1\pm 1.8$ &$\mathbf{18.2\pm 0.6}$\\
		&& RLR& $16.7\pm 2.5$ & $32.2\pm 2.1$& $15.0\pm 1.9$& $45.8\pm 2.0$& $22.5\pm 2.4$& $7.7\pm 0.5$&$19.3\pm 0.3$ &$6.7\pm 0.1$ &$25.2\pm 0.6$ &$10.6\pm 0.2$\\
		&&  Ours& ${24.6\pm 1.5}$ & $\mathbf{44.8\pm 4.7}$& $\mathbf{22.2\pm 2.5}$& $\mathbf{68.6\pm 4.7}$& $\mathbf{33.4\pm 3.3}$& $\mathbf{9.7\pm 0.1}$&${29.2\pm 2.1}$ &${11.1\pm 0.9}$ &${41.6\pm 3.5}$ &${17.5\pm 1.4}$\\
		\hline\hline
	\end{tabular}
\end{lrbox}
\scalebox{0.55}{\usebox{\tablebox}}
}
\end{table*}

\begin{table*}[htbp]
	{\large
		\centering
		\caption[]{Multi-label zero-shot recognition performance (mean$\pm$SEM\%) of the baseline models and our full model using \textbf{Hinge rank loss} in three evaluation scenarios under different data split settings. \textbf{Notation}:
			GZSL -- generalized ZSL scenario; KnownA -- Known-action only scenario; UnseenA -- Unseen-action only scenario. Other notations are the same as described in Table \ref{table_hyper}.
		}
		\label{table_baseline_hinge}
		\begin{lrbox}{\tablebox}
			\begin{tabular}{c|c|c|ccccc|ccccc}
				\hline\hline
				\multirow{2}{*}{\textbf{Data Split}} & \multirow{2}{*}{\textbf{Evaluation Scenario}}  & \multirow{2}{*}{\textbf{Model}} &  \multicolumn{5}{c|}{\textbf{Breakfast}} & \multicolumn{5}{c}{\textbf{Charades}}\\ \cline{4-13}
				& & &\textbf{L-MAP} & \textbf{I-MAP}   & \textbf{P}  & \textbf{R}  & \textbf{${\rm F_1}$} & \textbf{L-MAP} & \textbf{I-MAP}   & \textbf{P}  & \textbf{R}  & \textbf{${\rm F_1}$}\\
				
				\hline
				\multirow{15}{*}{IFS}& \multirow{5}{*}{GZSL}
				& RGS&$10.9\pm 0.0$&$15.4\pm 0.0$& $8.8\pm 0.0$& $10.2\pm 0.0$& $9.4\pm 0.0$& $5.9\pm 0.0$ & $8.5\pm 0.0$ &$5.6\pm 0.0$ &$3.2\pm 0.0$ &$4.1\pm 0.0$\\
				&& NRC& $28.1\pm 0.4$ & $50.1\pm 1.0$& $36.7\pm 0.7$& $42.3\pm 0.8$& $39.3\pm 0.7$& $9.2\pm 0.1$ &$21.7\pm 0.7$ &$22.2\pm 0.7$ &$12.7\pm 0.4$ &$16.2\pm 0.5$\\
				&& WSE& $31.1\pm 0.7$ &$50.4\pm 0.3$ &$36.6\pm 0.5$ &$42.2\pm 0.6$ &$39.2\pm 0.6$ &$8.6\pm 0.0$ &$20.1\pm 0.5$ &$20.3\pm 0.7$ &$11.6\pm 0.4$ &$14.7\pm 0.5$\\
				&& RLR& $30.3\pm 1.2$ & $51.8\pm 1.8$& $37.4\pm 1.4$& $43.2\pm 1.6$& $40.1\pm 1.5$& $9.5\pm 0.1$ &$21.4\pm 0.9$ &$22.4\pm 0.8$ &$12.8\pm 0.5$ &$16.3\pm 0.6$\\
				&& Ours& $\mathbf{32.7\pm 0.4}$& $\mathbf{53.4\pm 0.8}$ & $\mathbf{38.9\pm 0.5}$& $\mathbf{44.9\pm 0.5}$& $\mathbf{41.7\pm 0.5}$& $\mathbf{10.0\pm 0.1}$ &$\mathbf{22.6\pm 0.4}$ &$\mathbf{23.1\pm 0.5}$ &$\mathbf{13.2\pm 0.3}$ &$\mathbf{16.8\pm 0.4}$\\
				\cline{2-13}
				&\multirow{5}{*}{KnownA}
				& RGS&$11.4\pm 0.2$&$17.4\pm 0.1$& $9.6\pm 0.2$& $12.9\pm 0.0$& $11.0\pm 0.1$& $6.1\pm 0.1$ & $9.4\pm 0.1$ &$5.8\pm 0.1$ &$4.3\pm 0.0$ &$4.9\pm 0.0$\\
				&& NRC& $30.5\pm 0.6$ & $54.9\pm 0.9$& $36.6\pm 0.4$& $49.3\pm 0.4$& $42.0\pm 0.2$& $10.1\pm 0.3$&$25.4\pm 0.5$ &$23.2\pm 0.6$ &$17.2\pm 0.2$ &$19.8\pm 0.4$\\
				&& WSE& ${33.0\pm 0.9}$ &$55.4\pm 0.7$ &${37.1\pm 0.6}$ &${49.9\pm 0.6}$ &${42.6\pm 0.5}$ &$9.3\pm 0.2$ &$23.7\pm 0.4$ &$21.5\pm 0.5$ &$16.0\pm 0.2$ &$18.3\pm 0.3$\\
				&& RLR& $34.9\pm 1.8$ & $\mathbf{59.0\pm 1.3}$& $\mathbf{38.6\pm 0.7}$& $51.9\pm 1.1$& $\mathbf{44.3\pm 0.8}$& ${10.8\pm 0.2}$&$\mathbf{26.5\pm 0.7}$ &${23.7\pm 0.8}$ &${17.7\pm 0.4}$ &${20.3\pm 0.5}$\\
				&&  Ours&$\mathbf{35.6\pm 0.5}$ & ${58.2\pm 1.0}$& $\mathbf{38.6\pm 0.3}$& $\mathbf{52.0\pm 0.5}$& $\mathbf{44.3\pm 0.0}$& $\mathbf{10.9\pm 0.2}$&${26.4\pm 0.3}$ &$\mathbf{24.1\pm 0.3}$ &$\mathbf{17.9\pm 0.2}$ &$\mathbf{20.6\pm 0.1}$\\
				\cline{2-13}
				&\multirow{5}{*}{UnseenA}
				& RGS&$8.5\pm 0.6$& $30.7\pm 0.6$& $6.1\pm 0.6$& $50.0\pm 0.0$& $10.9\pm 0.9$& $5.4\pm 0.0$ & $13.9\pm 0.0$ &$5.1\pm 0.0$ &$12.5\pm 0.0$ &$7.2\pm 0.0$\\
				&& NRC& $17.6\pm 1.6$  & $46.7\pm 3.2$& $\mathbf{9.1\pm 1.2}$& $\mathbf{73.4\pm 4.0}$& $\mathbf{16.1\pm 2.0}$& $6.7\pm 0.4$&$21.8\pm 1.2$ &$9.3\pm 0.4$ &$23.1\pm 1.4$ &$13.2\pm 0.6$\\
				&& WSE&$\mathbf{22.7\pm 2.1}$ &${43.6\pm 6.0}$ &$\mathbf{9.1\pm 1.9}$ &$72.6\pm 8.3$ &$\mathbf{16.1\pm 3.1}$ &${6.6\pm 0.4}$ &$21.3\pm 1.9$ &$8.7\pm 0.7$ &$21.7\pm 2.6$ &$12.4\pm 1.1$\\
				&& RLR& $10.5\pm 2.0$ & $35.5\pm 2.1$& $5.9\pm 0.3$& $48.4\pm 2.1$& $10.5\pm 0.5$& $5.6\pm 0.3$&$14.7\pm 1.9$ &$5.0\pm 0.7$ &$12.4\pm 2.0$ &$7.1\pm 1.1$\\
				&&  Ours&${20.3\pm 0.6}$ & $\mathbf{47.6\pm 1.6}$& ${8.8\pm 1.0}$& ${71.6\pm 2.2}$& ${15.7\pm 1.6}$& $\mathbf{7.3\pm 0.5}$&$\mathbf{22.6\pm 1.5}$ &$\mathbf{9.6\pm 0.7}$ &$\mathbf{23.9\pm 2.0}$ &$\mathbf{13.7\pm 1.0}$\\
				\hline\hline
				\multirow{15}{*}{LFS}& \multirow{5}{*}{GZSL}
				& RGS&$15.2\pm 1.5$&$17.8\pm 0.4$& $11.3\pm 0.4$& $10.2\pm 0.0$& $10.7\pm 0.2$& $5.2\pm 0.1$ & $7.9\pm 0.1$& $5.1\pm 0.1$ &$3.2\pm 0.0$ &$3.9\pm 0.0$\\
				&& NRC& $22.6\pm 2.0$ & $26.6\pm 1.7$& $18.5\pm 1.3$& $16.7\pm 1.2$& $17.5\pm 1.2$& $8.9\pm 0.1$&$20.7\pm 0.1$ &$21.2\pm 0.3$ &$13.3\pm 0.3$ &$16.4\pm 0.3$\\
				&& WSE& ${24.4\pm 1.7}$ & ${31.3\pm 3.0}$ &${23.4\pm 1.9}$ &${21.1\pm 1.4}$ &${22.2\pm 1.6}$ &$8.1\pm 0.2$ &$19.4\pm 0.2$ &$19.4\pm 0.4$ &$12.2\pm 0.3$ &$14.9\pm 0.3$\\
				&& RLR& $22.2\pm 1.4$ & $31.6\pm 1.6$& $24.3\pm 2.1$& $21.8\pm 1.6$& $23.0\pm 1.8$& ${9.0\pm 0.1}$&$19.9\pm 0.7$ &$20.7\pm 1.0$ &$13.0\pm 0.5$ &$16.0\pm 0.6$\\
				&&  Ours&$\mathbf{24.7\pm 1.4}$ & $\mathbf{32.9\pm 2.9}$& $\mathbf{24.6\pm 2.5}$& $\mathbf{22.1\pm 1.9}$& $\mathbf{23.3\pm 2.2}$& $\mathbf{9.2\pm 0.1}$&$\mathbf{21.1\pm 0.4}$ &$\mathbf{22.1\pm 0.7}$ &$\mathbf{13.9\pm 0.4}$ &$\mathbf{17.1\pm 0.5}$\\
				\cline{2-13}
				&\multirow{5}{*}{KnownA}
				& RGS&$14.7\pm 1.9$& $17.8\pm 0.2$& $10.1\pm 0.2$& $12.8\pm 0.0$& $11.3\pm 0.1$& $5.0\pm 0.1$ & $8.0\pm 0.1$ &$4.8\pm 0.1$ &$3.7\pm 0.0$ &$4.2\pm 0.0$\\
				&& NRC& $23.0\pm 2.6$ & $29.4\pm 2.7$& $18.5\pm 1.5$& $23.6\pm 2.2$& $20.7\pm 1.8$& $8.8\pm 0.1$&$22.6\pm 0.4$ &$21.3\pm 0.4$ &$16.1\pm 0.0$ &$18.3\pm 0.1$\\
				&& WSE& ${24.9\pm 2.4}$ &${35.1\pm 3.8}$ &${23.4\pm 2.0}$ &${29.7\pm 2.0}$ &${26.2\pm 2.0}$ &$7.9\pm 0.2$ &$21.1\pm 0.5$ &$19.6\pm 0.4$ &$14.8\pm 0.1$ &$16.9\pm 0.2$\\
				&& RLR& $23.6\pm 2.6$ & $36.1\pm 2.7$& $24.3\pm 2.2$& $31.0\pm 3.2$& $27.2\pm 2.6$& $\mathbf{9.2\pm 0.1}$&${22.7\pm 0.9}$ &${21.0\pm 1.0}$ &${15.9\pm 0.5}$ &${18.1\pm 0.7}$\\
				&&  Ours& $\mathbf{25.2\pm 1.8}$ & $\mathbf{37.1\pm 3.4}$& $\mathbf{25.0\pm 2.0}$& $\mathbf{31.9\pm 2.7}$& $\mathbf{28.1\pm 2.3}$& ${9.1\pm 0.0}$&$\mathbf{23.2\pm 0.4}$ &$\mathbf{22.1\pm 0.5}$ &$\mathbf{16.6\pm 0.2}$ &$\mathbf{19.0\pm 0.3}$\\
				\cline{2-13}
				&\multirow{5}{*}{UnseenA}
				& RGS&$16.8\pm 1.6$& $34.2\pm 1.3$& $16.2\pm 1.6$& $50.0\pm 0.0$& $24.4\pm 1.8$& $6.9\pm 0.1$ & $19.4\pm 0.1$ &$6.7\pm 0.1$ &$25.0\pm 0.0$ &$10.5\pm 0.1$\\
				&& NRC& $21.6\pm 0.6$ & ${39.6\pm 2.8}$& $\mathbf{19.7\pm 1.4}$& $\mathbf{61.4\pm 4.6}$& $\mathbf{29.7\pm 1.8}$& $9.5\pm 0.2$&${30.7\pm 2.3}$ &${11.8\pm 0.7}$ &${44.0\pm 2.8}$ &${18.6\pm 1.1}$\\
				&& WSE& $\mathbf{23.3\pm 0.2}$ &$\mathbf{41.2\pm 4.3}$ &${19.0\pm 2.8}$ &$58.4\pm 5.7$ &${28.6\pm 3.8}$ &${9.3\pm 0.1}$ &$29.9\pm 1.6$ &$11.5\pm 0.2$ &$42.9\pm 1.3$ &${18.1\pm 0.4}$\\
				&& RLR& $18.9\pm 1.1$ & $33.3\pm 1.2$& $16.4\pm 0.1$& $51.5\pm 5.4$& $24.7\pm 0.6$& $7.4\pm 0.5$&$18.5\pm 0.8$ &$6.4\pm 0.5$ &$23.8\pm 1.4$ &$10.1\pm 0.7$\\
				&&  Ours& $\mathbf{23.3\pm 2.0}$ & ${40.2\pm 5.1}$& ${19.4\pm 3.6}$& ${59.0\pm 7.7}$& ${29.1\pm 5.0}$& $\mathbf{10.0\pm 0.1}$&$\mathbf{30.9\pm 2.6}$ &$\mathbf{12.0\pm 0.9}$ &$\mathbf{44.9\pm 3.8}$ &$\mathbf{18.9\pm 1.5}$\\
				\hline\hline
			\end{tabular}
		\end{lrbox}
		\scalebox{0.55}{\usebox{\tablebox}}
	}
\end{table*}

\subsection{Results on Comparison to Baseline Models}
\label{subsect:res_baseline}

Tables \ref{table_baseline_log} and \ref{table_baseline_hinge} summarize all the results yielded by four baseline models described in Section \ref{sect_ablation} and the full model described in Section \ref{sect_jlel}, with the use of regularized RankNet loss and hinge rank loss described in Section \ref{sect_loss} respectively. The experimental results are reported based on two different data split settings described in Section \ref{sect_datasplit}, \emph{instance-first split} (IFS) and \emph{label-first split} (LFS), under three different evaluation scenarios  described in Section \ref{sect_evaluation_scenarios}; i.e., \emph{generalized ZSL}, \emph{known-action only}  and \emph{unseen-action only} scenarios. For reliability, we report the \textit{mean} and \textit{standard error of the mean} (SEM) of results ($k=5$ used in evaluation metrics, i.e., Eqs.(\ref{eq_precision}-\ref{eq_f1})) over three randomly generated known/unseen label splits for each evaluation scenario.

For the IFS setting, it is observed from Tables \ref{table_baseline_log} and \ref{table_baseline_hinge} that all the baseline models and the full model perform significantly better than the RGS, a random guess model, on two datasets regardless of evaluation scenarios apart from the RLR model under the unseen-action only scenario. Due to a lack of knowledge transfer in a random label representation, the zero-shot performance of the RLR is expected. Overall, the full model outperforms all the baseline models on both datasets in the generalized ZSL and unseen-action only scenarios regardless of evaluation metrics. A comparison to the WSE suggests that the performance of the full model is generally superior to this baseline on both datasets under different evaluation scenarios, which lends evidence to support the necessity of the semantic embedding learning in multi-label learning problems.
Also, we observe that the full model outperforms the RLR on Breakfast but fails to do so on Charades in the know-action only scenario when using RankNet loss (Table \ref{table_baseline_log}).
This result suggests that a semantic representation of labels is not critically important for known actions in multi-label learning when the semantic embedding learning has been employed to explore the between-action relations from label co-occurrences.
This observation further implies that the semantic embedding learning cannot explore the semantic relations between labels properly unless there are sufficient training examples for different actions.
Nevertheless, the performance in the unseen-action only and the generalized ZSL scenarios clearly indicates the importance of the semantic representation of an action label for knowledge transfer required by ZSL. It is also evident from Tables \ref{table_baseline_log} and \ref{table_baseline_hinge} that the full model always outperforms the NRC  where there are no recurrent connections. Thus, the comparison to the baseline models clearly suggest that the performance gain is brought by the use of an LSTM layer in the visual model and the semantic embedding learning fulfilled in the semantic model.

For the LFS setting, results shown in Tables \ref{table_baseline_log} and \ref{table_baseline_hinge} suggest that all the baseline models perform significantly better than a random guess. Overall, the full model generally outperforms those baseline models on both datasets. In few circumstances, however, the full model slightly under-performs the WSE on Breakfast in terms of L-MAP with a tiny margin (Table \ref{table_baseline_log}). Besides, it is observed from Table \ref{table_baseline_log} that in the unseen-action only scenario, our model slightly under-performs the WSE and the NRC on Charades although it yields the best performance on  Breakfast. While from Table \ref{table_baseline_hinge} we can observe that our full model performs the best on Charades but not on Breakfast. These results reveal that the two employed rank losses are complementary when learning the joint embedding space.

In summary, the comparison to the elaborated baseline models facilitates the understanding of different components and ranking loss functions employed in our proposed framework for multi-label zero-shot human action recognition. Two different ranking losses used in our framework yield the similar performance overall. By comparison to four baseline models, the full model generally leads to better results on two datasets measured with different evaluation metrics in all three evaluation scenarios, although the experimental results also reveal the limitation of components used in the full model to be investigated in our future studies.

\begin{table*}[!htbp]
	{\large
		\centering
		\caption[]{Multi-label zero-shot recognition performance (mean$\pm$SEM\%) of five state-of-the-art methods and ours with the reference to random guess in three evaluation scenarios under different data split settings. The notations are the same as used in Tables \ref{table_hyper}-\ref{table_baseline_hinge}.}
		\label{table_comparative}
\begin{lrbox}{\tablebox}
	\begin{tabular}{c|c|c|ccccc|ccccc}
		\hline\hline
		\multirow{2}{*}{\textbf{Data Split}} & \multirow{2}{*}{\textbf{Evaluation Scenario}}  & \multirow{2}{*}{\textbf{Model}} &  \multicolumn{5}{c|}{\textbf{Breakfast}} & \multicolumn{5}{c}{\textbf{Charades}}\\ \cline{4-13}
		& & &\textbf{L-MAP} & \textbf{I-MAP}   & \textbf{P}  & \textbf{R}  & \textbf{${\rm F_1}$} & \textbf{L-MAP} & \textbf{I-MAP}   & \textbf{P}  & \textbf{R}  & \textbf{${\rm F_1}$}\\
		
		\hline
		\multirow{21}{*}{IFS}& \multirow{7}{*}{GZSL}
		& RGS &$10.9\pm 0.0$&$15.4\pm 0.0$& $8.8\pm 0.0$& $10.2\pm 0.0$& $9.4\pm 0.0$& $5.9\pm 0.0$ & $8.5\pm 0.0$ &$5.6\pm 0.0$ &$3.2\pm 0.0$ &$4.1\pm 0.0$\\
		&& DSP\citep{lampert2014attribute} & $21.3\pm 0.0$ & $25.4\pm 0.9$& $16.8\pm 0.4$& $19.4\pm 0.4$& $18.0\pm 0.4$& $7.9\pm 0.0$ &$12.5\pm 0.1$ &$12.5\pm 0.2$ &$7.2\pm 0.1$ &$9.1\pm 0.1$\\
		&& ConSE\citep{norouzi2013zero}& $16.1\pm 0.2$ & $29.6\pm 0.3$& $20.3\pm 0.1$& $23.4\pm 0.1$& $21.7\pm 0.1$ &$7.3\pm 0.0$ &$13.9\pm 0.3$ &$14.6\pm 0.5$ &$8.3\pm 0.3$ &$10.6\pm 0.4$\\
		&& COSTA\citep{mensink2014costa}& $19.7\pm 0.2$ & $37.4\pm 0.3$& $28.8\pm 0.4$& $33.2\pm 0.5$& $30.8\pm 0.5$ &$8.5\pm 0.1$ &$16.9\pm 0.7$ &$17.6\pm 0.9$ &$10.1\pm 0.5$ &$12.8\pm 0.6$\\
		&& Fast0Tag\citep{zhang2016fast}& $22.3\pm 1.1$& $38.6\pm 0.2$ & $27.5\pm 0.1$& $31.8\pm 0.1$& $29.5\pm 0.1$& $9.3\pm 0.0$ &$20.6\pm 1.0$ &$20.6\pm 1.4$ &$11.8\pm 0.8$ &$15.0\pm 1.0$\\
		&& Fast0Tag+& $23.4\pm 0.3$& $40.6\pm 0.7$ & $29.2\pm 0.3$& $33.7\pm 0.3$& $31.3\pm 0.3$& $9.7\pm 0.1$ &$21.7\pm 0.6$ &$21.9\pm 0.8$ &$12.5\pm 0.5$ &$15.9\pm 0.6$\\
		&& Ours(RankNet)& $32.1\pm 0.8$& $53.3\pm 1.0$ & ${39.0\pm 0.9}$& ${45.0\pm 1.0}$& ${41.8\pm 0.9}$& ${9.7\pm 0.1}$ &${22.4\pm 0.4}$ &${22.8\pm 0.4}$ &${13.0\pm 0.2}$ &${16.5\pm 0.3}$\\
		&& Ours(Hinge) &${32.7\pm 0.4}$& ${53.4\pm 0.8}$ &$38.9\pm 0.5$ &$44.9\pm 0.5$ &$41.7\pm 0.5$ &$10.0\pm 0.1$ &$22.6\pm 0.4$ &$23.1\pm 0.5$ &$13.2\pm 0.3$ &$16.8\pm 0.4$\\
		&& Ours(Fusion) &$\mathbf{33.9\pm 0.4}$ &$\mathbf{54.7\pm 1.1}$ &$\mathbf{40.0\pm 1.1}$ &$\mathbf{46.1\pm 1.3}$ &$\mathbf{42.8\pm 1.2}$ &$\mathbf{10.1\pm 0.1}$ &$\mathbf{23.3\pm 0.5}$ &$\mathbf{23.7\pm 0.7}$ &$\mathbf{13.5\pm 0.4}$ &$\mathbf{17.2\pm 0.5}$\\
		\cline{2-13}
		&\multirow{7}{*}{KnownA}
		& RGS &$11.4\pm 0.2$&$17.4\pm 0.1$& $9.6\pm 0.2$& $12.9\pm 0.0$& $11.0\pm 0.1$& $6.1\pm 0.1$ & $9.4\pm 0.1$ &$5.8\pm 0.1$ &$4.3\pm 0.0$ &$4.9\pm 0.0$\\
		&& DSP\citep{lampert2014attribute} & $22.6\pm 0.5$ & $30.8\pm 1.9$& $19.1\pm 0.9$& $25.7\pm 1.3$& $21.9\pm 1.1$& $8.5\pm 0.1$&$13.9\pm 0.3$ &$12.4\pm 0.4$ &$9.2\pm 0.3$ &$10.6\pm 0.3$\\
		&& ConSE\citep{norouzi2013zero}& $17.1\pm 0.3$ & $33.0\pm 1.2$& $21.0\pm 0.2$& $28.2\pm 0.7$& $24.1\pm 0.4$ &$7.8\pm 0.2$ &$15.9\pm 0.4$ &$14.6\pm 0.7$ &$10.8\pm 0.3$ &$12.4\pm 0.5$\\
		&& COSTA\citep{mensink2014costa}& $22.1\pm 0.7$ & $41.6\pm 0.7$& $28.8\pm 0.4$& $38.7\pm 0.2$& $33.0\pm 0.3$ &$9.3\pm 0.2$ &$19.7\pm 0.6$ &$17.6\pm 0.9$ &$13.1\pm 0.5$ &$15.0\pm 0.6$\\
		&& Fast0Tag\citep{zhang2016fast}& $23.9\pm 1.3$& $44.3\pm 0.5$ & $29.8\pm 0.7$& $40.0\pm 0.6$& $34.1\pm 0.6$& $10.0\pm 0.2$ &$24.6\pm 0.6$ &$22.4\pm 0.7$ &$16.6\pm 0.3$ &$19.1\pm 0.4$\\
		&& Fast0Tag+& $25.4\pm 0.2$& $45.0\pm 0.5$ & $29.9\pm 0.4$& $40.2\pm 1.1$& $34.3\pm 0.6$& $10.5\pm 0.3$ &$25.7\pm 0.7$ &$23.3\pm 0.9$ &$17.3\pm 0.4$ &$19.8\pm 0.6$\\
		&&  Ours(RankNet)&$34.5\pm 1.0$ & ${57.8\pm 0.7}$& ${38.5\pm 1.0}$& ${51.8\pm 0.6}$& ${44.2\pm 0.9}$& ${10.5\pm 0.2}$&${26.1\pm 0.3}$ &${23.5\pm 0.5}$ &${17.5\pm 0.1}$ &${20.0\pm 0.2}$\\
		&& Ours(Hinge) &$\mathbf{35.6\pm 0.5}$ &${58.2\pm 1.0}$ &${38.6\pm 0.3}$ &${52.0\pm 0.5}$ &${44.3\pm 0.0}$ &$10.9\pm 0.2$ &$26.4\pm 0.3$ &$24.1\pm 0.3$ &$17.9\pm 0.2$ &$20.6\pm 0.1$\\
		&& Ours(Fusion) &$\mathbf{36.6\pm 0.7}$ &$\mathbf{59.4\pm 0.5}$ &$\mathbf{39.6\pm 0.9}$ &$\mathbf{53.2\pm 0.5}$ &$\mathbf{45.4\pm 0.8}$ &$\mathbf{11.0\pm 0.3}$ &$\mathbf{27.1\pm 0.4}$ &$\mathbf{24.6\pm 0.5}$ &$\mathbf{18.3\pm 0.2}$ &$\mathbf{21.0\pm 0.3}$\\
		\cline{2-13}
		&\multirow{7}{*}{UnseenA}
		& RGS &$8.5\pm 0.6$& $30.7\pm 0.6$& $6.1\pm 0.6$& $50.0\pm 0.0$& $10.9\pm 0.9$& $5.4\pm 0.0$ & $13.9\pm 0.0$ &$5.1\pm 0.0$ &$12.5\pm 0.0$ &$7.2\pm 0.0$\\
		&& DSP\citep{lampert2014attribute} & $15.9\pm 1.5$ & $27.0\pm 4.7$& $5.8\pm 1.6$& $47.6\pm 11.5$& $10.4\pm 2.7$& $6.2\pm 0.4$&$17.3\pm 1.4$ &$7.4\pm 0.8$ &$18.4\pm 1.9$ &$10.6\pm 1.1$\\
		&& ConSE\citep{norouzi2013zero}& $12.2\pm 0.3$ & $29.9\pm 5.0$& $6.2\pm 0.9$& $51.0\pm 6.9$& $11.1\pm 1.5$ &$5.8\pm 0.4$ &$17.3\pm 0.8$ &$7.0\pm 0.7$ &$17.2\pm 1.1$ &$9.9\pm 0.8$\\
		&& COSTA\citep{mensink2014costa}& $9.2\pm 1.2$ & $37.4\pm 2.8$& $7.4\pm 0.7$& $60.1\pm 2.5$& $13.1\pm 1.0$ &$6.0\pm 0.3$ &$15.5\pm 1.0$ &$6.3\pm 0.8$ &$15.4\pm 1.5$ &$8.9\pm 1.0$\\
		&& Fast0Tag\citep{zhang2016fast}& $15.3\pm 0.9$& $36.7\pm 4.1$ & $7.0\pm 1.4$& $55.9\pm 6.7$& $12.4\pm 2.3$& $7.1\pm 0.4$ &$20.2\pm 2.4$ &$8.3\pm 0.8$ &$20.8\pm 2.7$ &$11.9\pm 1.2$\\
		&& Fast0Tag+& $15.1\pm 1.2$& $39.4\pm 1.3$ & $7.4\pm 0.9$& $60.1\pm 3.6$& $13.2\pm 1.5$& $\mathbf{7.3\pm 0.4}$ &$19.3\pm 0.5$ &$8.1\pm 0.3$ &$20.0\pm 0.6$ &$11.5\pm 0.3$\\
		&&  Ours(RankNet)&${21.9\pm 0.3}$ & ${51.0\pm 4.5}$& ${9.4\pm 1.2}$& ${76.5\pm 6.8}$& ${16.7\pm 2.0}$& ${7.1\pm 0.4}$&${22.4\pm 2.1}$ &${9.5\pm 1.0}$ &${23.5\pm 2.7}$ &${13.5\pm 1.5}$\\
		&& Ours(Hinge) &$20.3\pm 0.6$ &$47.6\pm 1.6$ &$8.8\pm 1.0$ &$71.6\pm 2.2$ &$15.7\pm 1.6$ &$\mathbf{7.3\pm 0.5}$ &$22.6\pm 1.5$ &$9.6\pm 0.7$ &$23.9\pm 2.0$ &$13.7\pm 1.0$\\
		&& Ours(Fusion) &$\mathbf{22.3\pm0.4}$ &$\mathbf{52.9\pm 4.6}$ &$\mathbf{9.7\pm 1.6}$ &$\mathbf{78.8\pm 6.8}$ &$\mathbf{17.3\pm 2.6}$ &$\mathbf{7.3\pm 0.5}$ &$\mathbf{23.1\pm 1.8}$ &$\mathbf{9.9\pm 0.8}$ &$\mathbf{24.6\pm 2.5}$ &$\mathbf{14.1\pm 1.2}$\\
		\hline\hline
		\multirow{21}{*}{LFS}& \multirow{7}{*}{GZSL}
		& RGS &$15.2\pm 1.5$&$17.8\pm 0.4$& $11.3\pm 0.4$& $10.2\pm 0.0$& $10.7\pm 0.2$& $5.2\pm 0.1$ & $7.9\pm 0.1$& $5.1\pm 0.1$ &$3.2\pm 0.0$ &$3.9\pm 0.0$\\
		&& DSP\citep{lampert2014attribute} & $20.7\pm 1.7$ & $18.6\pm 1.9$& $11.0\pm 1.5$& $9.9\pm 1.0$& $10.4\pm 1.2$& $7.4\pm 0.1$&$12.1\pm 0.3$ &$12.4\pm 0.6$ &$7.7\pm 0.3$ &$9.5\pm 0.4$\\
		&& ConSE\citep{norouzi2013zero}& $18.5\pm 2.1$ & $20.2\pm 1.8$& $12.7\pm 1.1$& $11.4\pm 0.6$& $12.0\pm 0.8$ &$7.0\pm 0.1$ &$13.8\pm 0.1$ &$14.8\pm 0.1$ &$9.3\pm 0.1$ &$11.4\pm 0.1$\\
		&& COSTA\citep{mensink2014costa}& $19.3\pm 2.1$ & $22.7\pm 2.1$& $16.8\pm 1.0$& $15.1\pm 0.7$& $15.9\pm 0.8$ &$8.9\pm 0.1$ &$17.3\pm 0.1$ &$18.6\pm 0.1$ &$11.7\pm 0.2$ &$14.4\pm 0.1$\\
		&& Fast0Tag\citep{zhang2016fast}& $22.5\pm 1.5$& $24.3\pm 1.7$ & $16.2\pm 0.6$& $14.6\pm 0.1$& $15.4\pm 0.3$& $8.6\pm 0.1$ &$20.1\pm 0.4$ &$20.1\pm 0.9$ &$12.6\pm 0.6$ &$15.5\pm 0.7$\\
		&& Fast0Tag+& $21.9\pm 1.1$& $23.3\pm 1.1$ & $15.3\pm 0.2$& $13.8\pm 0.6$& $14.5\pm 0.4$& $9.0\pm 0.1$ &$20.9\pm 0.3$ &$21.4\pm 0.6$ &$13.5\pm 0.4$ &$16.5\pm 0.5$\\
		&&  Ours(RankNet)&${25.0\pm 1.4}$ & ${32.6\pm 2.9}$& ${23.6\pm 2.1}$& ${21.2\pm 1.5}$& ${22.3\pm 1.7}$& ${9.2\pm 0.1}$&${20.8\pm 0.3}$ &${21.6\pm 0.7}$ &${13.5\pm 0.4}$ &${16.6\pm 0.5}$\\
		&& Ours(Hinge) &$24.7\pm 1.4$ &$32.9\pm 2.9$ &$\mathbf{24.6\pm 2.5}$ &$\mathbf{22.1\pm 1.9}$ &$\mathbf{23.3\pm 2.2}$ &$9.2\pm 0.1$ &$21.1\pm 0.4$ &$22.1\pm 0.7$ &$13.9\pm 0.4$ &$17.1\pm 0.5$\\
		&& Ours(Fusion) &$\mathbf{25.5\pm 1.4}$ &$\mathbf{33.3\pm 2.5}$ &$\mathbf{24.6\pm 2.3}$ &$\mathbf{22.1\pm 1.6}$ &$23.2\pm 1.9$  &$\mathbf{9.6\pm 0.1}$ &$\mathbf{21.5\pm 0.4}$ &$\mathbf{22.6\pm 0.6}$ &$\mathbf{14.2\pm 0.4}$ &$\mathbf{17.4\pm 0.4}$\\
		\cline{2-13}
		&\multirow{7}{*}{KnownA}
		& RGS &$14.7\pm 1.9$& $17.8\pm 0.2$& $10.1\pm 0.2$& $12.8\pm 0.0$& $11.3\pm 0.1$& $5.0\pm 0.1$ & $8.0\pm 0.1$ &$4.8\pm 0.1$ &$3.7\pm 0.0$ &$4.2\pm 0.0$\\
		&& DSP\citep{lampert2014attribute} & $14.1\pm 1.2$ & $18.1\pm 2.7$& $8.5\pm 1.8$& $10.3\pm 2.1$& $9.4\pm 1.9$& $7.3\pm 0.1$&$12.9\pm 0.4$ &$12.3\pm 0.7$ &$9.3\pm 0.3$ &$10.6\pm 0.5$\\
		&& ConSE\citep{norouzi2013zero}& $12.2\pm 0.8$ & $22.1\pm 0.6$& $13.1\pm 0.6$& $14.6\pm 1.2$& $13.8\pm 0.9$ &$6.8\pm 0.1$ &$15.0\pm 0.1$ &$14.7\pm 0.2$ &$11.1\pm 0.1$ &$12.6\pm 0.1$\\
		&& COSTA\citep{mensink2014costa}& $19.8\pm 2.9$ & $25.1\pm 2.7$& $16.8\pm 1.0$& $21.3\pm 0.8$& $18.7\pm 0.9$ &$9.0\pm 0.1$ &$19.5\pm 0.1$ &$18.6\pm 0.1$ &$14.1\pm 0.2$ &$16.0\pm 0.1$\\
		&& Fast0Tag\citep{zhang2016fast}& $23.1\pm 2.0$& $26.5\pm 1.9$ & $15.9\pm 0.4$& $20.2\pm 0.8$& $17.8\pm 0.5$& $8.4\pm 0.1$ &$22.4\pm 0.7$ &$20.8\pm 0.7$ &$15.7\pm 0.2$ &$17.9\pm 0.4$\\
		&& Fast0Tag+& $22.2\pm 1.7$& $24.9\pm 1.5$ & $15.1\pm 0.6$& $19.3\pm 1.1$& $17.0\pm 0.8$& $9.0\pm 0.1$ &$22.8\pm 0.5$ &$21.3\pm 0.5$ &$16.1\pm 0.1$ &$18.3\pm 0.3$\\
		&&  Ours(RankNet)& ${25.2\pm 1.6}$ & ${35.7\pm 3.6}$& ${23.3\pm 2.0}$& ${29.6\pm 2.8}$& ${26.0\pm 2.3}$& ${9.1\pm 0.1}$&${23.0\pm 0.6}$ &${21.6\pm 0.7}$ &${16.3\pm 0.3}$ &${18.6\pm 0.5}$\\
		&& Ours(Hinge) &${25.2\pm 1.8}$ &$37.1\pm 3.4$ &$\mathbf{25.0\pm 2.0}$ &$\mathbf{31.9\pm 2.7}$ &$\mathbf{28.1\pm 2.3}$ &$9.1\pm 0.0$ &$23.2\pm 0.4$ &$22.1\pm 0.5$ &$16.6\pm 0.2$ &$19.0\pm 0.3$\\
		&& Ours(Fusion) &$\mathbf{25.8\pm 1.8}$ &$\mathbf{37.4\pm 2.7}$ &$24.6\pm 2.0$ &$31.3\pm 2.8$ &$27.6\pm 2.3$ &$\mathbf{9.5\pm 0.1}$ &$\mathbf{23.8\pm 0.6}$ &$\mathbf{22.7\pm 0.7}$ &$\mathbf{17.1\pm 0.3}$ &$\mathbf{19.5\pm 0.4}$\\
		\cline{2-13}
		&\multirow{7}{*}{UnseenA}
		& RGS &$16.8\pm 1.6$& $34.2\pm 1.3$& $16.2\pm 1.6$& $50.0\pm 0.0$& $24.4\pm 1.8$& $6.9\pm 0.1$ & $19.4\pm 0.1$ &$6.7\pm 0.1$ &$25.0\pm 0.0$ &$10.5\pm 0.1$\\
		&& DSP\citep{lampert2014attribute} & $16.6\pm 0.8$ & $28.8\pm 1.4$& $14.7\pm 0.6$& $42.8\pm 1.3$& $21.9\pm 0.7$& $8.5\pm 0.2$&$22.8\pm 0.9$ &$8.1\pm 0.4$ &$30.3\pm 1.3$ &$12.8\pm 0.6$\\
		&& ConSE\citep{norouzi2013zero}& $13.7\pm 1.7$ & $33.0\pm 0.7$& $14.7\pm 0.5$& $50.9\pm 4.6$& $22.7\pm 0.2$ &$7.9\pm 0.4$ &$23.2\pm 0.4$ &$8.5\pm 0.0$ &$31.9\pm 0.3$ &$13.5\pm 0.1$\\
		&& COSTA\citep{mensink2014costa}& $18.2\pm 1.1$ & $35.8\pm 1.2$& $16.9\pm 1.7$& $52.2\pm 2.0$& $25.4\pm 2.0$ &$7.6\pm 0.2$ &$24.9\pm 1.0$ &$9.2\pm 0.4$ &$34.4\pm 1.8$ &$14.5\pm 0.7$\\
		&& Fast0Tag\citep{zhang2016fast}& $21.1\pm 1.1$& $39.1\pm 4.0$ & $18.5\pm 3.6$& $56.0\pm 6.5$& $27.8\pm 4.9$& ${9.5\pm 0.1}$ &$28.4\pm 2.5$ &$11.0\pm 0.5$ &$41.1\pm 2.2$ &$17.4\pm 0.8$\\
		&& Fast0Tag+& $21.2\pm 0.8$& $\mathbf{45.5\pm 3.4}$ & $19.9\pm 1.8$& $61.7\pm 2.4$& $30.0\pm 2.1$& $9.3\pm 0.1$ &${30.2\pm 2.9}$ &${11.6\pm 0.7}$ &${43.2\pm 3.1}$ &${18.2\pm 1.2}$\\
		&&  Ours(RankNet)& ${24.6\pm 1.5}$ & $44.8\pm 4.7$& $\mathbf{22.2\pm 2.5}$& $\mathbf{68.6\pm 4.7}$& $\mathbf{33.4\pm 3.3}$& ${9.7\pm 0.1}$&${29.2\pm 2.1}$ &${11.1\pm 0.9}$ &${41.6\pm 3.5}$ &${17.5\pm 1.4}$\\
		&& Ours(Hinge) &$23.3\pm 2.0$ &$40.2\pm 5.1$ &$19.4\pm 3.6$ &$59.0\pm 7.7$ &$29.1\pm 5.0$ &$10.0\pm 0.1$ &$30.9\pm 2.6$ &$12.0\pm 0.9$ &$44.9\pm 3.8$ &$18.9\pm 1.5$\\
		&& Ours(Fusion) &$\mathbf{24.7\pm 1.5}$ &$42.6\pm 6.1$ &$19.6\pm 3.9$ &$59.7\pm 8.6$ &$29.5\pm 5.4$ &$\mathbf{10.2\pm 0.1}$ &$\mathbf{31.1\pm 2.7}$ &$\mathbf{12.2\pm 0.9}$ &$\mathbf{45.6\pm 3.7}$ &$\mathbf{19.2\pm 1.4}$\\
		\hline\hline
	\end{tabular}
\end{lrbox}
\scalebox{0.55}{\usebox{\tablebox}}
}
\end{table*}

\begin{table*}[!htbp]
	{\large
		\centering
		\caption[]{Multi-label recognition performance (mean$\pm$std\%) of five state-of-the-art methods and ours.\\
		}
		\label{table_conventional}
\begin{lrbox}{\tablebox}
	\begin{tabular}{c|ccccc|ccccc}
		\hline\hline
		\multirow{2}{*}{\textbf{Model}} &  \multicolumn{5}{c|}{\textbf{Breakfast}} & \multicolumn{5}{c}{\textbf{Charades}}\\ \cline{2-11}
		&\textbf{L-MAP} & \textbf{I-MAP}   & \textbf{P}  & \textbf{R}  & \textbf{F1} & \textbf{L-MAP} & \textbf{I-MAP}   & \textbf{P}  & \textbf{R}  & \textbf{F1}\\
		
		\hline
		DSP\citep{lampert2014attribute} & $21.6\pm 0.0$ & $25.2\pm 0.0$& $16.6\pm 0.0$& $19.1\pm 0.0$& $17.7\pm 0.0$& $8.0\pm 0.0$ &$12.6\pm 0.0$ &$13.0\pm 0.0$ &$7.4\pm 0.0$ &$9.5\pm 0.0$\\
		ConSE\citep{norouzi2013zero}& $16.7\pm 0.0$ & $30.8\pm 0.0$& $21.1\pm 0.0$& $24.4\pm 0.0$& $22.7\pm 0.0$ &$7.4\pm 0.0$ &$14.6\pm 0.0$ &$15.8\pm 0.0$ &$9.0\pm 0.0$ &$11.5\pm 0.0$\\
		COSTA\citep{mensink2014costa}& $21.5\pm 0.0$ & $39.3\pm 0.0$& $29.5\pm 0.0$& $34.0\pm 0.0$& $31.6\pm 0.0$ &$9.1\pm 0.0$ &$18.6\pm 0.0$ &$19.5\pm 0.0$ &$11.2\pm 0.0$ &$14.2\pm 0.0$\\
		Fast0Tag\citep{zhang2016fast}& $21.6\pm 1.5$& $40.7\pm 0.6$ & $28.7\pm 0.6$& $33.1\pm 0.7$& $30.7\pm 0.6$& $9.5\pm 0.0$ &$23.4\pm 0.0$ &$23.7\pm 0.1$ &$13.6\pm 0.1$ &$17.3\pm 0.1$\\
		Fast0Tag+& $23.2\pm 0.3$& $42.1\pm 0.5$ & $30.4\pm 0.3$& $35.1\pm 0.3$& $32.6\pm 0.3$& $10.0\pm 0.0$ &$24.3\pm 0.2$ &$24.6\pm 0.3$ &$14.1\pm 0.2$ &$17.9\pm 0.2$\\
		Ours (RankNet)& ${32.7\pm 0.4}$& ${54.0\pm 0.5}$ & ${39.1\pm 0.5}$& ${45.1\pm 0.5}$& ${41.9\pm 0.5}$& ${10.2\pm 0.1}$ &${24.9\pm 0.4}$ &${25.5\pm 0.4}$ &${14.6\pm 0.3}$ &${18.5\pm 0.3}$\\
		Ours (Hinge)& ${33.8\pm 0.2}$& ${55.0\pm 0.3}$ & $\mathbf{39.7\pm 0.2}$& $\mathbf{45.8\pm 0.2}$& $\mathbf{42.5\pm 0.2}$& ${10.5\pm 0.1}$ &${25.2\pm 0.1}$ &${25.5\pm 0.3}$ &${14.6\pm 0.2}$ &${18.6\pm 0.2}$\\
		Ours(Fusion)& $\mathbf{34.1\pm 0.2}$& $\mathbf{55.2\pm 0.2}$ & $\mathbf{39.7\pm 0.3}$& $\mathbf{45.8\pm 0.4}$& $\mathbf{42.5\pm 0.4}$& $\mathbf{10.8\pm 0.1}$ &$\mathbf{25.8\pm 0.1}$ &$\mathbf{26.3\pm 0.2}$ &$\mathbf{15.0\pm 0.1}$ &$\mathbf{19.1\pm 0.1}$\\
		\hline\hline
	\end{tabular}
\end{lrbox}
\scalebox{0.55}{\usebox{\tablebox}}
}
\end{table*}

\subsection{Results on Comparison to State-of-the-Art Methods}
\label{subsect:res_comp}

Table \ref{table_comparative} summarizes the experimental results of the comparative study described in Section \ref{sect_comparative}. Multi-label ZSL performance of five different methods including Fast0Tag+ (our extension for Fast0Tag) with the reference to a random guess is reported to be compared with our proposed framework where two different rank losses and their fusion are employed, respectively. Again, all the experiments are conducted with two different data split settings and evaluated under three evaluation scenarios, as described in Section \ref{subsect:res_baseline}.
For reliability, we report the mean and the SEM of results ($k=5$ used in evaluation metrics, i.e., Eqs(\ref{eq_precision}-\ref{eq_f1})) over three randomly generated known/unseen label splits under each evaluation scenario.

For the IFS setting, it is seen from Table \ref{table_comparative} that all the models perform better than random guess in most of evaluation scenarios. However, DSP and ConSE result in the poorer performance than random guess in the unseen-action only scenario on Breakfast in terms of some specific metric, e.g., I-MAP. Overall, DSP and ConSE under-perform other methods considerably in terms of all five evaluation metrics under all three evaluation scenarios. Such results demonstrate that simply combining semantic representations of co-occurred multiple labels into one collective representation leads to catastrophic loss of semantic information, which is mainly responsible for the poor performance of DSP and ConSE in multi-label recognition.  Fast0Tag generally outperforms COSTA on two datasets in terms of most of evaluation metrics.  While COSTA learns a classifier for each label separately without considering a relationship among co-occurred labels, the consideration of such a relationship in Fast0Tag accounts for the better performance.
By incorporating the semantic embedding learning into Fast0Tag, Fast0Tag+, our extension of Fast0Tag, constantly improves the performance of its original version in most circumstances on two datasets.  Once again, this result lends us evidence to justify the necessity of semantic embedding learning used in our framework for zero-shot multi-label ZSL. In contrast, our model trained with either RankNet loss or the hinge rank loss generally outperforms all five models significantly in terms of five evaluation metrics under different evaluation scenarios on two datasets, as highlighted with bold-font in Table \ref{table_comparative}. By comparing our model to Fast0Tag+, we see three main differences between them as follows: \emph{visual representations}, \emph{network architectures} for the visual model and \emph{loss} functions.
Regarding visual representations, our model uses the segment-based visual features for an instance while Fast0Tag+ employs an instance-level holistic visual representation. For network architectures, we employ an LSTM layer with recurrent connections to capture temporal coherence among segments of a video clip while Fast0Tag+ simply uses a feed-forward network. As described in Section \ref{sect_loss}, we use an alternative loss function to that in Fast0Tag. Thus, those differences together leverage our performance gain over Fast0Tag+, which yet again lends us evidence to support our proposed framework. Finally,
it is observed from Table \ref{table_comparative} that in the IFS setting, the RankNet and the hinge rank losses perform differently on two datasets; the hinge rank loss generally outperforms the RankNet loss on both datasets with the exceptions of unseen action only scenarios on Breakfast. Nevertheless, the fusion of two models trained with different losses leads to the best performance in most circumstances as highlighted with bold-font in Table \ref{table_comparative}. Such results reveal that two losses behave quite differently and the diversity can be exploited via fusion, which provides useful information to develop more effective rank loss functions.

\begin{figure*}[htbp]
		\includegraphics[width=\linewidth]{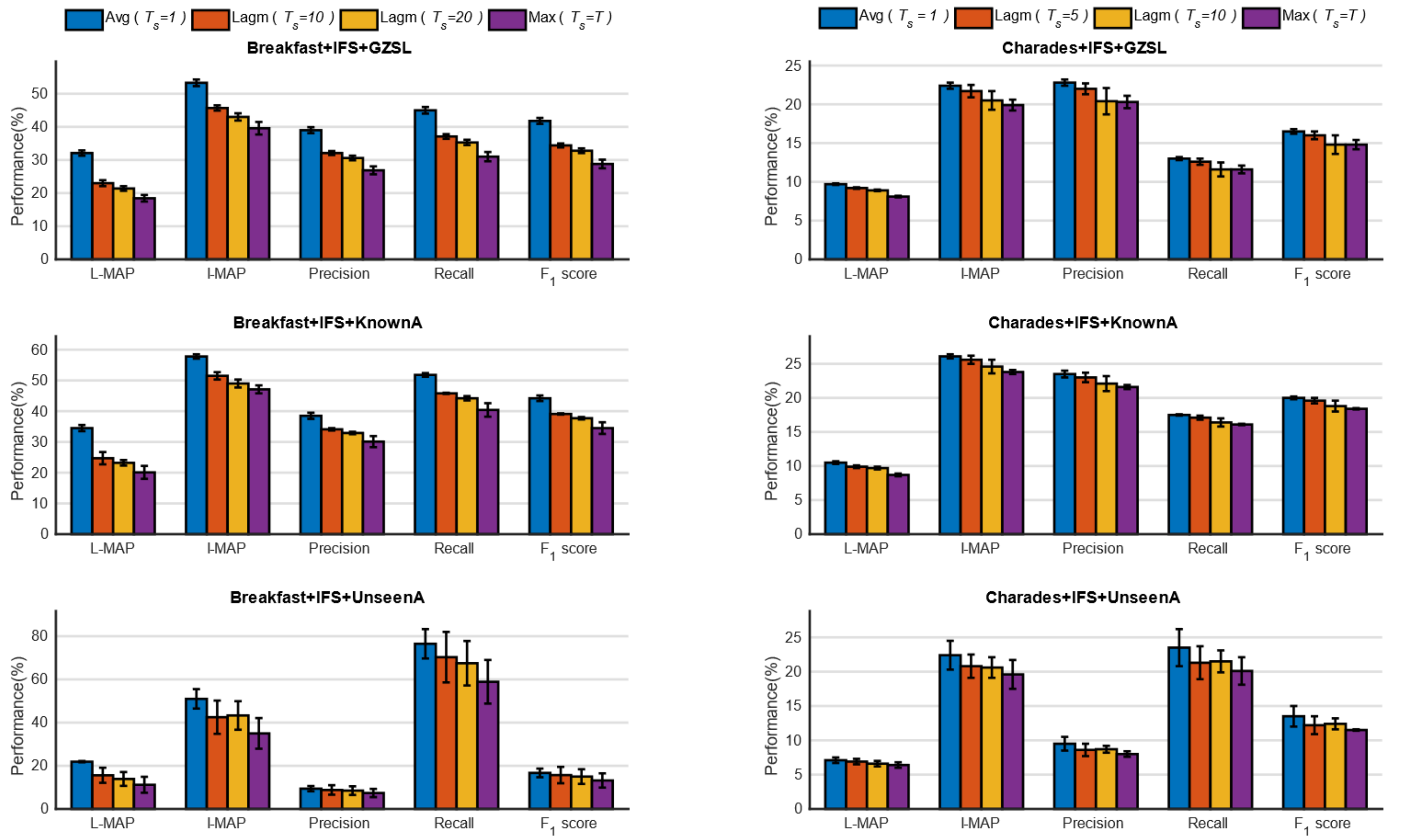}
		\caption{Performance of different pooling strategies used in our framework. Avg: average pooling; Lagm: local average global maximum pooling; Max: maximum pooling. }
		\label{fig_pooling}
		
\end{figure*}

For the LFS setting, experimental results suggest that most of the models in question have similar behavior to that in the IFS setting, as shown in Table \ref{table_comparative}. Once again, DSP and ConSE generally perform worse than other models and even under-perform random guess on Charades in the unseen-action only scenario. While COSTA yields better performance than DSP and ConSE overall, it generally under-performs Fast0Tag, Fast0Tag+ and ours in all three evaluation scenarios. In the LFS setting, our model trained with different rank losses generally outperforms others in most circumstances except for the unseen-action only scenario on Breakfast where Fast0Tag+ performs better than ours marginally in terms of I-MAP.
Regarding two rank losses used in our experiments, the hinge rank loss marginally outperforms the RankNet loss in most circumstances on two datasets. Once again, the fusion of results brought by two rank losses further improves the performance in most circumstances, which provides the further evidence on the complementary aspect of two different rank losses.
As described in Section \ref{subsect:res_baseline}, the LFS setting is more challenging than the IFS setting and some salient visual features on test instances corresponding to unseen actions could completely miss in training examples. In this case, the use of a segment-level based visual representation and an LSTM layer in the visual model may not be able to generalize well due to a lack of training examples. Although such a result does not sufficiently favour the use of a segment-level based visual representation and an LSTM layer in the visual model in the presence of limited training data, it is no doubt that introducing a semantic model to Fast0Tag leverages the performance gain. Once again, experimental results here along with those compared to the baseline models under our LFS setting reveal a \emph{training data sparsity} issue that has to be addressed in any future multi-label zero-shot human action recognition study.

Furthermore, Table \ref{table_conventional} shows the experimental results in conventional multi-label human action recognition, i.e., all the actions are known in learning. In this circumstance, only the IFS setting is applicable. Hence, we use the same IFS setting as described in Section \ref{sect_datasplit} but, unlike what has been done for simulating a zero-shot scenario, do not reserve any actions. Also we use the same procedure as done for zero-shot learning to search for optimal hyper-parameters for five models and ours and repeat the experiments on the same data split as the IFS setting for three trials with different parameter initialization. As a result, we report the mean and standard deviation (std) of three-trial results yielded by different methods. It is evident from Table \ref{table_conventional} that our model trained with either of two rank losses as well as their fusion outperform others in conventional multi-label recognition on both datasets. Without unseen classes, our model trained with the hinge rank loss generally performs slightly better than its counterpart trained with the RankNet loss. Once again, the fusion of results generated by those two models leads to the best performance. To see the degraded performance in a zero-shot scenario, we can compare the performance in the generalized ZSL evaluation scenario under the IFS setting, as shown in Table \ref{table_comparative}, to that reported in Table \ref{table_conventional}. By such a comparison, it is seen that the zero-shot performance of our model drops with a narrow margin (approximately less than 10\% overall in terms of five different evaluation metrics). Given the fact that 10 out of 49 and 40 out of 157 human actions are reserved as unseen labels on Breakfast and Charades, respectively, this comparison on experimental results suggests that our proposed framework yields the promising performance for multi-label zero-shot human action recognition, which is close to the performance in multi-label human action recognition. Experimental results shown in Table \ref{table_conventional} also suggest that other state-of-the-art methods behave similarly to ours in general.
However, we also observe an unusual phenomenon from their performance; i.e., by a comparison to the generalized ZSL performance reported in
Table \ref{table_comparative}, DSP yields slightly worse performance in multi-label recognition in terms of four of five evaluation metrics on Breakfast and so do Fast0Tag and Fast0Tag+ in terms of L-MAP. By a closer look at the dataset and results in two experiments as well as our analysis, we find that at least two factors account for this unusual phenomenon: a) co-occurred labels associated with most of video clips on Breakfast are redundant in light of semantics, and b) the single collective semantic representation of co-occurred multiple labels used in DSP is insensitive to missing of few co-occurred labels due to the label information redundancy and the information loss resulting from the average operation in forming the single representation. Thus, we reckon that this phenomenon is rather specific to the nature of this dataset and the ZSL setting where there are only a small number of unseen labels.

\subsection{Results on Pooling Strategy}
	
 We report the performance of three pooling strategies in terms of five evaluation criteria. It is evident from Figure \ref{fig_pooling} the average pooling always performs the best and the maximum pooling performs the worst regardless of evaluation criteria. In addition, the local average global maximum pooling performs better when $T_s$ is set smaller. Such results imply that our framework interprets the visual information at a global level that tends to recognize actions appearing in a video clip rather than a local level that identifies the accurate boundaries between different actions. From our empirical study, it is observed that the average pooling takes into account all information in a video to yield the relatedness scores while the maximum pooling uses only the local information regarding an abrupt change in visual domain but likely overlooks a large portion of useful information related to the nature of actions.
 Nevertheless, the maximum pooling might be beneficial for unsupervised action localization in  the weak supervision setting, which is beyond the scope of this paper but worth studying in future.

In summary, our comparative study suggests that our proposed framework yields the favourable results and outperforms the existing state-of-the-art methods in general. The average pooling generally outperforms other alternatives in question.
Also, our experimental results demonstrate challenges in multi-label ZSL via our novel LFS setting especially when training data are less correlated to test instances associated with unseen classes in both semantic and visual domains.

\begin{figure*}[htbp]	
	\includegraphics[width=\linewidth]{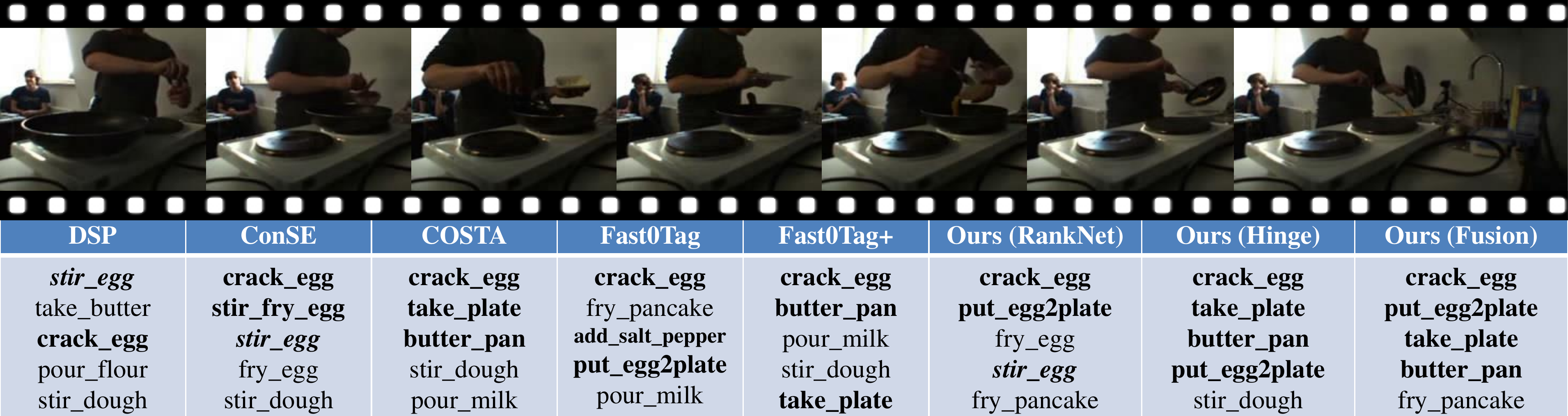}
	\caption{A test video clip in the IFS setting and the top-5 labels predicted by different methods. Its ground-truth labels are \textbf{take\_bowl}, \textbf{crack\_egg}, \textbf{put\_egg2plate}, \textbf{take\_plate},\textit{\textbf{stir\_egg}}, \textbf{pour\_egg2pan}, \textbf{stir\_fry\_egg}, \textbf{add\_salt\_pepper}, \textbf{butter\_pan}.}
	\label{fig_video275}
\end{figure*}

\begin{figure*}[htbp]
	\includegraphics[width=\linewidth]{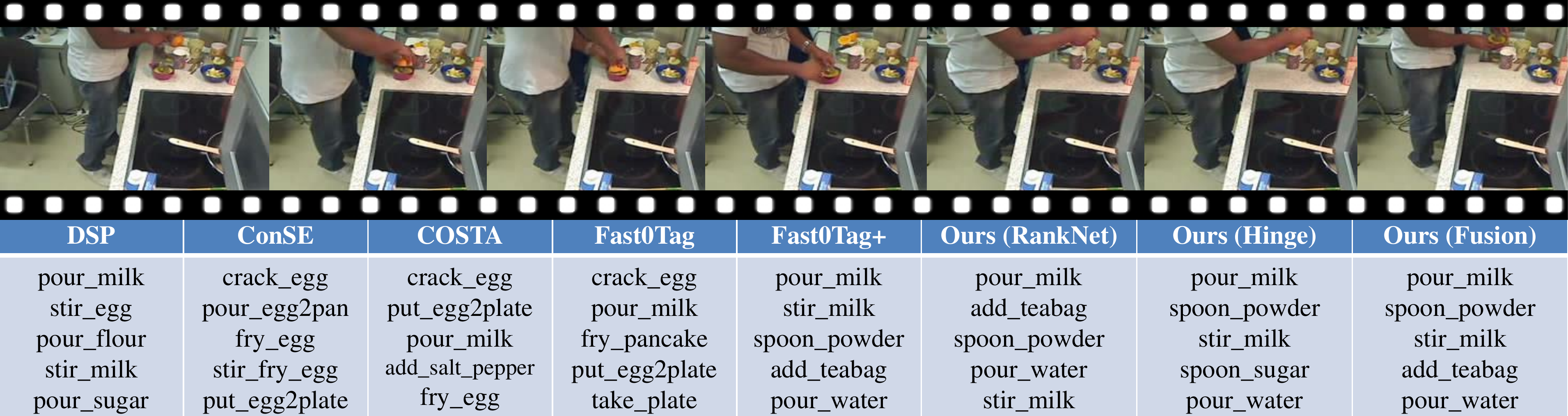}
	\caption{A test video clip in the IFS setting and the top-5 labels predicted by different methods. Its ground-truth labels are  \textbf{cut\_orange}, \textbf{squeeze\_orange}, \textbf{pour\_juice}.}
	\label{fig_video653}
\end{figure*}

\begin{figure*}[htbp]
	\includegraphics[width=\linewidth]{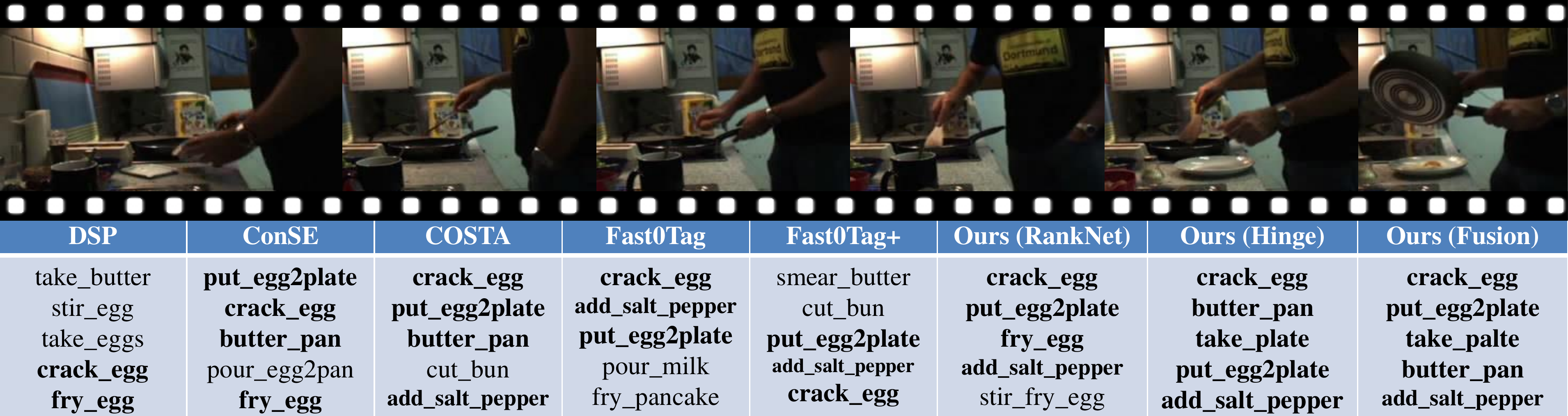}
	\caption{A test video clip in the IFS setting and the top-5 labels predicted by different methods. Its ground-truth labels are \textbf{crack\_egg}, \textbf{fry\_egg}, \textbf{put\_egg2plate}, \textbf{take\_plate}, \textbf{add\_salt\_pepper}, \textbf{butter\_pan}. }
	\label{fig_video465}
	
\end{figure*}

\begin{figure*}[htbp]	
	\includegraphics[width=\linewidth]{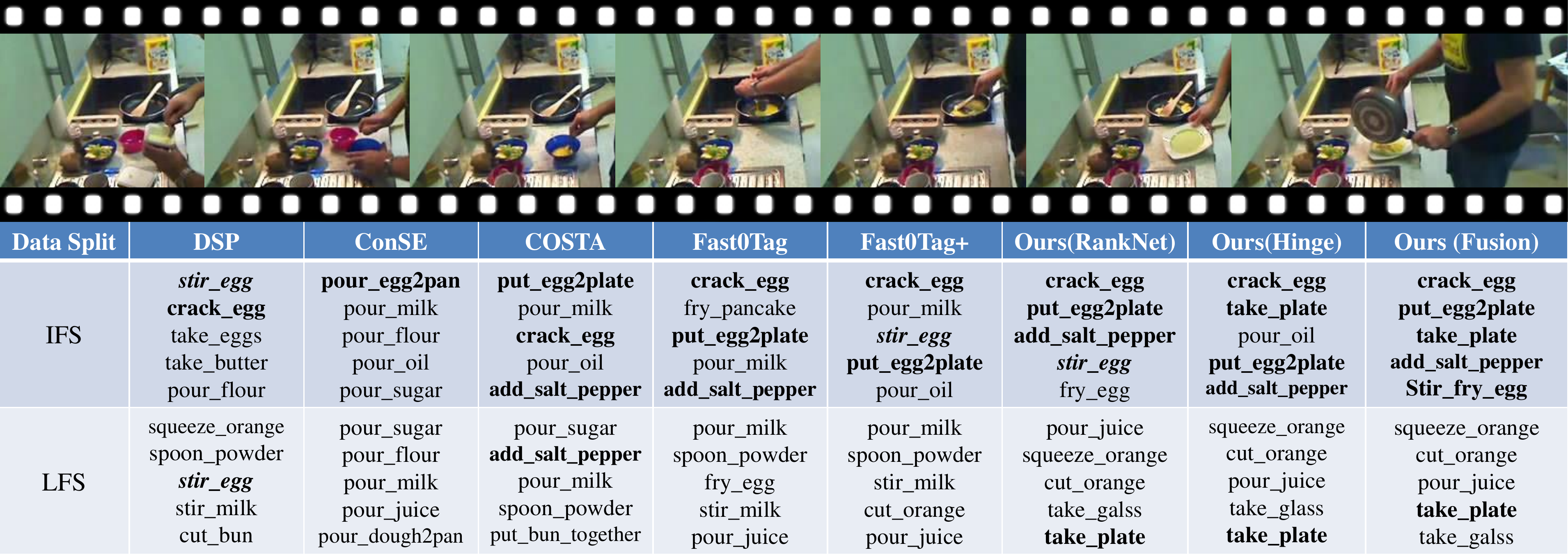}
	\caption{A test video clip appearing in in the IFS and LFS settings and the top-5 labels predicted by different methods in two data split settings. Its ground truth labels are \textbf{take\_bowl}, \textbf{crack\_egg}, \textbf{put\_egg2plate},  \textbf{take\_plate}, \textit{\textbf{stir\_egg}}, \textbf{pour\_egg2pan}, \textbf{stir\_fry\_egg}, \textbf{add\_salt\_pepper}, \textbf{butter\_pan} in the IFS setting, and \textit{\textbf{take\_bowl}}, \textbf{crack\_egg}, \textbf{put\_egg2plate}, \textbf{take\_plate}, \textit{\textbf{stir\_egg}}, \textit{\textbf{pour\_egg2pan}}, \textbf{stir\_fry\_egg}, \textbf{add\_salt\_pepper}, \textit{\textbf{butter\_pan}} in the LFS setting, respectively.}
	\label{fig_video505IFS3}
\end{figure*}

\subsection{Visual Inspection}
\label{subsect:res_vis}

In general, visual inspection provides a manner that helps us understand the behaviour of a method intuitively. To gain an intuitive insight into the multi-label zero-shot human action recognition, we visualize a number of typical test video clips on Breakfast and the top-5 labels predicted by different state-of-the-art methods described in Section \ref{sect_comparative} and ours in terms of semantic relatedness scores. Our visual inspection mainly focuses on understanding of the behaviour of our model and issues arising from our work.
As a result, Figures \ref{fig_video275}-\ref{fig_video505IFS3} illustrate several key frames to human actions in typical test video clips and the top-5 predicted labels by different methods, where a correctly predicted known label is highlighted with bold font and a correctly predicted unseen label is marked with bold-italic font.

For the IFS setting, Figures \ref{fig_video275}-\ref{fig_video465} illustrate three typical results yielded by different methods.
Figure \ref{fig_video275} exemplifies the success of our model, where four out of the top-5 labels predicted by our model are the ground-truth actions and no other methods can match the performance of our model. This exemplified test instance suggests that the use of an LSTM layer in our visual model facilitates the recognition of distinctive actions in a video clip.
Figure \ref{fig_video653} shows a test instance where all the methods fail to have any ground-truth labels in their top-5 predicted labels.  Our visual inspection on this test instance reveals that non-trivial objects pertaining to different actions are concentrated in a small region located in top-right of frames  in this video clip. Thus, it is extremely difficult to capture the useful information in the visual domain, which poses a challenge to all the existing human action recognition techniques.
Figure \ref{fig_video275}-\ref{fig_video505IFS3} reveal that our models trained with two rank losses yield different results for a test instance. Specially in Figure \ref{fig_video505IFS3}, three of the top-5  labels predicted by two models are in common, however, the fusion method described in Section \ref{sect_multi-label ZSL} successfully predicts five ground truth labels. These instances vividly demonstrate the different aspects of two rank losses and the synergy achieved by their fusion.
Besides, these test instances illustrated in Figures \ref{fig_video275}-\ref{fig_video505IFS3}
also provide some insight regarding the behaviour of other state-of-the-art models used in our comparative study. For example, ConSE is more likely to yield the labels regarding frequently used words in a human action domain. For those test instances shown in Figures \ref{fig_video275}-\ref{fig_video465}, at least four out of the top-5 labels predicted by ConSE are regarding different actions taken on ``egg". For the instance shown in Figure  \ref{fig_video505IFS3}, all the top-5 labels predicted by ConSE are completely regarding ``pour" actions commonly taken in kitchen.
This limitation is due to the fact that ConSE uses a single collective semantic representation resulting from averaging the semantic representations of multiple co-occurred labels, which favors those frequently used word vectors but diminishes the opportunity of finding out infrequently used word vectors in prediction.

\begin{figure*}[htbp]
		\includegraphics[width=\linewidth]{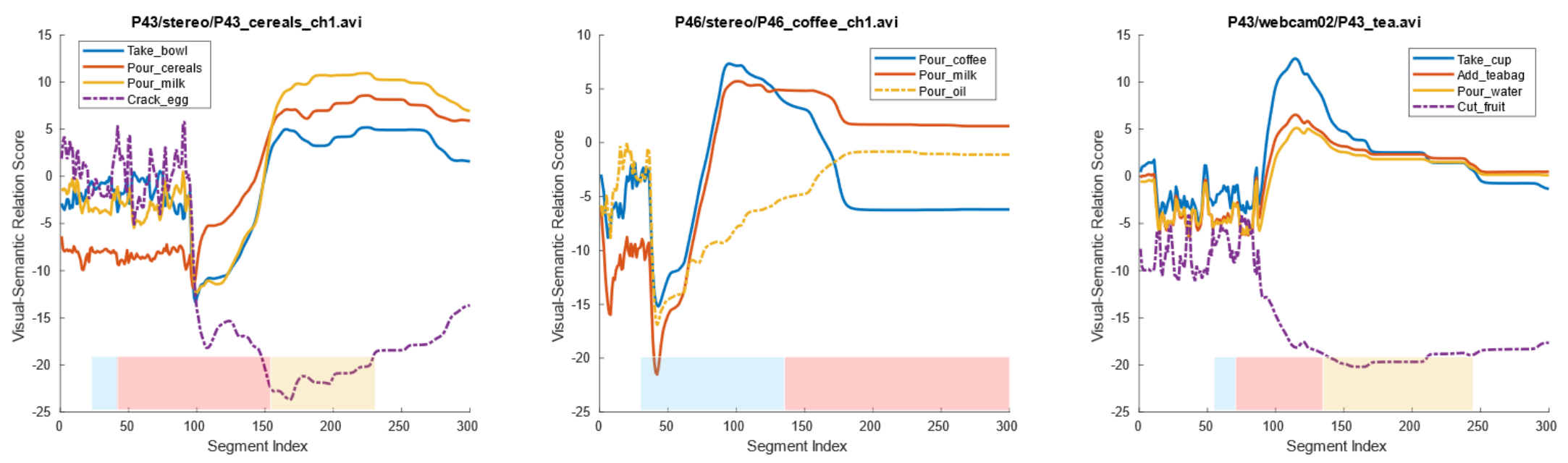}
		\caption{Visual-semantic relatedness scores of segments from three test video clips in the Breakfast dataset with the reference to the ground-truth location of actions. The horizontal bars (colored with blue, red and yellow) denote the ground-truth localization of different actions in a video clip. Their relatedness scores are depicted with solid lines of the corresponding colors while the dashed lines expresses the related scores of an illustrative negative action.}
		\label{fig_vsscores}
\vspace*{-3mm}
\end{figure*}

Experimental results reported in Tables \ref{table_baseline_log}-\ref{table_comparative}
suggest that all the  models including ours generally perform worse under the LFS setting than under the IFS setting. On the one hand, the LFS setting results in a \emph{training data sparsity} issue in contrast to the IFS setting. To see this issue, let us take the first split on Breakfast  as an example. In this split shown in Table \ref{table_datasplit}, there are 1,196 and 1,019 training examples in the IFS and the LFS setting, respectively. However, the number of training examples pertaining to specific known actions is significantly different in two split settings due to different data split protocols described in Section \ref{sect_datasplit}. For example, there are 330, 217, 254, 109 and 156 training examples with target labels,``crack\_egg", ``put\_egg2plate", ``take\_plate", ``stir\_fry\_egg" and ``add\_salt\_pepper", respectively, in the IFS setting. In contrast, there are only 23, 23, 81, 23 and 20 examples with the same target labels, respectively, in the LFS setting.
On the other hand, there is a major difference between those models and ours; i.e., our visual model employs a hidden layer of recurrent connection to capture temporal coherence underlying intrinsic visual features while those state-of-the-art models used in our comparative study do not have such a mechanism. It is well known that a learning model of a higher complexity or a larger capacity demands more informative training data. To this end, the training data sparsity issue affects the performance of our model more severely than other models; it is evident that the performance gain from the use of an LSTM layer in our visual model disappears due to a lack of sufficient training data required in training our visual model for capture temporal coherence.

To understand the difference between the IFS and the LFS settings and the training data sparsity issue intuitively, we illustrate the results yielded by the state-of-the-art methods and ours on a common instance appearing in test sets in two data split settings, as shown in Figure \ref{fig_video505IFS3}. It is evident that four out of the top-5 action labels predicted by our model are the ground truth and all other models can predict some of ground-truth actions correctly under the IFS setting. In contrast, however, none of the models correctly predicts more than one ground-truth action for this exactly same test instance under the LFS setting. The visual inspection on this test instance clearly demonstrates the distinction between two data split settings; i.e., visual features associated with unseen actions are available in the IFS setting (an unrealistic scenario) but unavailable in the LFS setting (a realistic scenario), and the \emph{training data sparsity} issue in the LFS setting, which poses a big challenge to all the existing multi-label ZSL methods including ours.

\subsection{Correspondence between Visual and Semantic Embedding}
\label{subsect:correspond}

To understand the behavior of our framework further, we explicitly investigate how our framework captures the correspondence between visual and semantic embedding by comparing the related scores against the ground truth of action boundaries further annotated manually. To this end, we take a closer look at the visual-semantic relatedness scores in terms of video segments corresponding to multiple actions in a video clip of a long duration.

In our experiment,  we select three representative test video clips from the Breakfast dataset and exhibit the segment-level relatedness scores against the ground truth in Figure \ref{fig_vsscores}.  The horizontal  bars colored with blue, red and yellow in each plot express the ground-truth localization of actions in three video clips. Their relatedness scores are depicted with solid lines in the same colours, and the dashed lines denote the relatedness scores corresponding to some selected negative actions. It is observed from Figure \ref{fig_vsscores} that the relatedness scores are usually below zero for all the actions at the beginning and with subtle changes within a period until reaching the point after which the relatedness scores of positive actions start increasing whilst the scores of negative actions get decreasing or stable. By comparing the moment when the significant change of relatedness scores occurs against the ground-truth localization of actions, we observe that the relatedness scores do not necessarily correspond to the accurate action location in a video clip. By a closer look at the left and middle plots in Figure \ref{fig_vsscores}, one can see that the occurrence of a specific action would lead to a higher relatedness scores of this action than other positive actions not appearing within this visual segment during this period. However, this observation does not appear in the right plot. Hence, our experimental results suggest that our framework yields only reasonable relatedness scores once capturing sufficient visual information. The similar relatedness scores of multiple positive actions throughout the segment sequence indicate our framework learns the label co-occurrences well and hence recognize a set of actions collectively. However, it may fail in action localization which is beyond the scope of this paper.

\subsection{Model Complexity}

The architecture complexity of our learning model depends on the number of hidden layers, hidden units and their types as well as their connections used in neural networks to implement the visual and the semantic models for a data set.

In our current implementation, the  number of parameters in the visual model varies from  4.9 to 24.7 millions, and the number of parameters in the semantic model varies from 0.15 to 0.77 millions under different hyper-parameter settings. Obviously, the semantic model has much fewer parameters compared with the visual model, suggesting that introducing a semantic model does not incur a much higher computational burden but leads to the performance gain. In general, our model often takes longer training time than other state-of-the-art learning models used in our comparative study due to the use of a LSTM layer to capture temporal coherence.

Practically, with a GTX1080Ti GPU, the averaging time spent for training our learning model is roughly 13 minutes on Charades (i.e., 40s per epoch multiplies approximately 20 epochs) and one hour on Breakfast due to a larger number of time steps (T=300). One limitation of our learning model is a large memory requirement for training. Recall that the visual representations have to be reserved for use in the training of semantic model, it is required to load one large matrix with a size of  $n\times d_e \times T$ into memory . In our implementation, the amount of GPU memory used for training on Charades and Breakfast is 3.5 GB and 10.5 GB, respectively.

\section{Concluding Remarks}
\label{sect:conclusion}

In this paper, we have formulated human action recognition as a multi-label zero-shot learning problem and provide an effective solution by proposing a novel framework via joint latent ranking embedding learning. To carry out our framework, we employ a neural network of the heterogeneous architecture for visual embedding, where an LSTM layer is used to facilitate capturing temporal coherence information underlying different actions from weakly annotated video data. Also, we advocate the use of semantic embedding learning to facilitate bridging the semantic gap and effective knowledge transfer, which is implemented by a feed-forward neural network. All the above contributions have been thoroughly verified via our comparative study with various well-motivated settings. Experimental results on two benchmark multi-label human action datasets suggest that our proposed framework generally outperforms not only the baseline systems but also several state-of-the-art multi-label ZSL approaches in all the different test scenarios.

Although we have demonstrated favourable results on two benchmark datasets in comparison to state-of-the-art approaches, our observations on the performance of all the approaches used in our comparative study including ours suggest that the existing multi-label ZSL techniques are not ready for a real application; the instance-first split setting fails to simulate real multi-label zero-shot human action recognition scenarios while the performance becomes even worse under the label-first split setting that simulates a real scenario.
Nevertheless, our experimental results including visual inspection provide the insightful information for improving our proposed framework. In our ongoing work, we would address issues arising from our experiments and observations with proper techniques. To address the \emph{training data sparsity} issue revealed in our
experiments, we would develop unsupervised learning algorithms to discover salient yet intrinsic visual features from unlabelled video clips and further incorporate proper temporal constraints into our rank loss functions to better capture temporal coherence. Moreover, we would consider diverse pooling strategies and introduce attention mechanisms to our model for improving implicit salient feature extraction and accurate localization of different yet complex actions involved in a video clip during the visual embedding learning. Also, we would employ alternative semantic representations developed by ourselves \cite{wang17alternative}, which encode the semantic relatedness between action labels more accurately, in the semantic embedding learning to facilitate knowledge transfer.

While our framework is proposed especially for multi-label zero-shot human action recognition, we would highlight that it is directly applicable to multi-label human action recognition without modification as demonstrated in our experiments. Also, our framework is easy to adapt for tackling various multi-label ZSL problems in different domains. For example, we can apply our framework to miscellaneous multi-label zero-shot classification tasks on temporal or sequential data, e.g., acoustic event classification,  straightforward as well as multi-label zero-shot learning tasks on static data, e.g., object recognition, by replacing a neural network of the heterogeneous architecture only with a neural network of only feed-forward connections in the visual model. Thus, we are going to explore such extensions and applications in our future work.

\section*{Acknowledgement}

The authors would like to thank the anonymous reviewers for their valuable comments that improve
the presentation of this manuscript.

\bibliographystyle{elsarticle-num}

\footnotesize
\bibliography{mybibfile}  
\end{document}